\documentclass{article}

\usepackage{arxiv}

\usepackage[utf8]{inputenc} 
\usepackage[T1]{fontenc}    
\usepackage{hyperref}       
\usepackage{url}            
\usepackage{booktabs}       
\usepackage{amsfonts}       
\usepackage{amsmath}        
\usepackage{amssymb}        
\usepackage{multirow}       
\usepackage{nicefrac}       
\usepackage{microtype}      
\usepackage{graphicx}
\usepackage{color}

\graphicspath{{./}}

\title{A Multi-Agent System for Motor Design Optimization Using an FEA--AI Hybrid Approach}

\author{
  Jinseong Han \\
  Cho Chun Shik Graduate School of Mobility\\
  Korea Advanced Institute of Science and Technology (KAIST)\\
  Daejeon 34051, Republic of Korea \\
  \texttt{nethan1012@kaist.ac.kr} \\
  \And
  Sunwoong Yang\textsuperscript{*} \\
  Department of Mechanical Engineering\\
  Hanyang University\\
  Ansan 15588, Republic of Korea \\
  \texttt{sunwoongy@hanyang.ac.kr} \\
  \And
  Namwoo Kang\textsuperscript{**} \\
  Cho Chun Shik Graduate School of Mobility\\
  Korea Advanced Institute of Science and Technology (KAIST)\\
  Daejeon 34051, Republic of Korea \\
  Narnia Labs, Daejeon, Republic of Korea \\
  \texttt{nwkang@kaist.ac.kr} \\
}

\begin{document}
\maketitle
\renewcommand{\thefootnote}{}%
\footnotetext{\textsuperscript{*, **}Corresponding authors.}%

\begin{abstract}
This study presents a large language model (LLM)-based multi-agent framework for interior permanent magnet synchronous motor (IPMSM) design optimization that mitigates limitations of conventional workflows: expertise-dependent problem setup and data preparation, the prohibitive computational cost of finite element analysis (FEA), and the unreliability of AI surrogates in unexplored regions. To this end, we first introduce a \textit{Design agent} that formulates the optimization problem in natural language, leveraging retrieval-augmented generation to improve answer accuracy on motor-design problems from below 50\% to 67--80\%. Furthermore, a \textit{Training agent} autonomously repairs improperly defined design spaces by reasoning over solver-failure history, raising the success ratio of the geometry sampling from 28\% to 84\% for AI training. Additionally, to resolve cost and reliability simultaneously, an \textit{Optimization agent} employs an uncertainty-aware FEA--AI hybrid model: the AI surrogate is the primary evaluator, and FEA is selectively invoked where predictive uncertainty is high. Under the same FEA budget, this hybrid model achieves up to 44\% lower iron loss in single-objective and 22.5\% higher hypervolume in multi-objective optimization than conventional FEA-only search. Under the same evaluation budget, it reduces computation time by 52--55\% while retaining 90--92\% of FEA-only hypervolume. Conversely, AI-only search converges to false optima, leaving half its Pareto designs infeasible. Notably, a controller agent adaptively updates the uncertainty threshold that triggers FEA each round, eliminating manual tuning and achieving 5.8\% lower single-objective iron loss than with a fixed threshold. These results establish domain-specialized LLM agents with uncertainty-aware hybrid evaluation as a reliable, scalable paradigm for simulation-driven design automation.
\end{abstract}


\section{Introduction}

The demand for high power density and high efficiency in electric vehicles and industrial drives has increased the importance of optimal design for interior permanent magnet synchronous motors (IPMSMs)~\cite{liu2016research,pellegrino2012comparison}. However, practical IPMSM optimization remains difficult because engineers must simultaneously balance conflicting objectives and coupled multi-physics constraints. Typical objectives include maximizing average torque and minimizing losses, torque ripple, and cogging torque, while typical constraints include demagnetization margin, rotor-bridge stress/thickness, voltage/current limits, and manufacturing minimum thickness~\cite{ji2024design,vidanalage2018multimodal,liu2022design}. Mature evolutionary algorithms can handle such multi-objective, multi-constraint formulations, so meeting these requirements now depends less on the optimization algorithm itself than on how the surrounding design workflow is set up, evaluated, and automated. Although recent advances in large language model (LLM) agents offer new possibilities for such automation, their direct application to motor design is hindered by domain-specific barriers. This paper distills these difficulties into four coupled bottlenecks (manual design and data preparation, FEA cost, surrogate reliability, and the lack of domain-adapted LLM automation) and proposes a multi-agent framework that addresses each of them.

The first bottleneck is the manual design and data-preparation workflow that surrounds the optimizer. Engineers must repeatedly define design variables, objective/constraint formulations, analysis conditions, and automation parameters before optimization can be run; junior engineers with limited motor-design domain knowledge struggle with this high-dimensional setup, senior engineers rebuild it whenever requirements change, and these setup and operation tasks can constitute a substantial share of total optimization lead time~\cite{kang2025generative}. The same manual burden extends to data preparation. Reliable per-variable bounds are rarely known in advance, and under wide or uncertain ranges many geometries sampled by design of experiments (DOE) are infeasible through component intersections or non-manufacturable, non-meshable shapes, so they fail geometry validation or FEA meshing and shrink the analysis-feasible training set~\cite{nakata2017automatic}. Conventional pipelines typically respond by discarding the infeasible samples or by imposing conservative bounds that limit design-space exploration, leaving the range refinement itself as manual trial and error.

Once the problem is defined and a feasible dataset secured, a second bottleneck appears in the optimization itself: the computational cost of high-fidelity evaluation. Motor objectives and constraints are available only as outputs of an FEA solver that provides no analytic gradients, so practical IPMSM optimization relies mainly on population-based evolutionary algorithms such as the genetic algorithm (GA) and the non-dominated sorting genetic algorithm II (NSGA-II), treating the solver as a black box~\cite{deb2002nsga2,kim2025optimal}. The robustness of these algorithms comes from evaluating many candidates: the number of evaluations scales with population size $\times$ number of generations, so a single optimization readily demands thousands of evaluations. Each evaluation, moreover, involves far more than a single electromagnetic solve. Practical motor design typically handles multiple conflicting objectives~\cite{duan2013review}, incorporates multi-physics factors such as structural stress and thermal behavior~\cite{bramerdorfer2018modern}, and, for industrial design targets such as efficiency maps, assesses each candidate geometry over hundreds of operating points~\cite{fatemi2016large}. Attempting to meet this entire evaluation load with FEA alone, at up to several hours per candidate, therefore incurs an enormous computational cost, which stands as the fundamental limitation of FEA-only optimization. Under a fixed computational budget, the only recourse is to shrink the population or the number of generations, which narrows exploration and limits attainable solution quality.

To reduce this burden, many studies adopt AI-surrogate-based optimization (often coupled with GA), replacing most of the expensive performance evaluations with low-cost predictions~\cite{jin2011surrogate,shin2023topology}; in IPMSM design, this approach has been developed furthest through generative adversarial network (GAN)-based data augmentation combined with convolutional neural network (CNN) or vision-transformer predictors~\cite{shimizu2022automatic,shimizu2024automatic}, greatly accelerating candidate evaluation. However, the reliance on surrogates introduces a third bottleneck: reliability. Because a surrogate is trained on a finite dataset, its predictive confidence is inherently uneven over the design space and regions of low reliability are unavoidable~\cite{lakshminarayanan2017deep,hullermeier2021aleatoric,abdar2021review}; an optimization guided by such a model therefore inherits this limited reliability. The problem deepens in out-of-distribution (OOD) regions, where prediction accuracy itself degrades~\cite{asanuma2025multi}. Without in-loop high-fidelity correction, these errors accumulate through candidate selection and model-guided updates until the search converges toward designs that appear optimal only according to the surrogate~\cite{jones2001taxonomy,queipo2005surrogate}. To resolve the cost bottleneck of FEA and the reliability bottleneck of surrogates together, FEA--AI hybrid workflows have emerged that re-verify selected candidates with FEA; in existing practice, however, the decision of when and where to invoke FEA rests on engineer-defined heuristics, making the verification ad hoc and hard to reproduce across problems~\cite{gong2025comparative,ma2024topology}. A principled criterion for this decision does exist: Bayesian optimization and variance-based active learning allocate expensive evaluations exactly where a probabilistic surrogate is least certain~\cite{jones1998ego,forrester2009recent,bae2024cfd}. Yet these methods are formulated for sequential single-point search, leaving their uncertainty-driven allocation unexploited in the population-based evolutionary loops on which motor design practice relies.

A fourth bottleneck is the lack of LLM automation that is specialized to the motor domain and deployable in-house. LLM agents have recently begun to automate design and optimization workflows~\cite{guo2026llm,ren2026agenticontrol}, including multi-agent airfoil design optimization and LLM-assisted hyper-heuristic search~\cite{fan2026airfoilagent,zhong2025llmoa}; to the best of our knowledge, however, such frameworks have not yet been applied to electric-motor design optimization. Nor is simply transplanting them viable, as they are built on general-purpose commercial API backbones. These closed-weight backbones lack motor-domain knowledge and leave users little room to tune the model for their own domain and intended use, so the resulting automation cannot be specialized to the target task. Passing proprietary geometries, requirements, and performance data through an external service also raises security and confidentiality concerns~\cite{yao2024llmsecurity}. Filling this gap therefore requires more than adopting an existing framework: it calls for LLM automation that users can tune to their own domain and purpose and deploy fully in-house.

To address these four bottlenecks, we propose an end-to-end multi-agent framework for IPMSM design optimization based on LLM-driven agent collaboration. A \textit{Design agent} grounds the problem-formulation dialogue in motor-textbook knowledge through retrieval-augmented generation (RAG)~\cite{lewis2020rag,wan2025empowering}, so that engineers with limited domain expertise can define the optimization problem in natural language. A \textit{Training agent} then secures the analysis-feasible training dataset, autonomously correcting an improperly defined design space by reasoning over geometry-validation and solver-failure logs. An \textit{Optimization agent} conducts the evolutionary search with an uncertainty-aware FEA--AI hybrid model, which switches each candidate evaluation between the fast AI surrogate and high-fidelity FEA, bringing the uncertainty-driven evaluation allocation of Bayesian optimization and active learning into the population-based evolutionary loop and thereby improving both the efficiency and reliability of the optimization process. Finally, every agent runs on a locally deployed 20B open-weight LLM adapted to the motor domain through retrieval grounding, keeping proprietary design data fully on-premises.

Our main contributions are as follows:
\begin{itemize}
  \item \textbf{Fully autonomous IPMSM design.} We introduce a multi-agent system that makes the entire design workflow autonomous from problem formulation to a verified optimal design. The \textit{Design agent} formulates the optimization problem through natural-language interaction, so that engineers with limited domain expertise can easily define the design problem. The \textit{Training agent} autonomously repairs improperly defined design spaces, executes FEA to build the training dataset, and trains the AI surrogate model on it. The \textit{Optimization agent} then completes the design search with the FEA--AI hybrid model.
  \item \textbf{FEA--AI hybrid model.} In the optimization stage, we introduce an FEA--AI hybrid model that uses the AI surrogate as the primary evaluator and calls FEA where the predictive uncertainty provided by the surrogate is high, simultaneously resolving the cost bottleneck of FEA and the accuracy (reliability) bottleneck of the AI surrogate. We conduct an ablation study of this hybrid model against FEA-only and AI-only baselines, which reveals that it achieves the best overall trade-off among solution quality, reliability, and computation cost.
  \item \textbf{The first local-LLM-based framework specialized for motor design.} To the best of our knowledge, this is the first study to apply an LLM-based multi-agent framework to motor design and optimization. We deploy an open-weight LLM locally, ground it in motor-design knowledge through RAG, and employ it as dedicated agents, each specialized for a specific stage of the workflow. This serves as a proof of concept that a domain-specialized LLM system for design automation can be built entirely in-house, with proprietary design data kept on-premises.
\end{itemize}

The remainder of this paper is organized as follows. Section~2 presents the proposed methodology, including the \textit{Design}, \textit{Training}, and \textit{Optimization agents}. Section~3 reports two preliminary studies: a RAG study that compares \textit{Design-agent} outputs with and without motor-domain retrieval grounding (Section~3.1) and a demonstration of the log-informed resampling loop on a representative user-defined scenario (Section~3.2). Section~4 presents the FEA--AI hybrid model setup together with a threshold-sensitivity analysis of rule-based switching (Section~4.1) and an agent-based switching study (Section~4.2). Section~5 presents an ablation study of FEA-only, AI-only, and FEA--AI hybrid model (rule-based and agent-based switching) strategies: first on a single-objective problem (Section~5.1) and then on a multi-objective problem (Section~5.2), each under the same FEA-call budget and the same evaluation budget. Section~6 concludes the paper.

\section{Methodology}

\begin{figure*}[!t]
\centering
\includegraphics[width=0.98\linewidth]{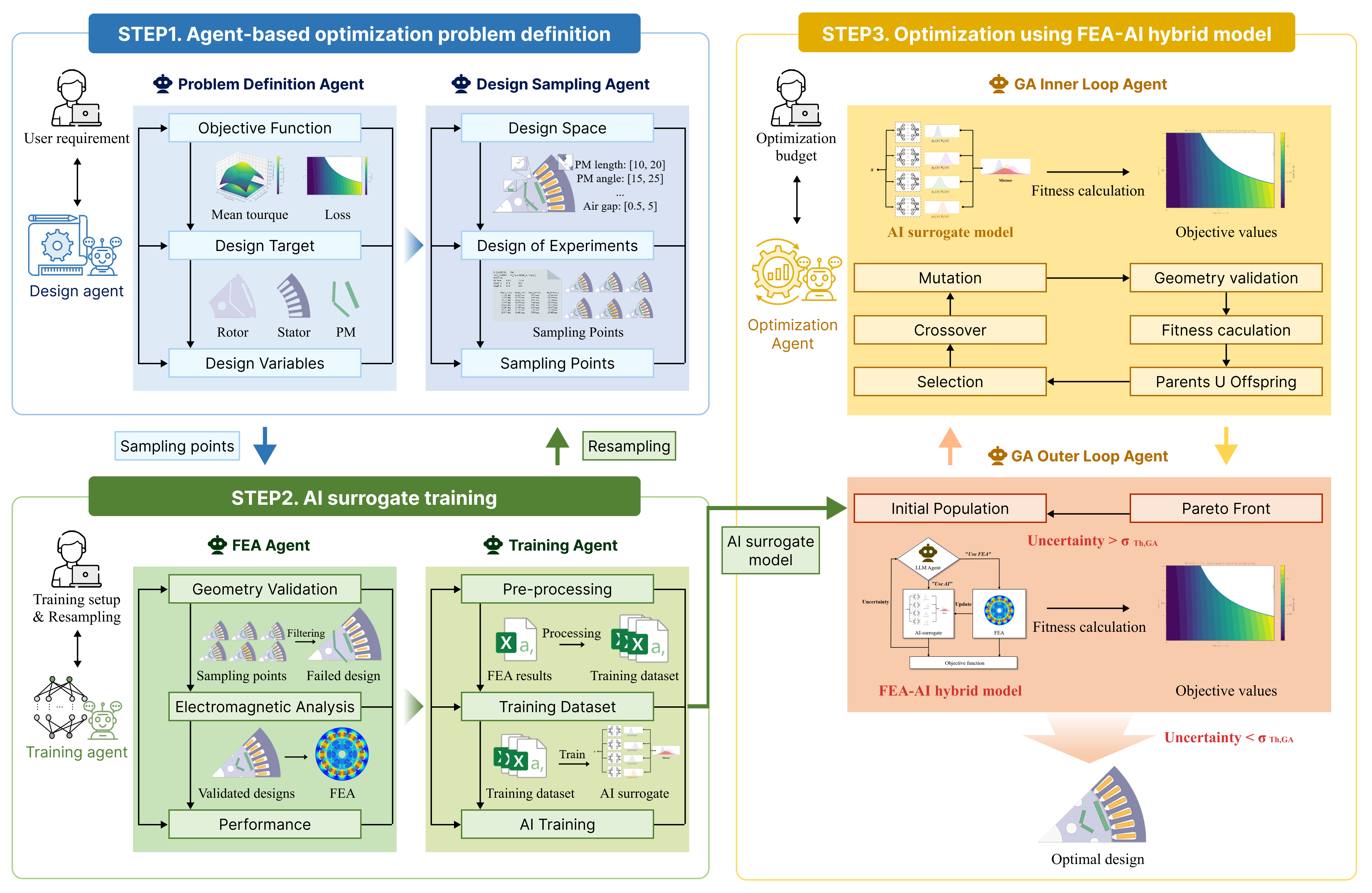}
\caption{Overall framework of the proposed multi-agent system for IPMSM design optimization. Three orchestration agents coordinate their sub-agents across the design, training, and optimization stages.}
\label{fig:overall_framework}
\end{figure*}

To address the challenges identified in Section~1, we propose an end-to-end multi-agent FEA--AI hybrid framework for IPMSM design optimization. As shown in Fig.~\ref{fig:overall_framework}, the proposed methodology has three stages, and in each stage a top-level agent acts as an orchestrator that decomposes the stage objective into specialized sub-agent tasks and coordinates the corresponding sub-agents:
\begin{itemize}
  \item \textbf{Step~1 -- \textit{Design agent}:} formalizes optimization requirements and generates DOE samples through its \textit{problem definition} and \textit{design sampling} sub-agents.
  \item \textbf{Step~2 -- \textit{Training agent}:} uses its sub-agents to analyze infeasible geometries and resample the design space, to execute electromagnetic FEA, and to train a surrogate model capable of uncertainty quantification (UQ).
  \item \textbf{Step~3 -- \textit{Optimization agent}:} conducts the optimization search with two sub-agents: a GA \textit{inner-loop agent} performs the routine evolutionary search, while a GA \textit{outer-loop agent} evaluates reliability-critical candidates with the FEA--AI hybrid model through uncertainty-driven switching.
\end{itemize}
The agents assigned to the three steps exchange standardized artifacts and can invoke one another when problem redefinition, resampling, surrogate update, or optimization refinement is required, allowing the workflow to proceed through agent-to-agent interaction even without intermediate user intervention. Before detailing the three steps in Sections~2.1--2.3, we first describe the shared agent-serving setup on which all three stages run.

\begin{figure}[!htbp]
\centering
\includegraphics[width=0.7\linewidth]{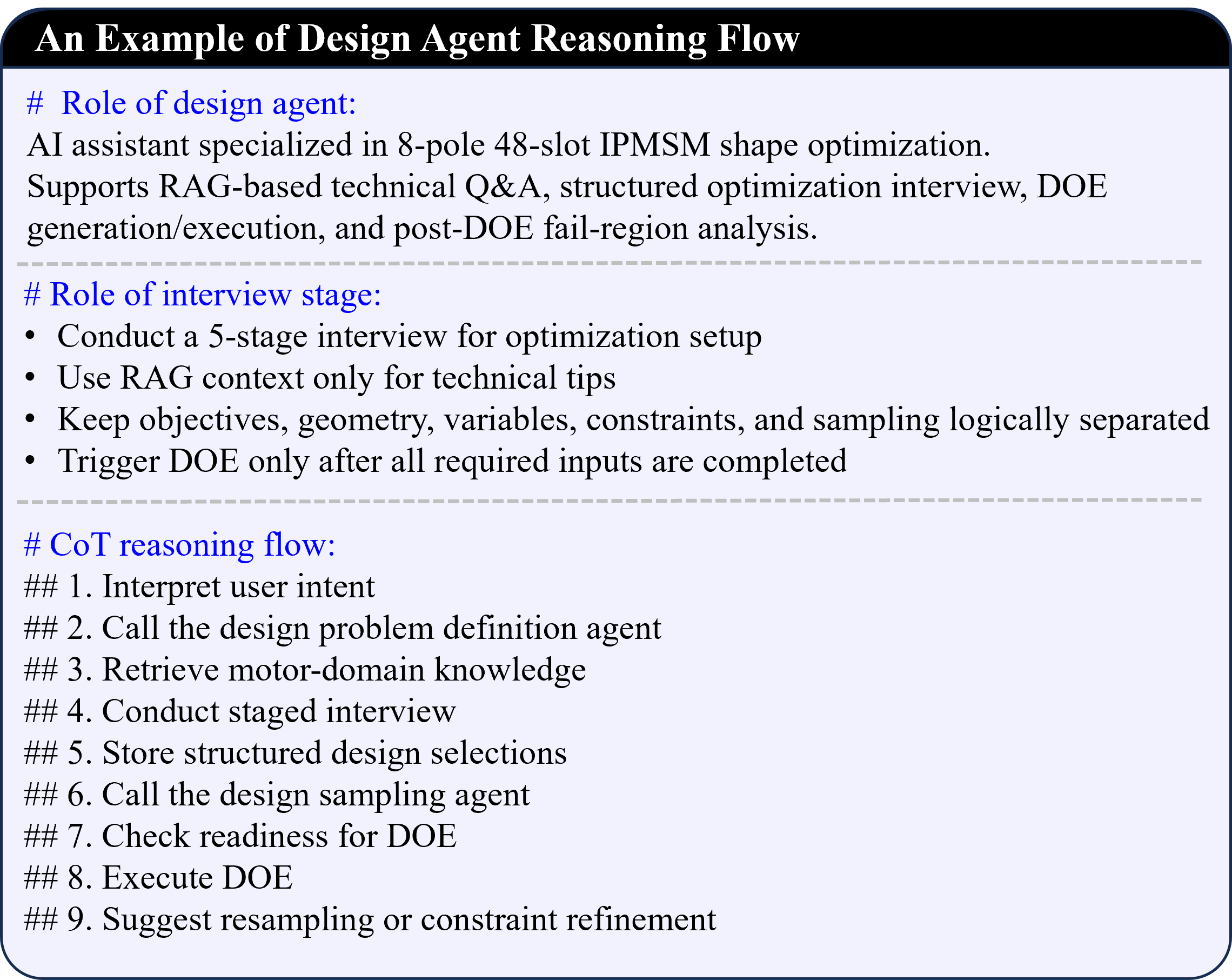}
\caption{Role-assignment and chain-of-thought (CoT) prompt template of the \textit{Design agent}.}
\label{fig:design_agent_cot_prompt}
\end{figure}

\paragraph{Agent setup.}
The multi-agent system adopts GPT-oss 20B as a lightweight open-weight local LLM backbone and runs it through an Ollama-based local serving environment. Compared with larger 60B-scale models or cloud-hosted proprietary services such as Gemini-, Claude-, or GPT-5-class APIs, this 20B-parameter model is easier to deploy in local engineering environments with lower compute overhead while retaining sufficient instruction-following and reasoning capability for structured agent tasks. This design is practical for confidentiality-sensitive motor-design projects because prompts, intermediate artifacts, and optimization datasets can remain on-premises. This backbone choice is validated quantitatively in Section~3.1.

Building on this local deployment setup, we implement agent orchestration with LangChain and LangGraph, and assign each agent a prompt-defined role using RAG+CoT templates with explicit input requirements, reasoning order, and structured output constraints (Fig.~\ref{fig:design_agent_cot_prompt}). The same template family is specialized for each agent's task: retrieval-grounded recommendation for the \textit{Design agent}, log-informed reasoning for the \textit{Training agent}, and rule-constrained switching decisions for the \textit{Optimization agent}, as detailed in Sections~2.1--2.3.

\begin{figure}[!htbp]
\centering
\includegraphics[width=0.9\linewidth]{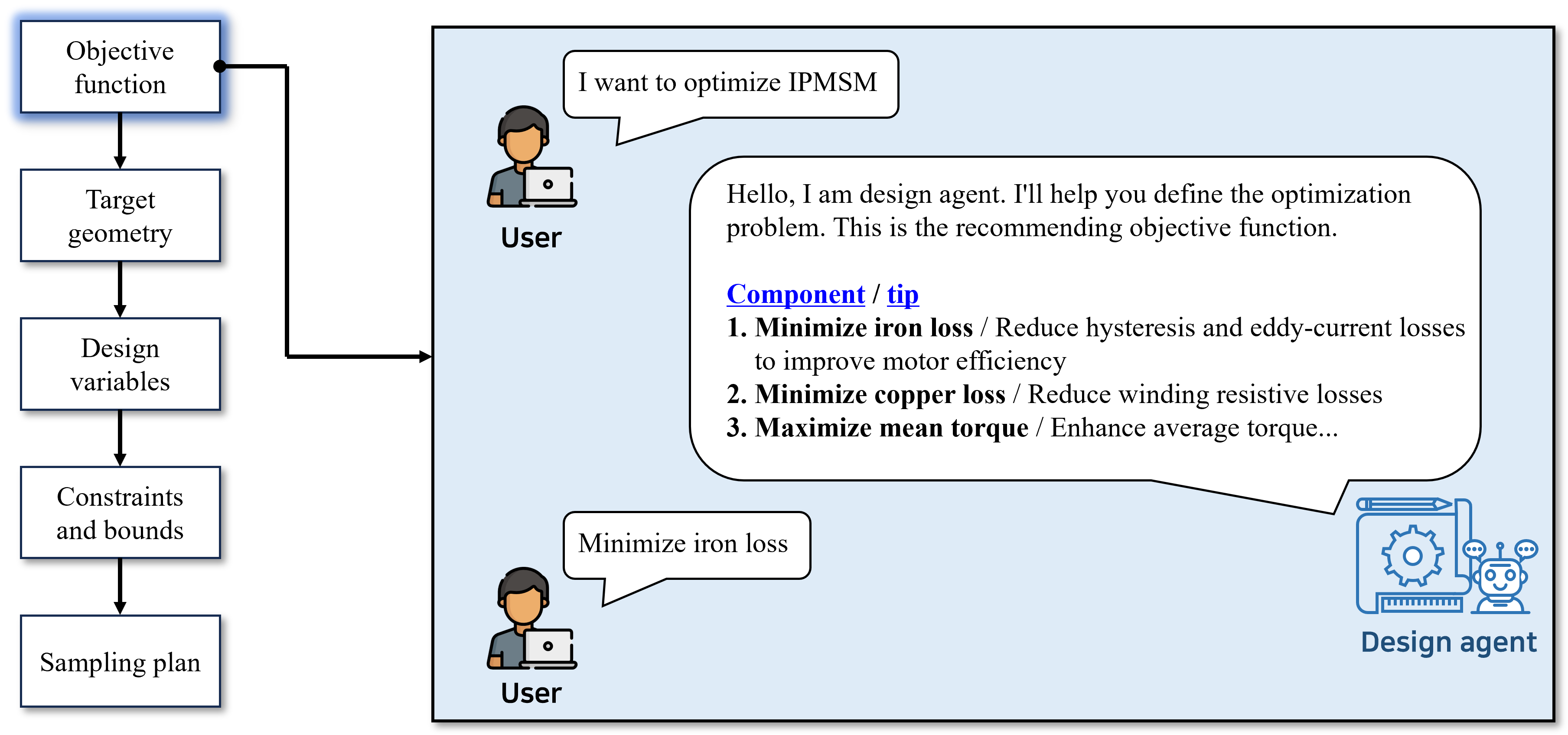}
\caption{Example problem-formulation interaction with the \textit{problem definition agent} for defining the objective function.}
\label{fig:design_agent_chat}
\end{figure}

\subsection{Design agent (Agent-guided optimization problem definition)}
The \textit{Design agent} orchestrates two sub-agents: a \textit{problem definition agent} and a \textit{design sampling agent}. The objective of this stage is to transform vague natural-language requests into a structured optimization card and a DOE-ready sampling plan with minimal manual tuning.

\paragraph{Retrieval pipeline.}
The \textit{Design agent} is grounded in a permanent-magnet motor reference~\cite{gieras2009permanent}. The source document is parsed, filtered to remove boilerplate, and segmented into overlapping passages, each of which is embedded with a local \texttt{nomic-embed-text} model and indexed in a FAISS vector store, while a lexical BM25 index~\cite{robertson2009bm25} is built over the same passages. At query time, the \textit{Design agent} retrieves through a \emph{hybrid retriever}: the dense (embedding) and BM25 rankings are
combined by reciprocal rank fusion~\cite{cormack2009rrf}, and the top $k{=}16$ passages are
injected into the agent's chain-of-thought prompt as a dedicated grounding block, so that
objective/variable candidates and engineering tips are conditioned on retrieved motor-domain
text rather than on parametric memory alone~\cite{lewis2020rag}. Combining lexical and dense
matching improves recall of both conceptual passages and specific numerical values reported in
the reference, as analyzed in Section~\ref{sec:rag_quant}. Generation uses the same
on-premises GPT-oss 20B backbone, so the full retrieval-and-reasoning pipeline remains local.

\subsubsection*{Problem definition agent}
Formulating the optimization problem is the first bottleneck identified in Section~1: a vague performance intent must be translated into an objective function, a design target, design variables, constraints and bounds, and a sampling plan, and each of these choices normally depends on expert motor-design knowledge. The \textit{problem definition agent} automates this translation as a RAG-grounded structured dialogue. The user first provides optimization requirements in natural language, and the agent, retrieving motor textbook/reference knowledge through the hybrid retrieval pipeline described above, guides the user through a sequential decision flow: objective function $\rightarrow$ design target $\rightarrow$ design variables $\rightarrow$ constraints and bounds $\rightarrow$ sampling plan~\cite{lewis2020rag}.

At each stage of this flow, the agent retrieves the passages relevant to the current decision, reasons over them under the role-assigning CoT template of Fig.~\ref{fig:design_agent_cot_prompt}, and returns candidate lists together with engineering tips that explain the practical consequences of each option. The user can accept a recommendation, adjust it, or supply an alternative in natural language, and each confirmed decision conditions the candidates proposed at the next stage. The stage-wise guidance covers:
\begin{itemize}
  \item \textbf{Objective function:} candidate objective functions and their trade-off criteria (illustrated in Fig.~\ref{fig:design_agent_chat});
  \item \textbf{Design target:} target-component candidates (rotor, permanent magnet (PM), stator, coil) and their relevance to the selected objectives (Appendix~\ref{app:problem_def}, Fig.~\ref{fig:pd_design_target});
  \item \textbf{Design variables:} variable candidates and the expected influence of each variable on the selected objectives (Appendix~\ref{app:problem_def}, Fig.~\ref{fig:pd_design_variables});
  \item \textbf{Constraints and bounds:} default constraints and variable bounds with accompanying risk notes (Appendix~\ref{app:problem_def}, Fig.~\ref{fig:pd_constraints});
  \item \textbf{Sampling plan:} sample-count guidance based on the number of selected variables and the available computational budget (Appendix~\ref{app:problem_def}, Fig.~\ref{fig:pd_sampling_plan}).
\end{itemize}
This staged decomposition reduces the high-dimensional setup task to a sequence of individually reviewable decisions: junior engineers receive domain guidance at every step rather than specifying the entire problem from scratch, while senior engineers can rapidly override the defaults when design requirements change.

Through this interactive process, the confirmed decisions are consolidated into a structured \emph{optimization card}, which serves as the machine-readable interface between the stages: the \textit{design sampling agent} reads it to generate the DOE plan, and the \textit{Training} and \textit{Optimization agents} of Steps~2 and~3 operate on the same card. Data generation and FEA--AI hybrid search therefore remain aligned with the user-confirmed problem definition, while the engineer retains a traceable checkpoint that can be inspected and revised before any FEA cost is incurred.

\subsubsection*{Design sampling agent}
\begin{figure}[!htbp]
\centering
\begin{minipage}{0.49\linewidth}\centering
  \includegraphics[width=\linewidth]{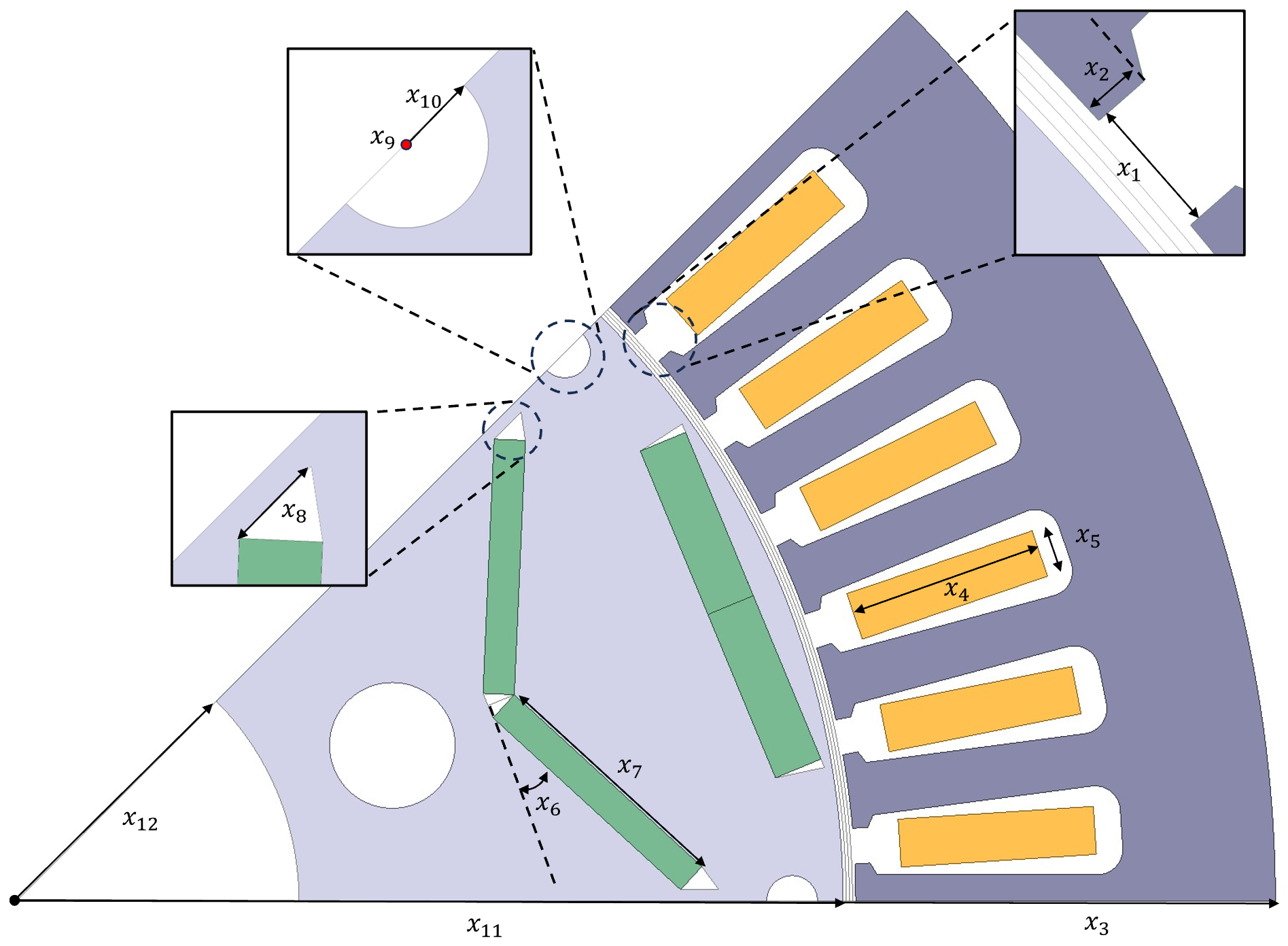}\\[0.2em]
  {\small (a)}
\end{minipage}\hfill
\begin{minipage}{0.49\linewidth}\centering
  \includegraphics[width=\linewidth]{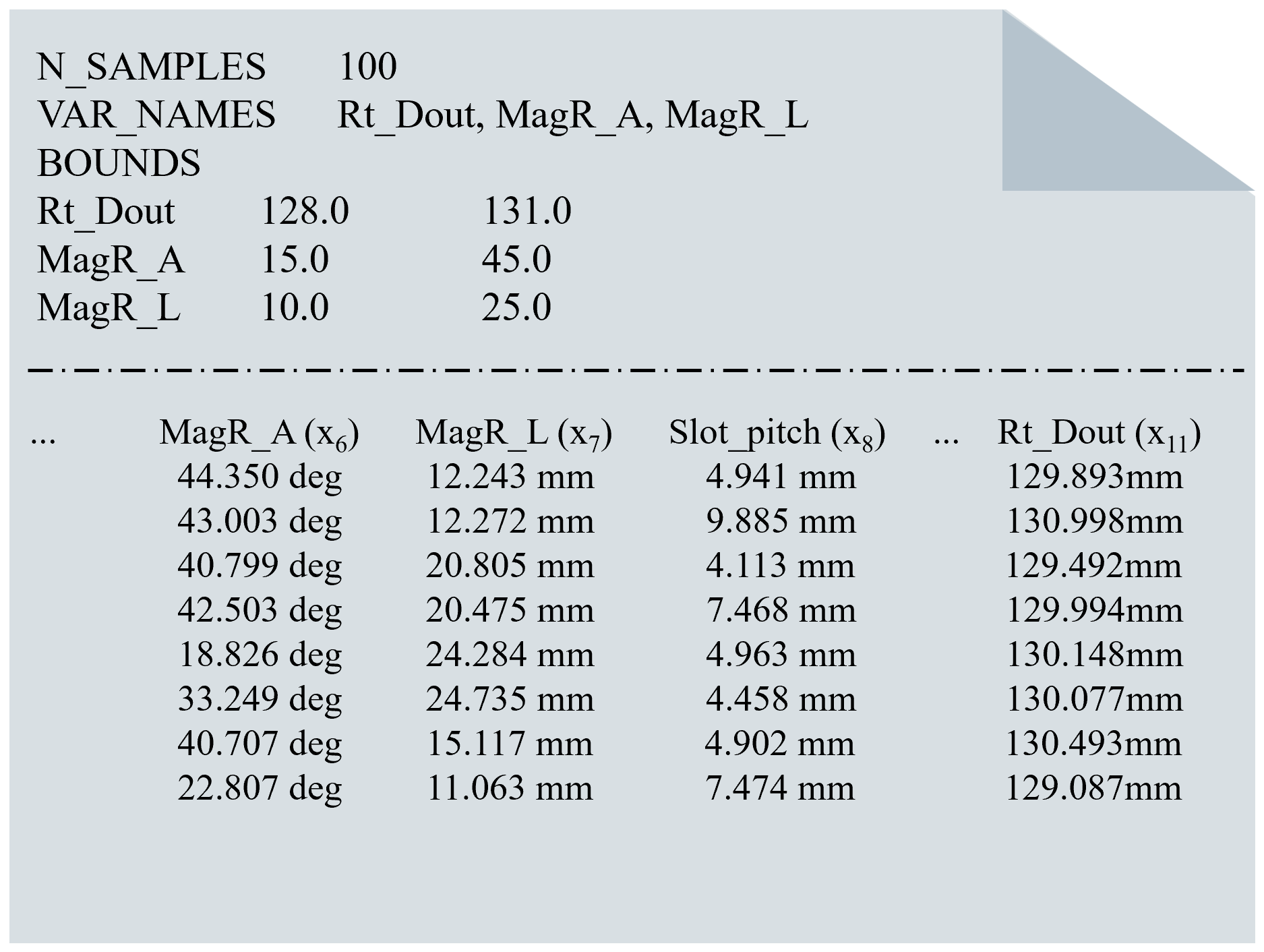}\\[0.2em]
  {\small (b)}
\end{minipage}
\caption{Design-sampling results after the \textit{Design-agent} interaction: (a)~Parameterized IPMSM rotor geometry with the twelve design variables selected through the dialogue. (b)~The DOE sampling points generated over the selected variable ranges.}
\label{fig:parameterized_ipmsm_doe}
\end{figure}

Once the interaction with the \textit{Design agent} is completed and the optimization problem is defined, the motor geometry is parameterized and linked to DOE generation, as shown in Fig.~\ref{fig:parameterized_ipmsm_doe}. In Fig.~\ref{fig:parameterized_ipmsm_doe}(a), the parameterized IPMSM is described using twelve geometric parameters: $x_1$ (gap between teeth tips), $x_2$ (teeth-tip height), $x_3$ (stator outer radius), $x_4$ (coil height), $x_5$ (coil width), $x_6$ (PM angle), $x_7$ (PM length), $x_8$ (slot pitch), $x_9$ (void location), $x_{10}$ (void radius), $x_{11}$ (rotor outer radius), and $x_{12}$ (rotor inner radius). Figure~\ref{fig:parameterized_ipmsm_doe}(b) shows sampling points generated by DOE in this parameter space~\cite{mckay1979lhs}.

Given the optimization card, the \textit{design sampling agent} defines variable-wise min/max bounds through user interaction. When users can specify reliable design ranges, the entered min/max values are used directly for DOE generation. Otherwise, the agent initializes provisional default bounds from the parameterized geometry template and domain-rule database, and these bounds can later be autonomously redefined through the log-informed resampling feedback loop described in Step~2. The sample count is then selected by the user according to the intended purpose and the available computational budget. The user-selected design variables are perturbed to generate multiple design candidates through DOE, while the remaining non-selected parameters are fixed at their default values before forwarding the DOE set to the \textit{Training agent}.

\subsection{Training agent (Dataset generation and surrogate training)}
The \textit{Training agent} orchestrates infeasible-geometry handling and resampling, electromagnetic analysis, preprocessing of electromagnetic-analysis outputs into a trainable dataset, and surrogate learning. This stage converts the design intent from Step~1 into a validated training dataset and a UQ-capable predictive model.

\subsubsection*{Resampling process}
\begin{figure}[!htbp]
\centering
\begin{minipage}{0.42\linewidth}\centering
  \includegraphics[width=\linewidth]{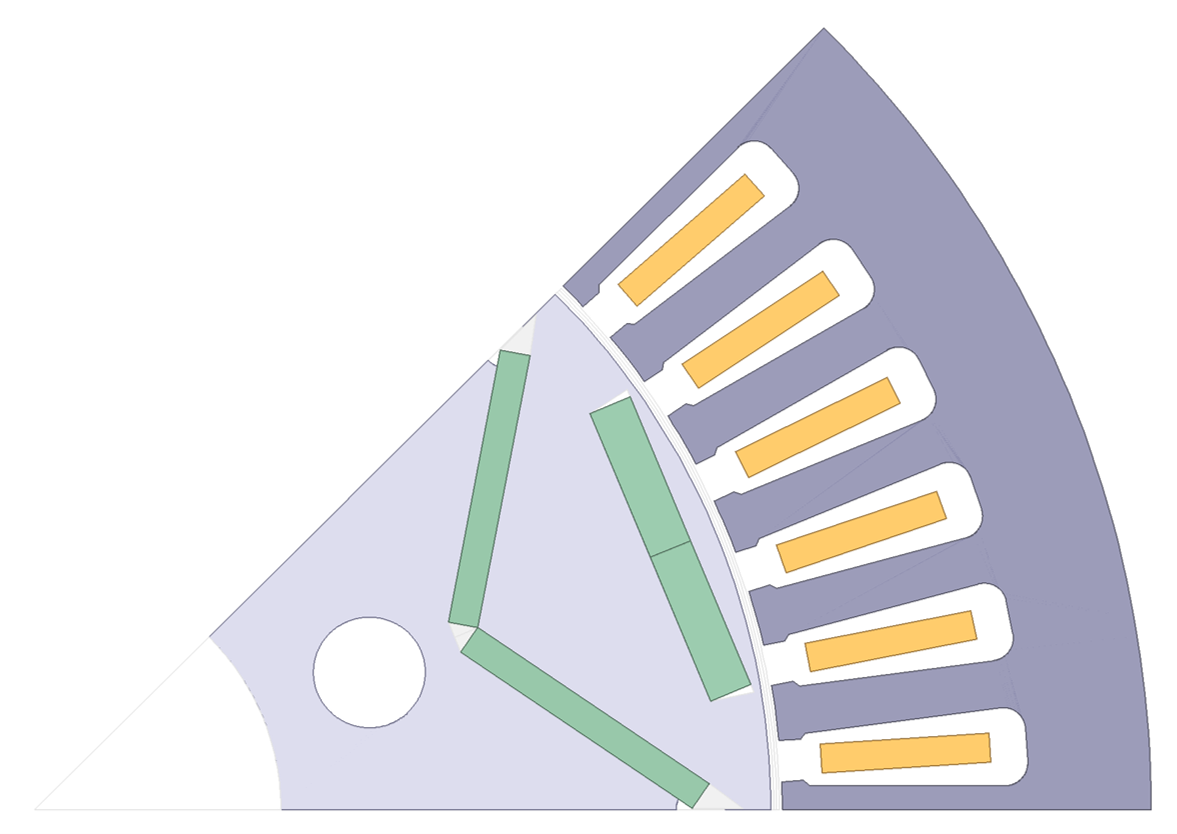}\\[0.2em]
  {\small (a)}
\end{minipage}\hfill
\begin{minipage}{0.42\linewidth}\centering
  \includegraphics[width=\linewidth]{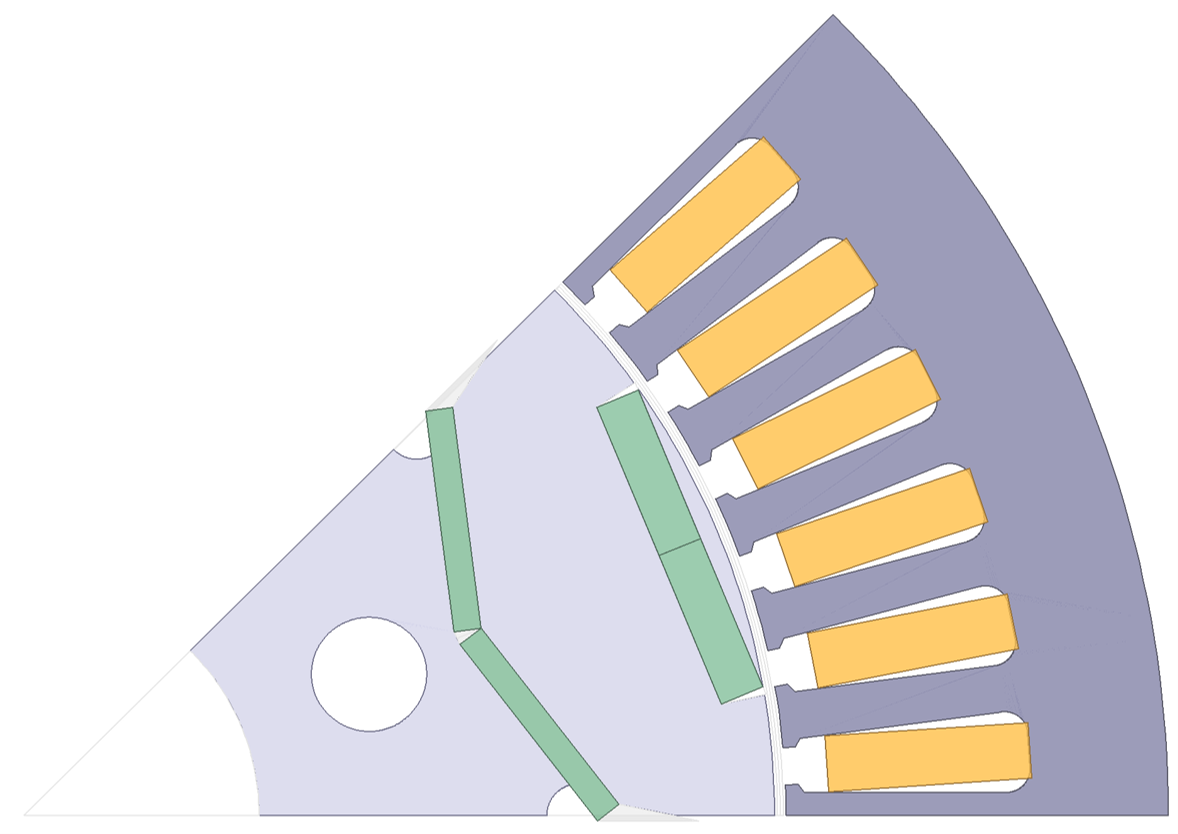}\\[0.2em]
  {\small (b)}
\end{minipage}

\vspace{0.6em}

\begin{minipage}{0.42\linewidth}\centering
  \includegraphics[width=\linewidth]{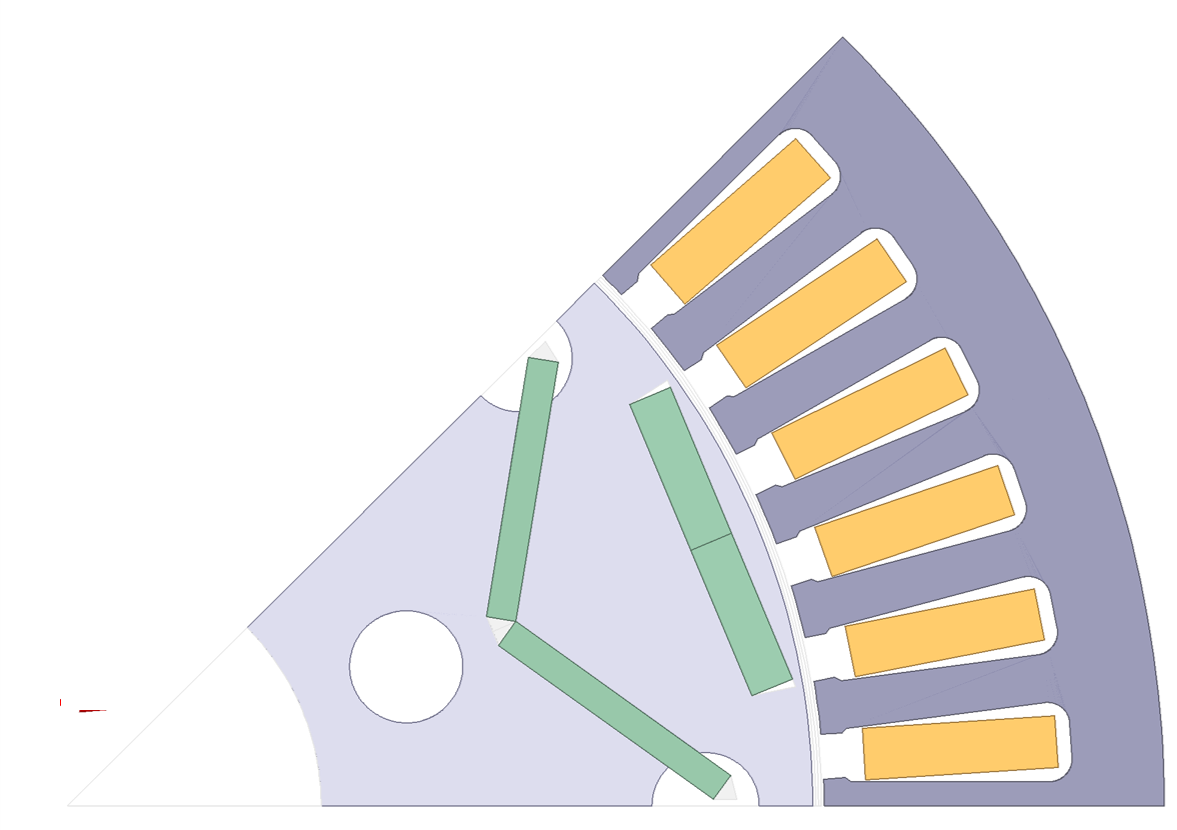}\\[0.2em]
  {\small (c)}
\end{minipage}\hfill
\begin{minipage}{0.42\linewidth}\centering
  \includegraphics[width=\linewidth]{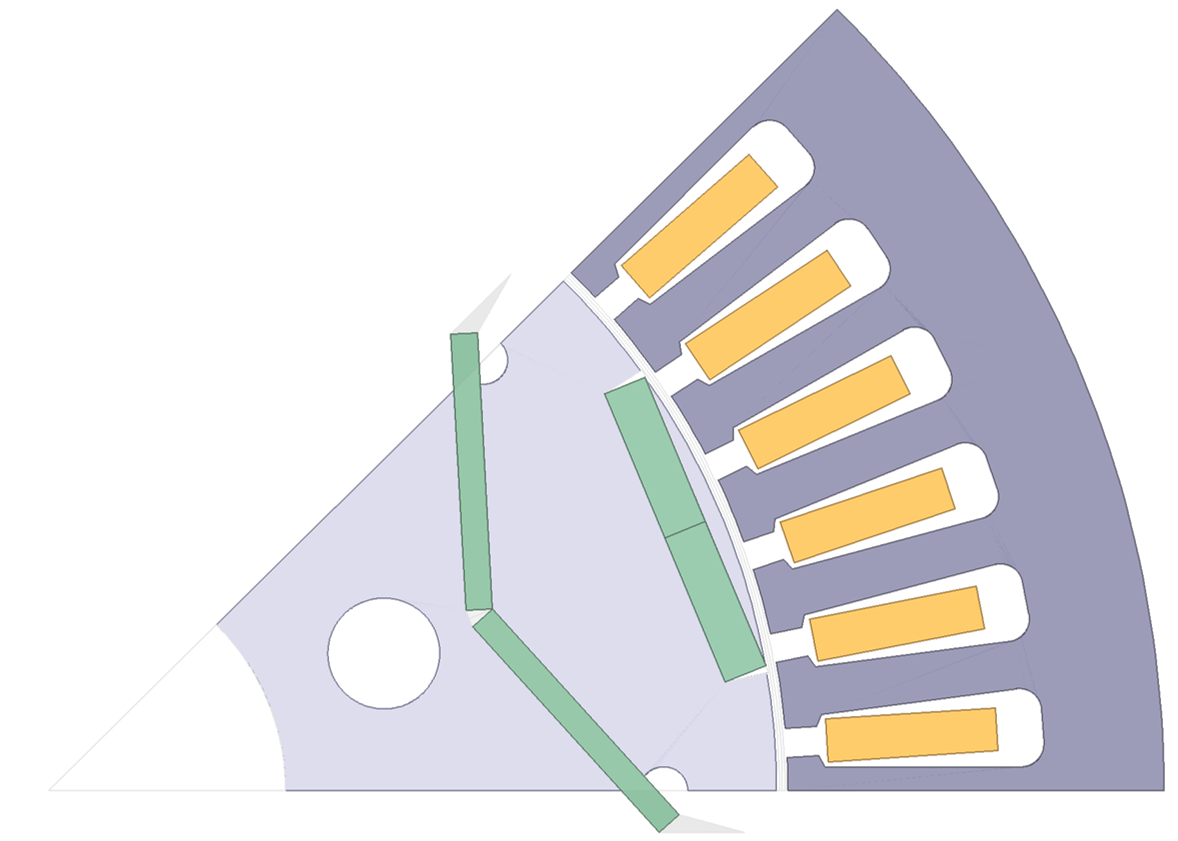}\\[0.2em]
  {\small (d)}
\end{minipage}
\caption{Examples of infeasible geometries from DOE sampling: component intersections (a)--(c) and protrusion outside the rotor boundary (d).}
\label{fig:fail_geometry_examples}
\end{figure}

The resampling process directly addresses the data-preparation gap of the manual design-workflow bottleneck identified in Section~1, and is implemented as an autonomous agentic feedback loop. Before each FEA batch, the \textit{Training agent} runs rule-based geometry validation on the DOE candidates, checking deterministic feasibility rules such as radial ordering, air-gap clearance, minimum bridge thickness, and PM/void/stator/rotor intersections. Candidates that pass are forwarded to the \textit{FEA agent} described in the next subsection, yet some still fail at the solver level through meshing errors or invalid component configurations; representative infeasible geometries are shown in Fig.~\ref{fig:fail_geometry_examples}. Every infeasible candidate, whether rejected by the validator or by the solver, is recorded in a geometry-validation log together with its design-variable vector, the violated rule or solver error, and the components involved. When the initial design space is wide or the sample count is large, these infeasible samples can dominate the batch, and the target training-set size for the surrogate cannot be secured.

To recover the missing samples, the framework closes a resampling feedback loop. An analysis of variance (ANOVA) over the infeasible and valid outcomes attributes the observed failure probability to individual variables and their interactions, and the LLM reasons over this statistical signal together with the recorded logs, aided by an explicit object-to-variable map that links each failing geometry object to the design variables controlling it, to identify fail-prone bounds and interactions while preserving safe regions for further sampling.

\begin{figure}[!htbp]
\centering
\includegraphics[width=\linewidth]{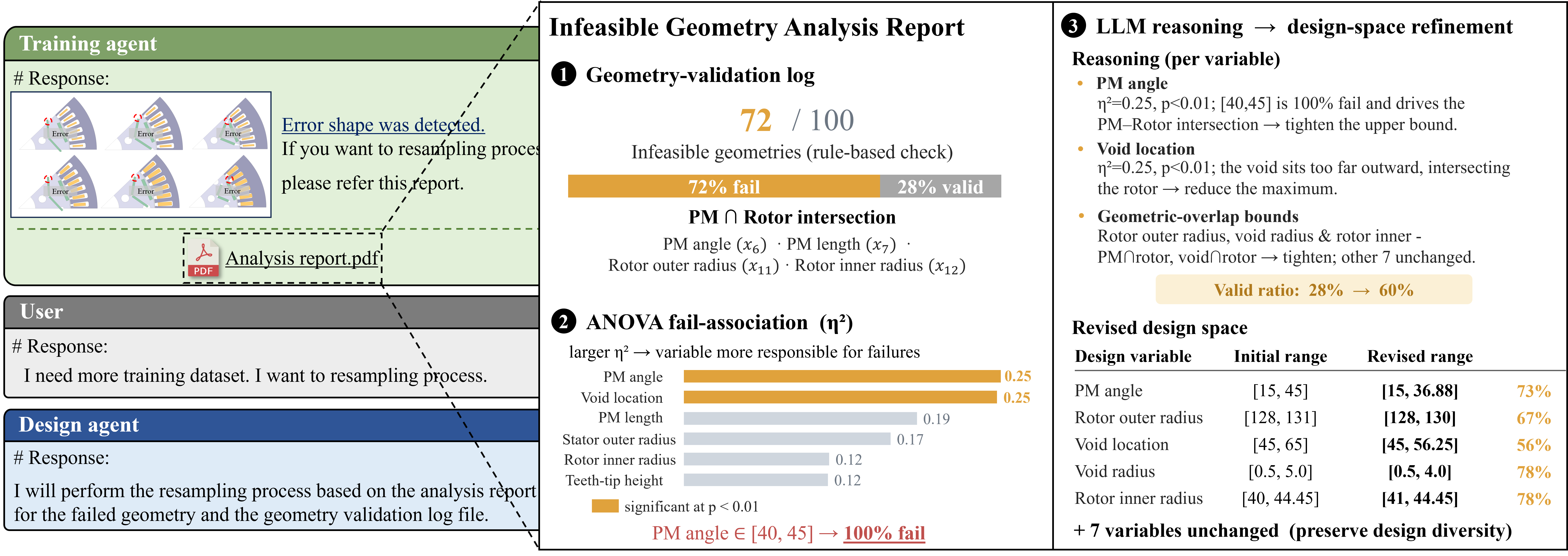}
\caption{Autonomous resampling loop for a worked example: agent-to-agent interaction between the \textit{Training agent} and the \textit{design sampling agent} (left) and the resulting analysis report with the geometry-validation log, ANOVA failure attribution ($\eta^2$), and revised variable ranges (right).}
\label{fig:resampling_interaction_process}
\end{figure}

Based on this reasoning, the \textit{Training agent} autonomously invokes the \textit{design sampling agent} of Step~1 (Section~2.1): redefining the design space is functionally the same operation as constructing the initial DOE, so the sub-agent is reused. Keeping the original optimization requirements and deterministic constraints, the \textit{design sampling agent} tightens the fail-prone bounds, determines how many additional points are required to recover the target valid count, and generates new DOE candidates for the next validation--FEA cycle. Each resampling step is compiled into an analysis report (right of Fig.~\ref{fig:resampling_interaction_process}) through which the user can inspect each revision and the evidence behind it, and the loop proceeds through agent-to-agent interaction (left of the same figure) until the requested number of analysis-feasible samples is secured or a predefined resampling limit is reached, after which the validated samples are forwarded to the surrogate-training stage.

\subsubsection*{FEA agent}
\begin{figure}[!htbp]
\centering
\includegraphics[width=0.8\linewidth]{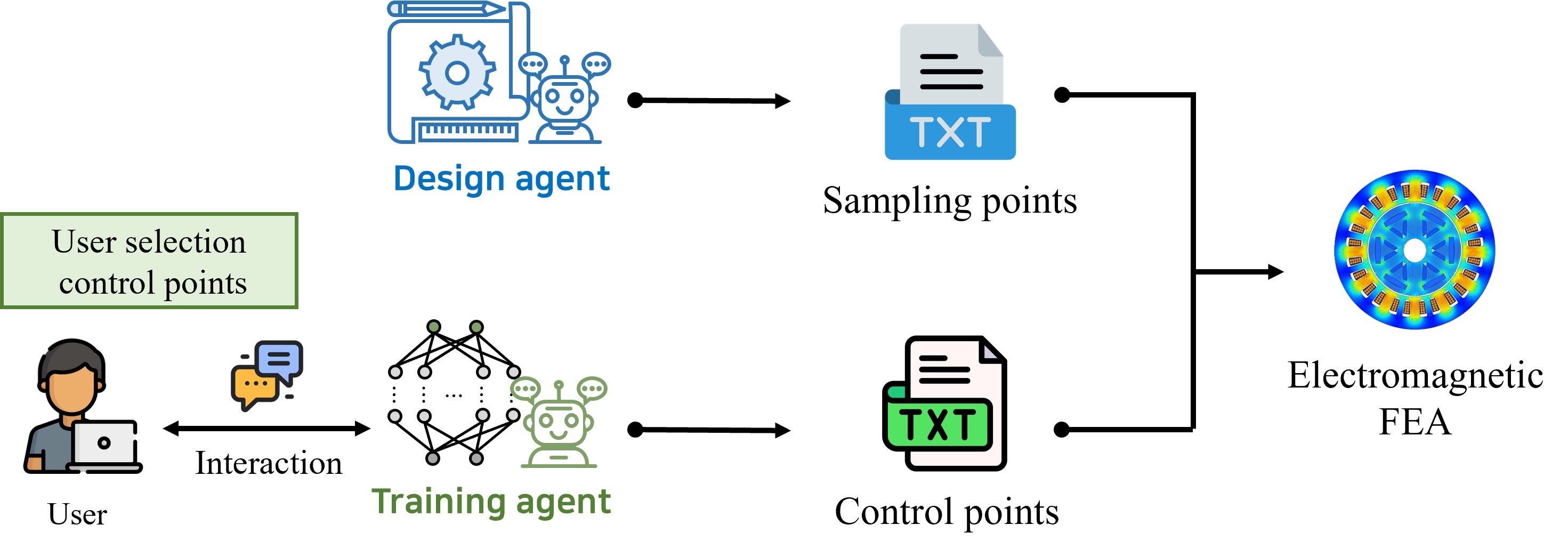}
\caption{Electromagnetic analysis stage of the \textit{Training agent}: analysis control-point setup and FEA execution over the DOE samples.}
\label{fig:electromagnetic_analysis_agent}
\end{figure}

The agent receives both the DOE sampling points generated in Step~1 and the additional points obtained through the resampling process, and performs high-fidelity electromagnetic analysis in Ansys Maxwell 2D on the reference IPMSM machine under the fixed analysis conditions summarized in Table~\ref{tab:machine_spec}. As illustrated in Fig.~\ref{fig:electromagnetic_analysis_agent}, the \textit{Training agent} first interacts with the user to determine analysis control points such as current, speed, and operating condition, then runs FEA for each sample and extracts core electromagnetic responses, including flux linkage and iron loss.

Average torque is computed from the extracted flux linkages and phase currents in the $dq$ frame, and iron loss from a modified Bertotti model, and the entire solver workflow, from project generation to result extraction, is automated through the \texttt{PyAEDT} library. The outputs are consolidated into a consistent simulation dataset containing the design variables, operating conditions, solver status, and the objective values defined in the optimization card, which directly feeds surrogate training.

\begin{table}[!htbp]
\centering
\caption{Specifications of the IPMSM case-study machine and the fixed analysis conditions (MTPA: maximum torque per ampere).}
\label{tab:machine_spec}
\begin{tabular}{ll}
\toprule
Item & Value \\
\midrule
Poles / stator slots & $8$ / $48$ (one-eighth periodic 2D model) \\
Air gap & $1$\,mm nominal (stator bore fixed; gap follows $x_{11}$) \\
Axial (stack) length & $153$\,mm \\
Permanent magnet & NdFeB, Arnold N30UH ($80\,^{\circ}\mathrm{C}$ BH data) \\
Core lamination & 30DH electrical steel ($k_h{=}71.72$, $k_c{=}0.251$, $k_e{=}12.16$) \\
Winding & three-phase, current-driven, $2$ parallel paths, coil pitch $5$ slots \\
Operating point & $I_{\mathrm{pk}}=140$\,A, $\beta=120^{\circ}$ (MTPA), $1000$\,rpm \\
Solver & Ansys Maxwell 2D transient (AEDT 2023.2), one electrical period \\
\bottomrule
\end{tabular}
\end{table}

\subsubsection*{Surrogate training agent}
The \textit{surrogate training agent} converts the valid electromagnetic dataset collected from the previous FEA stage into training data for objective-function prediction: each sample's geometry vector is aligned with its FEA-derived labels, and the dataset is cleaned and normalized.

\begin{figure}[!htbp]
\centering
\begin{minipage}[b]{0.56\linewidth}\centering
  \includegraphics[width=\linewidth]{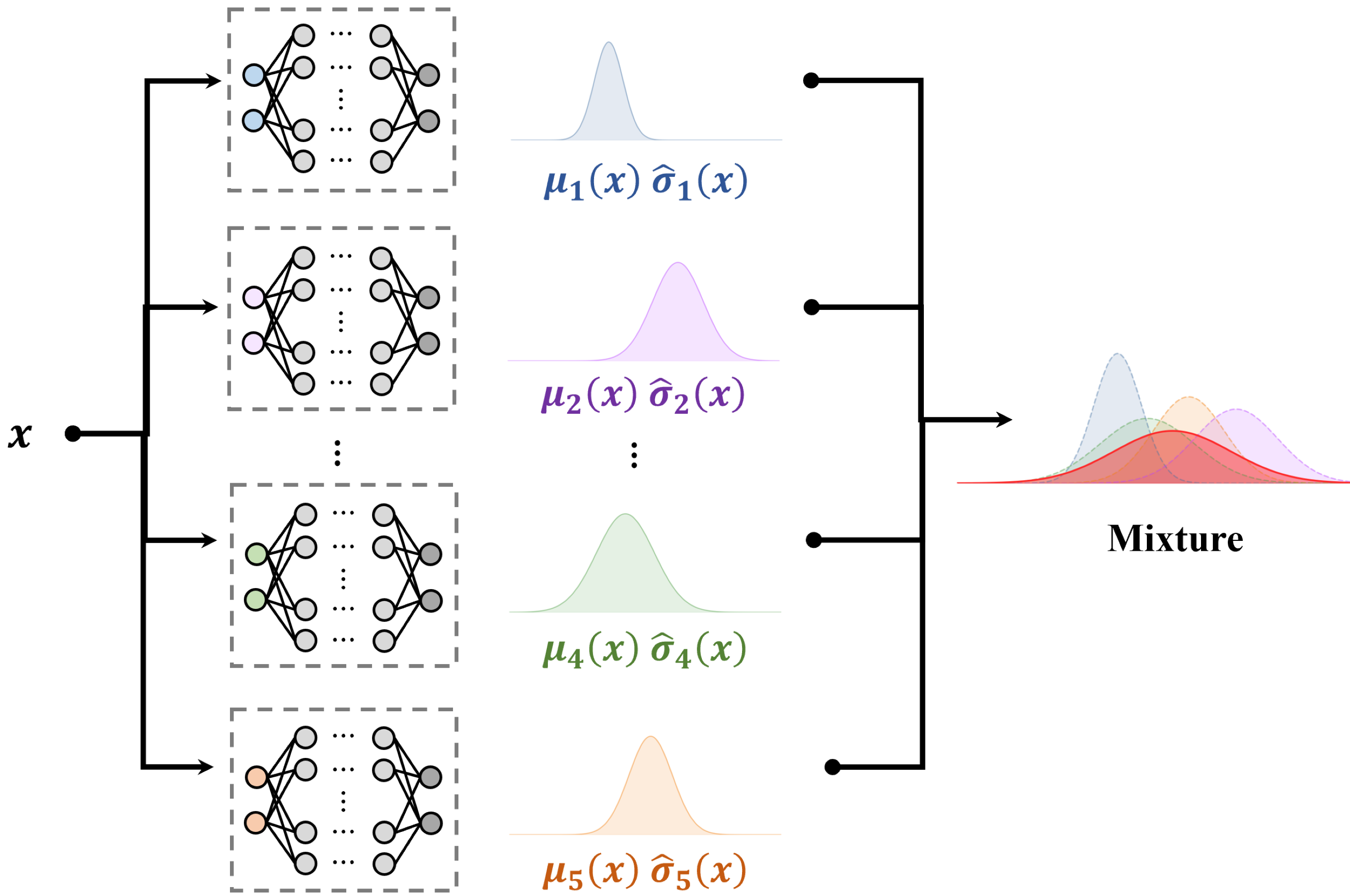}\\[0.2em]
  {\small (a)}
\end{minipage}\hfill
\begin{minipage}[b]{0.39\linewidth}\centering
  \includegraphics[width=\linewidth]{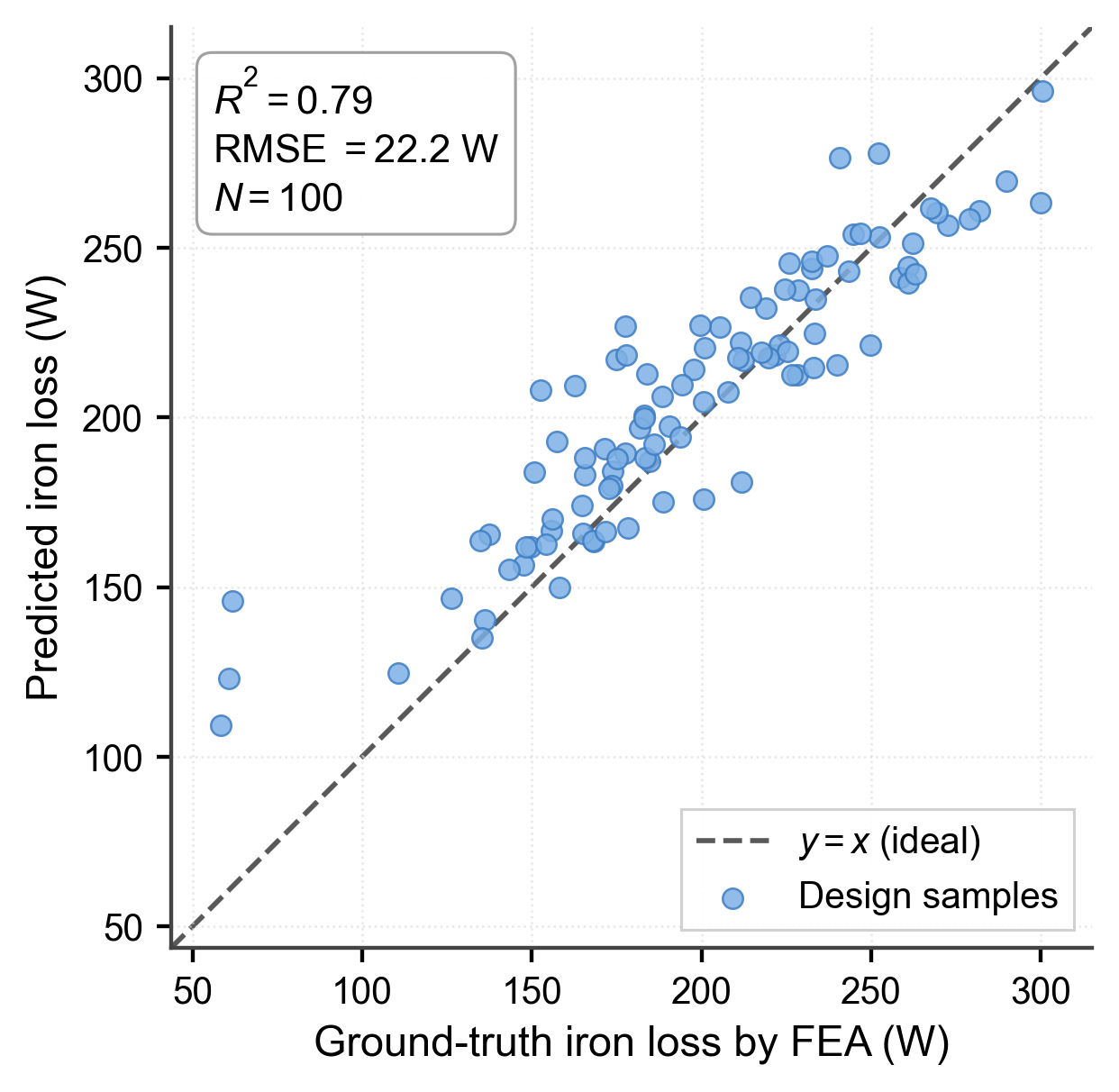}\\[0.2em]
  {\small (b)}
\end{minipage}
\caption{Deep-ensemble model trained by the \textit{surrogate training agent}: (a)~the deep-ensemble architecture and (b)~predicted values of the trained model versus the FEA ground truth.}
\label{fig:deep_ensemble_architecture}
\end{figure}

Figure~\ref{fig:deep_ensemble_architecture}(a) presents the default surrogate, a deep ensemble of five multilayer perceptron (MLP) regressors trained on this dataset~\cite{lakshminarayanan2017deep,yang2024deepensemble}. Each member maps the 12-dimensional design vector to a predicted objective value and a predictive variance, and Fig.~\ref{fig:deep_ensemble_architecture}(b) shows an example of the resulting prediction accuracy for the iron-loss objective, comparing the ensemble-mean prediction against the FEA ground truth. The layer and node settings can be adjusted through user--agent interaction when the optimization problem is reconfigured.

Following the standard uncertainty interpretation of deep ensembles, the predictive variance is decomposed into aleatoric and epistemic components~\cite{lakshminarayanan2017deep,kendall2017uncertainties}. For $M=5$ members with per-member outputs $\mu_m(x)$ and $\sigma_m^{2}(x)$, the ensemble mean and total predictive variance are
\begin{equation}
\mu(x)=\frac{1}{M}\sum_{m=1}^{M}\mu_m(x),\qquad
\sigma^{2}(x)=
\underbrace{\frac{1}{M}\sum_{m=1}^{M}\sigma_m^{2}(x)}_{\text{aleatoric}}
+
\underbrace{\frac{1}{M}\sum_{m=1}^{M}\left(\mu_m(x)-\mu(x)\right)^{2}}_{\text{epistemic}} .
\label{eq:ensemble_uncertainty_decomposition}
\end{equation}
The aleatoric term captures data- and label-related noise, whereas the epistemic term captures model uncertainty and grows in sparse or OOD regions. The total predictive standard deviation $\sigma(x)=\sqrt{\sigma^{2}(x)}$ therefore serves as the reliability indicator against the surrogate-reliability bottleneck of Section~1: high-uncertainty candidates are treated as low-confidence predictions and selectively corrected by FEA in the FEA--AI hybrid optimization loop.

To train the heteroscedastic outputs stably, each MLP is optimized with the $\beta$-NLL loss rather than the standard Gaussian negative log-likelihood (NLL), whose variance-scaled mean gradient under-weights high-variance samples; $\beta$-NLL counteracts this by reweighting each sample's NLL term with a stop-gradient factor of its predicted variance~\cite{seitzer2022pitfalls}:
\begin{equation}
\begin{aligned}
\mathcal{L}_{\beta\text{-NLL}} &= \frac{1}{N}\sum_{i=1}^{N}\left[\operatorname{sg}\!\left(\hat{\sigma}^{2}(x_i)\right)\right]^{\beta}\times \left[\frac{1}{2}\log \hat{\sigma}^{2}(x_i)+\frac{\left(y_i-\hat{\mu}(x_i)\right)^2}{2\hat{\sigma}^{2}(x_i)}+\frac{1}{2}\log(2\pi)\right],
\end{aligned}
\end{equation}
where $\hat{\mu}(x_i)$ and $\hat{\sigma}^{2}(x_i)$ are the network outputs for sample $i$, $N$ is the batch size, and $\operatorname{sg}(\cdot)$ denotes stop-gradient; the exponent $\beta$ controls the reweighting strength and is set to $0.5$ following the same study. Early stopping on the validation $\beta$-NLL prevents overfitting and unstable uncertainty estimates.

\subsection{Optimization agent (FEA--AI hybrid model-based optimization)}
\begin{figure}[!htbp]
\centering
\includegraphics[width=0.8\linewidth]{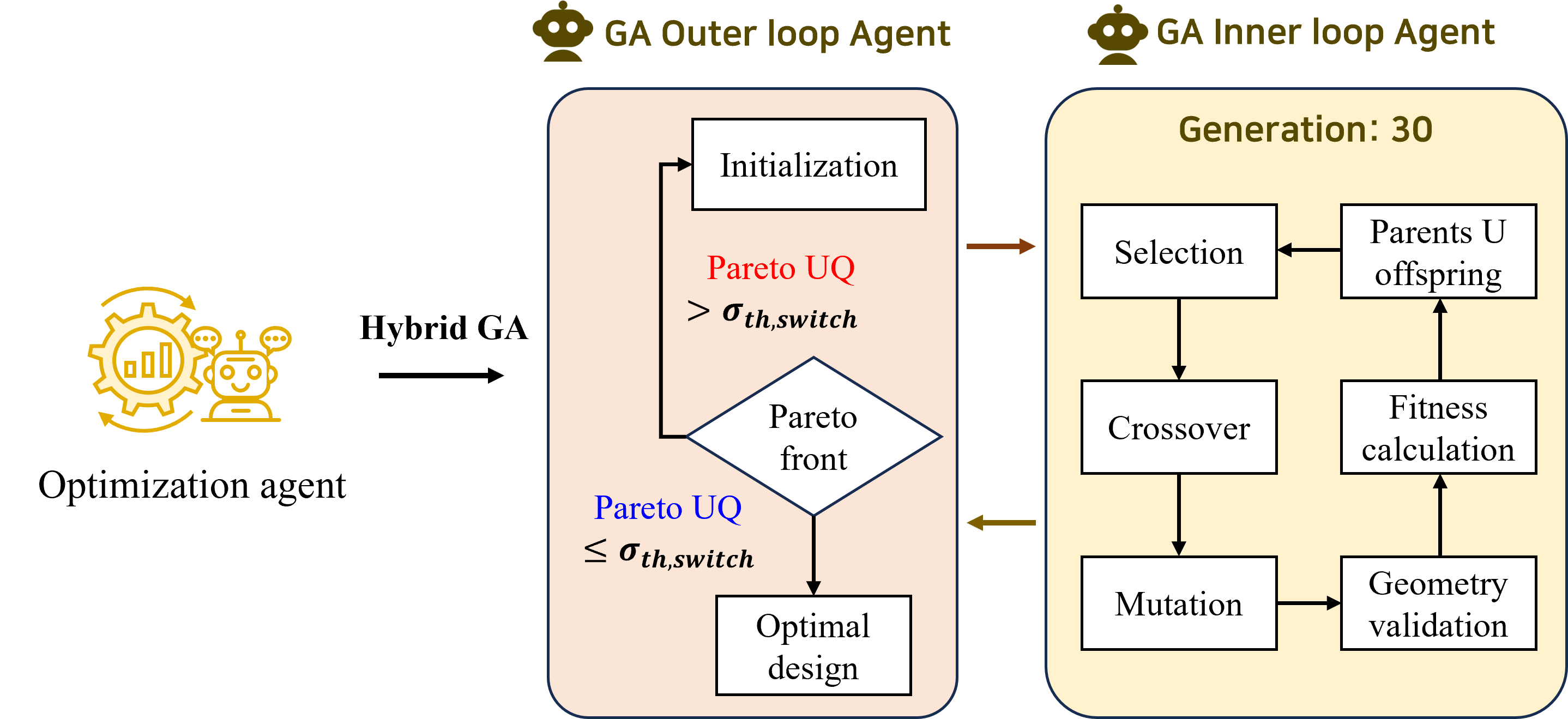}
\caption{Inner/outer-loop execution strategy of the \textit{Optimization agent}: the GA \textit{inner-loop agent} runs generation-level search with surrogate-only fitness evaluation, while the GA \textit{outer-loop agent} checks the returned candidate front with the uncertainty-gated FEA--AI hybrid model and feeds the verified results into the next round.}
\label{fig:optimization_agent_ga_process}
\end{figure}

Using the optimization card from Step~1 and the trained deep-ensemble surrogate from Step~2, the \textit{Optimization agent} conducts evolutionary design search through two coordinated sub-agents: a GA \textit{inner-loop agent} and a GA \textit{outer-loop agent}. The GA \textit{inner-loop agent} performs fast generation-level search using only the AI surrogate for routine fitness evaluation, whereas the GA \textit{outer-loop agent} monitors optimization-level uncertainty and invokes the FEA--AI hybrid model when reliability-critical samples must be checked. This separation retains the exploration speed of surrogate-assisted evolutionary search while adding targeted FEA reliability checks. Figure~\ref{fig:optimization_agent_ga_process} illustrates this inner/outer-loop execution strategy.

\subsubsection*{GA inner-loop agent}
The GA \textit{inner-loop agent} manages the generation-wise evolutionary search. Starting from a validated population, each generation applies evolutionary operators to produce offspring design-variable sets. Because these operators can generate infeasible motor geometries, each offspring is screened by the rule-based geometry validation of Section~2.2 before fitness evaluation, and rejected candidates are regenerated until the population size is restored.

Fitness is then evaluated with the AI surrogate model alone; no FEA is called inside this routine inner loop. Parents and offspring are merged and ranked for survivor selection, and the top-ranked designs, hereafter the \emph{candidate front}, are passed to the GA \textit{outer-loop agent} for reliability checking and reuse in the next optimization round.

\subsubsection*{GA outer-loop agent}
The GA \textit{outer-loop agent} controls the optimization process across multiple rounds. Each round starts from an initial population generated by Latin hypercube sampling (LHS)~\cite{mckay1979lhs} and screened by the same geometry validation as the inner loop; from the second round onward, the candidate front of the previous round is inserted into this population, preserving promising designs while newly sampled candidates keep exploring the design space.

\begin{figure}[!htbp]
\centering
\includegraphics[width=0.8\linewidth]{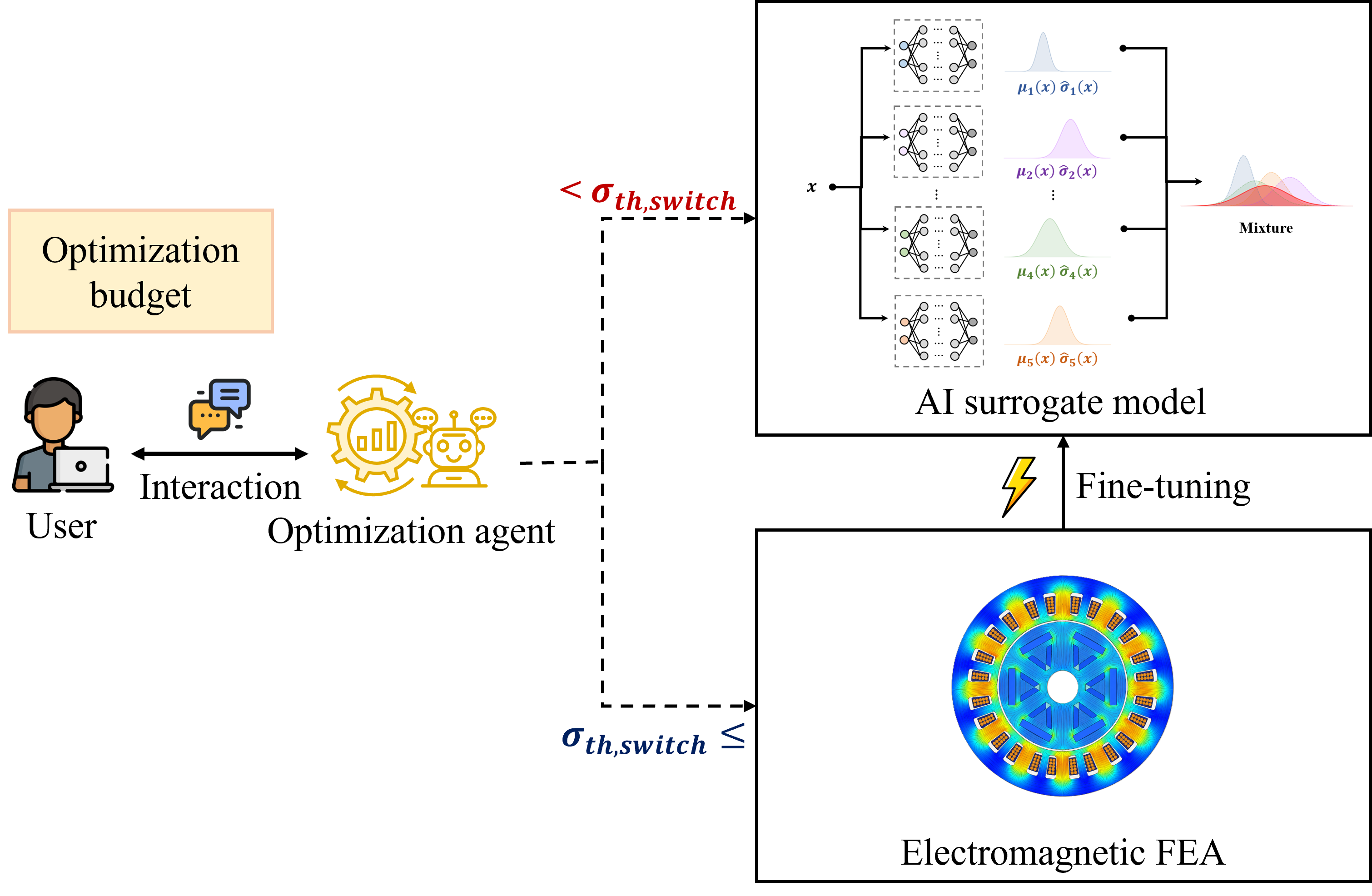}
\caption{FEA--AI hybrid model of the GA \textit{outer-loop agent}: each evaluation switches between AI-surrogate inference and high-fidelity FEA according to the uncertainty-based switching threshold.}
\label{fig:fea_ai_hybrid_uq_switch}
\end{figure}

Unlike the inner loop, the GA \textit{outer-loop agent} evaluates the reliability-critical samples, namely the initial population and the candidate front, with the FEA--AI hybrid model shown in Fig.~\ref{fig:fea_ai_hybrid_uq_switch}; candidates carried over from an earlier round keep their stored FEA labels. For each unverified candidate $x$, the surrogate predicts the objective $\mu(x)$ and the predictive standard deviation $\sigma(x)$ from Eq.~\ref{eq:ensemble_uncertainty_decomposition}, and the agent switches the evaluator according to
\begin{itemize}
  \item $\sigma(x) < \sigma_{\mathrm{th,switch}}$: use AI-surrogate prediction for fitness evaluation,
  \item $\sigma(x) \ge \sigma_{\mathrm{th,switch}}$: run high-fidelity FEA, use the FEA result as the label, and append the sample to the training dataset.
\end{itemize}
Because routine fitness is already served by the surrogate in the inner loop, high-fidelity FEA is confined to the high-uncertainty and front candidates rather than to every individual of every generation, which is how the FEA--AI hybrid model addresses the FEA-cost and surrogate-reliability bottlenecks of Section~1.

The switching threshold $\sigma_{\mathrm{th,switch}}$ is not fixed in advance: it is either held at a constant value (rule-based switching) or adaptively updated at each round by an LLM controller from the surrogate's live calibration (agent-based switching); both variants are studied in Section~4, and the threshold can also be adjusted through user--agent interaction.

The outer loop terminates on the reliability of the candidate front itself: its mean uncertainty $\bar{\sigma}_{\mathrm{front}}$ is compared with the same threshold $\sigma_{\mathrm{th,switch}}$, and the search proceeds to further rounds until $\bar{\sigma}_{\mathrm{front}} \le \sigma_{\mathrm{th,switch}}$, at which point the final optimal design is selected from the current candidate front. Because any front candidate whose own $\sigma(x)$ exceeds the threshold has already been relabeled by FEA in the same round, the retained front is reliable both on average and design-by-design.
After each round, the surrogate is fine-tuned on the FEA-augmented training set (round-wise active learning)~\cite{lakshminarayanan2017deep,gong2025comparative}. For multi-objective problems, NSGA-II replaces the standard GA and the candidate front becomes the Pareto front of non-dominated designs~\cite{deb2002nsga2}; the switching and termination rules apply unchanged.

\section{Preliminary studies: effects of RAG and resampling}

Before evaluating the full optimization pipeline, we first validate its two enabling components. Section~3.1 tests the RAG-grounded \textit{Design agent}: because the agent couples a local LLM with motor-textbook retrieval for domain specialization, we verify that this retrieval actually improves accuracy on motor-domain questions and that the local GPT-oss 20B backbone is competitive enough with other models to serve as the framework backbone. Section~3.2 tests the autonomous resampling loop performed by the \textit{Training agent}: we verify that the agent reasons correctly over the infeasible-geometry data and, through the resulting resampling, recovers the requested number of sampling points.

\subsection{RAG study}

The RAG study compares the same backbone with and without motor-textbook retrieval, with everything else held fixed, and assesses the grounding effect in two complementary ways: qualitatively, through representative answer comparisons, and quantitatively, through answer accuracy on a domain-specialized question set.

\subsubsection*{Qualitative analysis}
Table~\ref{tab:rag_examples} illustrates this grounding effect at the level of individual questions, on six representative items drawn from the domain-specialized set used for the quantitative evaluation below, spanning its three categories. In every case the same GPT-oss 20B backbone answers incorrectly without retrieval and returns the textbook answer with it. The effect is most pronounced for the book-specific worked-example quantities (e.g., the reported back-EMF $E_f=241.2$\,V and the peak air-gap flux density $B_{mg}=0.923$\,T): these values cannot be recovered from parametric knowledge, and without retrieval the model generates plausible but incorrect values ($10.5$ and $0.50$), precisely the type of hallucinated guidance that would otherwise contaminate the optimization card. Retrieval likewise corrects erroneous multi-step computations (the winding factor and the cogging-torque index), where the retrieved passages supply the correct textbook method rather than a memorized value, and definitional errors (the cogging-torque-index definition and the leakage-reactance components), where the ungrounded model substitutes a generic but incorrect recollection for the textbook definition.

\begin{table}[!htbp]
\centering
\caption{Representative questions from the domain-specialized set, answered by GPT-oss 20B without (No-RAG) and with textbook retrieval (RAG)~\cite{gieras2009permanent}. For every question, the RAG answers are correct, whereas the No-RAG answers are all incorrect on these motor-domain problems.}
\label{tab:rag_examples}
\footnotesize
\begin{tabular}{p{1.6cm}p{6.4cm}p{1.9cm}p{2.0cm}p{2.3cm}}
\toprule
Category & Question & Textbook & No-RAG & RAG \\
\midrule
Numerical & A 3-phase, 4-pole, 36-slot full-pitch winding. Compute the fundamental winding factor $k_{w1}$. & $0.96$ & $\times$~$0.18$ & $\checkmark$~$0.96$ \\
\addlinespace
Numerical & An IPMSM has $S=12$ slots and $2p=10$ poles. Using the textbook definition, compute the fundamental cogging-torque index $n_{\mathrm{cog}}$. & $6$ & $\times$~$30$ & $\checkmark$~$6$ \\
\addlinespace
Book-specific & A worked example analyzes a three-phase salient-pole PM synchronous motor with $s_1=72$ slots, $2p=6$ poles and a $10/12$ coil pitch. What rms back-EMF $E_f$ (in volts), excited by the rotor flux, does the textbook obtain? & $241.2$\,V & $\times$~$10.5$ & $\checkmark$~$241.2$\,V \\
\addlinespace
Book-specific & In the textbook's $2p=8$ PM brushless example with stator bore $D_{\mathrm{1in}}=0.132$\,m and $L_i=0.153$\,m, what peak air-gap flux density $B_{mg}$ (in T) is used? & $0.923$\,T & $\times$~$0.50$ & $\checkmark$~$0.923$\,T \\
\addlinespace
Conceptual & How does the textbook define the fundamental cogging-torque index $n_{\mathrm{cog}}$ in terms of the slot number $s_1$ and pole number $2p$? & $\dfrac{\mathrm{LCM}(s_1,2p)}{2p}$ & $\times$~$\dfrac{s_1}{2p}$ & $\checkmark$~$\dfrac{\mathrm{LCM}(s_1,2p)}{2p}$ \\
\addlinespace
Conceptual & What components make up the stator (armature) leakage reactance in the textbook? & slot, differential, and tooth-top leakage & $\times$~winding and core leakage & $\checkmark$~slot, differential, and tooth-top leakage \\
\bottomrule
\end{tabular}
\end{table}

The same pattern holds for full explanations, not only short answers: for an IPMSM saliency question ($L_q>L_d$ and reluctance torque), the No-RAG response is fluent but relies on a physically reversed mechanism, a failure mode that is difficult for non-expert users to detect, whereas the RAG response provides the correct magnetic-circuit reasoning; the full response pair is reproduced in Appendix~\ref{app:rag_response} (Fig.~\ref{fig:rag_saliency_comparison}).

\subsubsection*{Quantitative analysis}
\label{sec:rag_quant}%
To quantify this effect across models, we evaluate four backbones, two local open models (GPT-oss 20B, Llama-3 8B) and two commercial APIs (GPT-5-mini, Gemini-2.5-flash), with and without textbook retrieval under identical decoding, on $90$ domain-specialized questions derived from the motor reference~\cite{gieras2009permanent}: $30$ \emph{numerical} problems recomputed from textbook formulas, $30$ \emph{book-specific} lookups of values that appear only in particular worked examples, and $30$ \emph{conceptual} questions on textbook definitions. Numerical and book-specific answers are graded within a per-item tolerance ($\pm1$--$6\%$), conceptual answers by an LLM judge applied identically to both settings, and all four backbones share the \textit{Design agent}'s hybrid retriever ($k{=}16$).

\begin{figure}[!htbp]
\centering
\includegraphics[width=\linewidth]{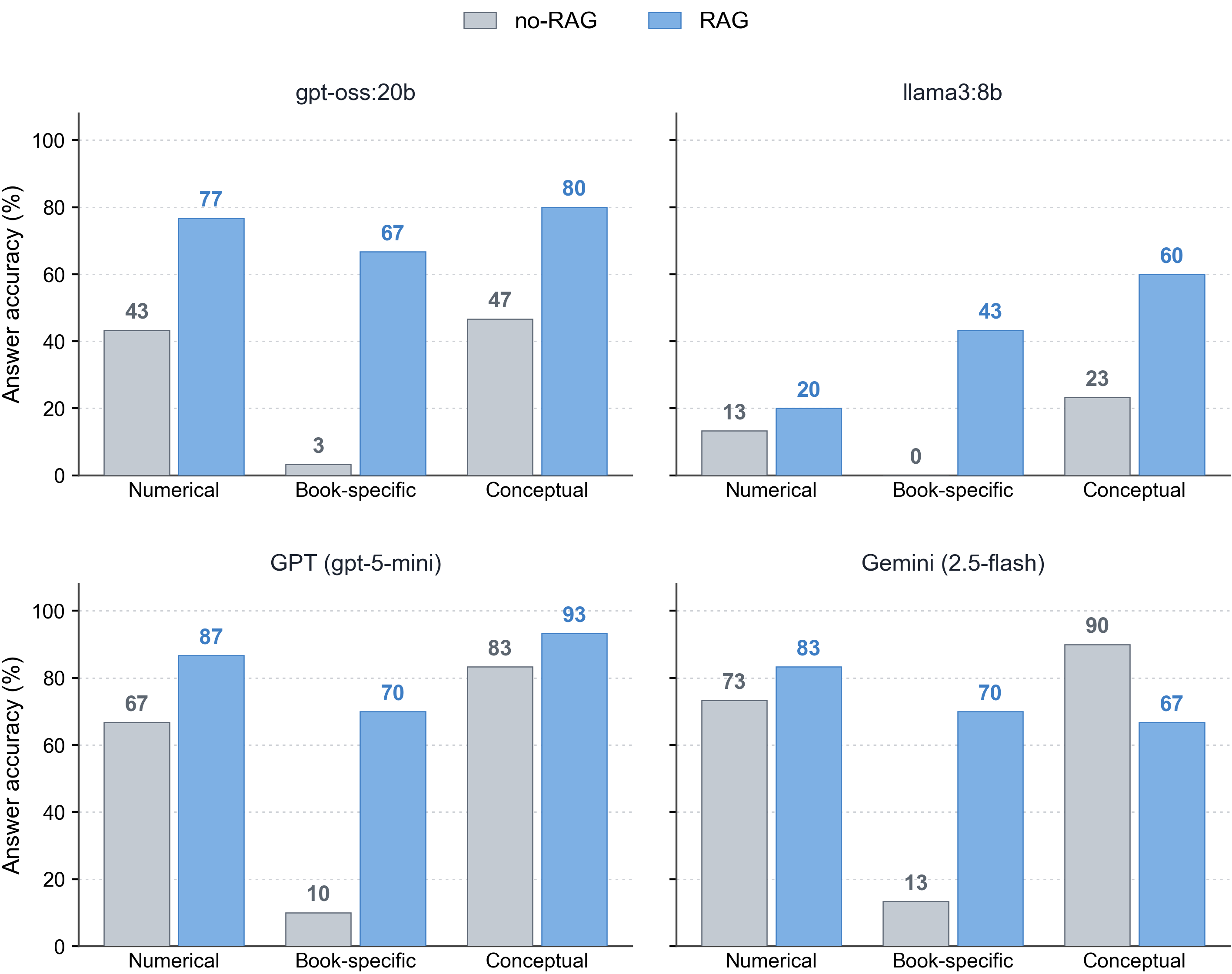}
\caption{Answer accuracy of each LLM backbone on the $90$ domain-specialized questions, which comprise three types: numerical, book-specific, and conceptual. Gray bars denote the No-RAG results and blue bars the RAG results. Most backbones achieve higher answer accuracy on the motor-domain questions when RAG is used.}
\label{fig:rag_quant_bar}
\end{figure}

Figure~\ref{fig:rag_quant_bar} summarizes the results. The most consistent effect appears on the \emph{book-specific} lookups, where retrieval raises every backbone from near zero ($0$--$13\%$) to $43$--$70\%$: even the commercial APIs cannot recover these values from parametric knowledge alone, and retrieval is what supplies them. For GPT-oss 20B, RAG also raises numerical accuracy from $43\%$ to $77\%$ and conceptual accuracy from $47\%$ to $80\%$; the smaller Llama-3 8B gains little on numerical problems because the multi-step arithmetic exceeds its capability, while the commercial backbones start high without RAG and still gain on numerical accuracy.

One exception is the conceptual score of Gemini-2.5-flash, which decreases under retrieval: instructed to ground its answer in the retrieved passage, the model, which already possesses the relevant knowledge, confines its answer to a narrowly focused retrieved passage and omits key points that it would otherwise include. Because the retriever is tuned for the dominant book-specific lookups, we retain it and report this as a documented trade-off.

Because retrieval improves every backbone in aggregate, the choice of backbone can rest on deployment considerations rather than accuracy. Although the most recent commercial model, GPT-5-mini, attains the highest accuracy with retrieval, the local GPT-oss 20B with RAG follows closely and nearly matches the commercial models on the hardest book-specific lookups ($67\%$ versus $70\%$). Because this small margin does not justify a commercial dependency, and because the locally deployed GPT-oss 20B can serve as the task-specific agents specialized for each stage of the motor design optimization workflow while keeping all prompts and proprietary design data on-premises, it is adopted as the backbone of the proposed framework, with hybrid-retrieval RAG as its default problem-definition interface.

\subsection{Resampling study}

This section demonstrates the log-informed resampling loop of the \textit{Training agent} (Section~2.2) on a representative user-defined scenario and reports how it secures the requested number of analysis-feasible samples from a low-yield initial design space.

\begin{figure}[!htbp]
\centering
\begin{minipage}[b]{0.56\linewidth}\centering
  \includegraphics[width=\linewidth]{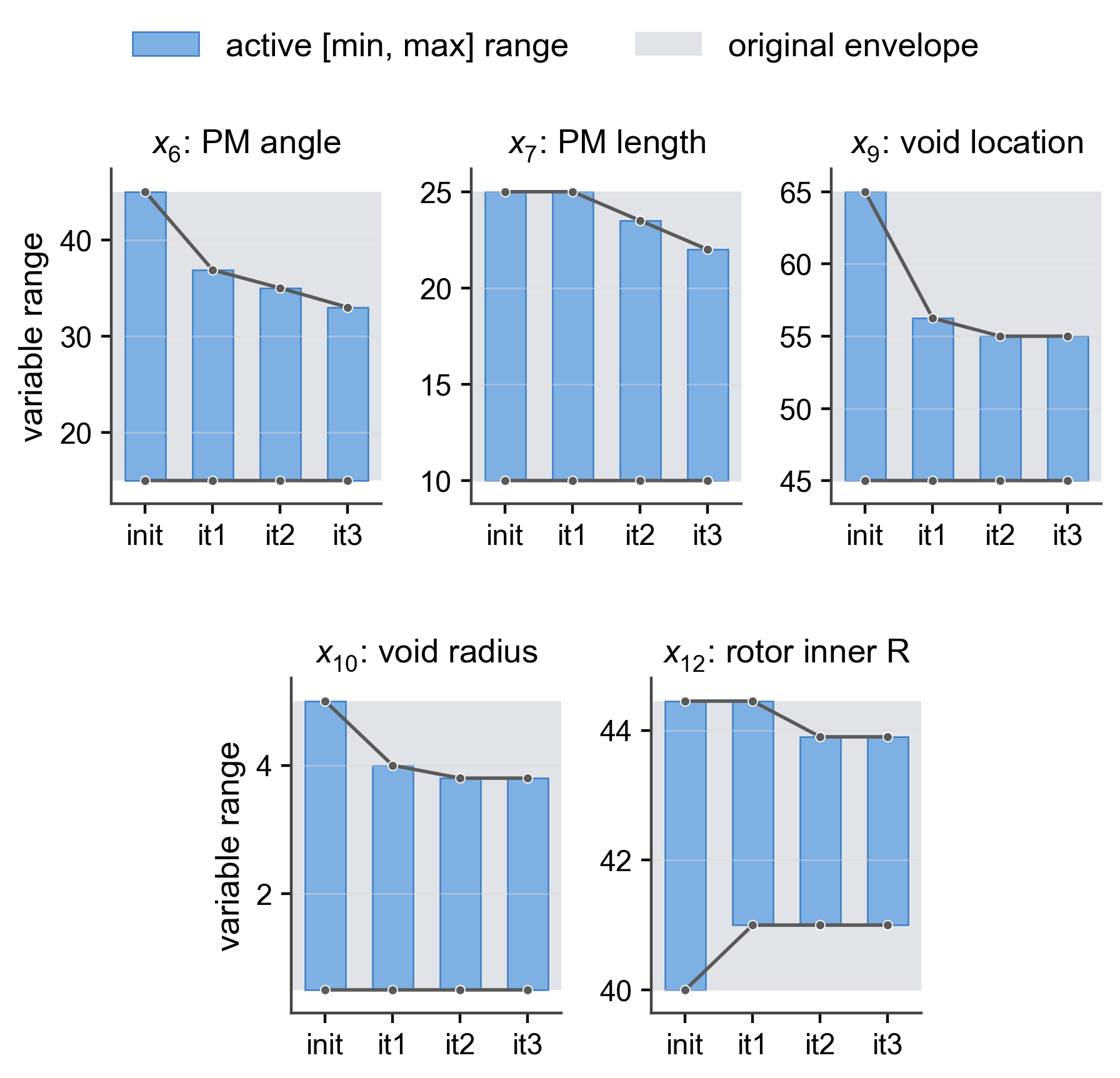}\\[0.2em]
  {\small (a)}
\end{minipage}\hfill
\begin{minipage}[b]{0.38\linewidth}\centering
  \includegraphics[width=\linewidth]{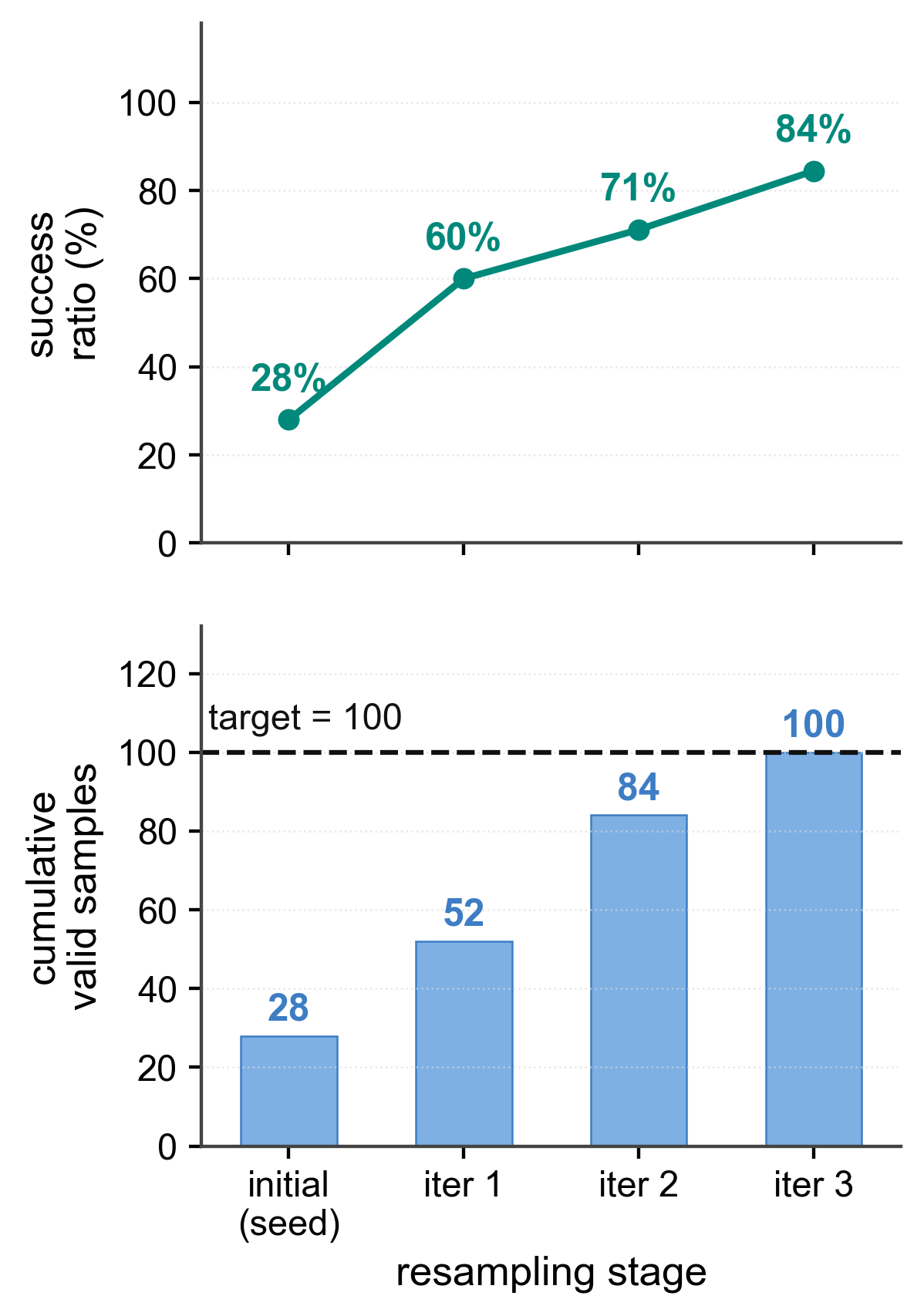}\\[0.2em]
  {\small (b)}
\end{minipage}
\caption{Agent-based autonomous resampling loop. (a)~The agent refines the design space through reasoning at every resampling iteration. (b)~The success ratio of the sampling points generated over the refined ranges rises accordingly, reaching the target within three iterations.}
\label{fig:resampling_analysis}
\end{figure}

Through interaction with the \textit{Design agent}, the user specifies a single-objective iron-loss minimization problem, selects all twelve geometry parameters $x_1$--$x_{12}$ of Fig.~\ref{fig:parameterized_ipmsm_doe}(a) as design variables, and requests $100$ analysis-feasible samples for surrogate training. Because the user provides no reliable per-variable ranges, the \textit{design sampling agent} initializes provisional default bounds and generates an initial Latin hypercube DOE of $100$ candidates over this intentionally wide design space.

When the rule-based geometry validation of Step~2 is applied to the initial DOE, only $28\%$ of the $100$ candidates prove analysis-feasible, far short of the requested count. From the geometry-validation log of the infeasible candidates, the ANOVA attribution identifies five fail-prone variables: $x_6$ (PM angle), $x_7$ (PM length), $x_9$ (void location), $x_{10}$ (void radius), and $x_{12}$ (rotor inner radius). These variables are precisely those controlling the geometry objects recorded as invalid or mutually intersecting in the log, confirming that the statistical attribution is consistent with the recorded failure mechanisms. A small, variable-independent fraction of the failures reflects solver-level meshing issues and cannot be removed by design-space refinement.

Guided by the object-to-variable map, the \textit{Training agent} then autonomously invokes the \textit{design sampling agent} to redefine the design space, narrowing the bounds of the fail-prone variables while retaining the original ranges of the low-influence variables to preserve sampling diversity. Figure~\ref{fig:resampling_analysis}(a) shows the result: across iterations, the active ranges of the five fail-prone variables are progressively tightened toward the feasible region while staying within the original envelope. The per-iteration decisions need not follow the $\eta^2$ ranking exactly: in the first refinement the agent also trimmed the rotor outer radius $x_{11}$, whose marginal $\eta^2$ is negligible but which the object map ties to the failing rotor geometry (Fig.~\ref{fig:resampling_interaction_process}), while the PM length $x_7$ was left unchanged until the second iteration.

Figure~\ref{fig:resampling_analysis}(b) shows the resulting convergence: the geometry success ratio rises monotonically from $28\%$ to $60\%$, $71\%$, and $84\%$ over three successive refinements, while the cumulative valid-sample count increases accordingly. The loop terminates automatically at the third iteration, once the requested $100$ valid points are secured, so the dataset is neither under- nor over-sampled. The complete demonstration, including every LLM reasoning call, runs in about $420$\,s on the local GPT-oss 20B backbone.

\section{FEA--AI hybrid model setup and optimization}

With the problem-definition interface established in Section~3.1, and with the $100$ analysis-feasible samples secured by the log-informed resampling loop of Section~3.2 serving as the surrogate training dataset, we now evaluate the proposed optimization pipeline.

\begin{table}[!htbp]
\centering
\caption{FEA--AI hybrid model and GA settings for the IPMSM optimization experiment.}
\label{tab:hybrid_ga_setting}
\begin{tabular}{lll}
\toprule
\textbf{Category} & \textbf{Item} & \textbf{Setting} \\
\midrule
\multirow{5}{*}{\textbf{AI architecture}}
 & Type & MLP deep ensemble \\
 & Hidden layers / nodes & $2$ / $64$ ($12{\to}64{\to}64{\to}2$, ReLU) \\
 & Ensemble size $M$ & $5$ \\
 & Variance head & Softplus $+\,10^{-6}$ \\
 & Parameters & $5{,}122$ \\
\midrule
\multirow{5}{*}{\textbf{Trained model}}
 & Train dataset & $100$ \\
 & Loss & $\beta$-NLL \\
 & Optimizer & Adam, $10^{-3}$, batch $32$ \\
 & Max epoch / early stopping & $300$ / patience $50$, best-NLL checkpoint \\
 & Normalization & Min--Max \\
\midrule
\multirow{2}{*}{\textbf{Uncertainty}}
 & UQ & aleatoric (mean) $+$ epistemic (var) \\
 & Switching metric & $\sigma_{\mathrm{total}}(x)$ in W, Eq.~\ref{eq:ensemble_uncertainty_decomposition} \\
\midrule
\multirow{3}{*}{\textbf{Active learning}}
 & Trigger & each round, $30$ epochs \\
 & Fine-tuning & full fine-tuning \\
 & Loss / Optimizer & $\beta$-NLL / Adam, $10^{-3}$, batch $32$ \\
\midrule
\multirow{3}{*}{\textbf{GA}}
 & Population / Generations & $25$ / $30$ \\
 & Crossover / Mutation & SBX ($p{=}0.9$, $\eta{=}15$) / Poly ($p{=}0.2$, $\eta{=}20$) \\
 & Selection / Replacement & Tournament ($k{=}3$) / Elitist truncation \\
\midrule
\textbf{Overall} & Random seeds & $5,\,42,\,777,\,2026$ \\
\bottomrule
\end{tabular}
\end{table}

The experiment is configured as a single-objective IPMSM optimization that minimizes iron loss, starting from the same initial geometry. The design-variable set, variable bounds, geometry-validation rules, and infeasible-geometry handling logic are fixed throughout the experiment so that the observed performance difference comes from the optimization execution strategy rather than from a changed design space. All experiments are executed in a local environment based on an AMD Ryzen 9 7900 12-Core CPU; the case-study machine and the fixed analysis conditions are those summarized in Table~\ref{tab:machine_spec} (Section~2.2).

Table~\ref{tab:hybrid_ga_setting} summarizes the FEA--AI hybrid model and GA settings used throughout the IPMSM case study. The surrogate is the deep ensemble of Section~2.2 ($M{=}5$, $\beta$-NLL loss), trained here on the $100$ analysis-feasible samples from the Section~3.2 resampling loop with min--max normalized inputs; its total predictive uncertainty $\sigma(x)$ in watts (Eq.~\ref{eq:ensemble_uncertainty_decomposition}) is the evaluator-switching signal, so the threshold is set directly in the physical unit of the iron-loss objective. The outer loop runs on a fixed schedule of five rounds under the per-candidate switch and round-level uncertainty stop of Section~2.3, with every round triggering active learning: the FEA-labeled samples acquired by the FEA--AI hybrid model are merged into the training set and the surrogate is fully fine-tuned for 30 epochs before the next round.

Surrogate generalization is evaluated on a $20\%$ held-out test split for each of the four random seeds after training. For the iron-loss prediction task used in the optimization experiment, the deep ensemble attains a four-seed-average $R^{2}=0.47$ with $\mathrm{RMSE}=31.2\,\mathrm{W}$ and $\mathrm{MAE}=23.1\,\mathrm{W}$, and a mean predictive standard deviation of about $22\,\mathrm{W}$. This accuracy is modest but sufficient for ranking candidates in the GA-based search; it is not, however, reliable enough to be used without verification, partly because of the numerical scatter of the FEA labels. This limitation is addressed by the selective FEA correction of the FEA--AI hybrid loop. Crucially, the predictive standard deviation $\sigma(x)$ lies on the same $10$--$30\,\mathrm{W}$ scale as the switching thresholds used throughout this section, so it meaningfully separates trustworthy predictions from out-of-distribution candidates that must be verified by FEA.

Rather than a preset heuristic constant, the switching threshold $\sigma_{\mathrm{th,switch}}$ is a key design parameter that directly balances high-fidelity FEA cost against surrogate reliability. Accordingly, Section~4.1 reports a rule-based switching sensitivity analysis over $\sigma_{\mathrm{th,switch}}\in\{12,18,24,30\}$\,W, and Section~5 then compares the resulting rule-based and agent-based FEA--AI hybrid variants against FEA-only and AI-only baselines.

For the GA setting, all cases share the same single-objective GA operators: LHS-based population initialization with a population of 25 evolved over 30 generations, tournament selection ($k{=}3$), simulated binary crossover (SBX, $p{=}0.9$, $\eta{=}15$), polynomial mutation ($p{=}0.2$, $\eta{=}20$), and elitist truncation replacement, with the top-$K$ ($K{=}5$) candidates carried between optimization rounds; the NSGA-II variant is exercised in the multi-objective study of Section~5.2. To account for stochasticity in sampling, surrogate training, and GA initialization, every configuration is repeated over the same four random seeds (5, 42, 777, 2026), used for every experiment in Sections~4 and~5, and metrics are averaged over these seeds unless a representative seed is indicated.

\subsection{Rule-based switching}

\begin{figure}[!htbp]
\centering
\begin{minipage}{0.49\linewidth}\centering
  \includegraphics[width=\linewidth]{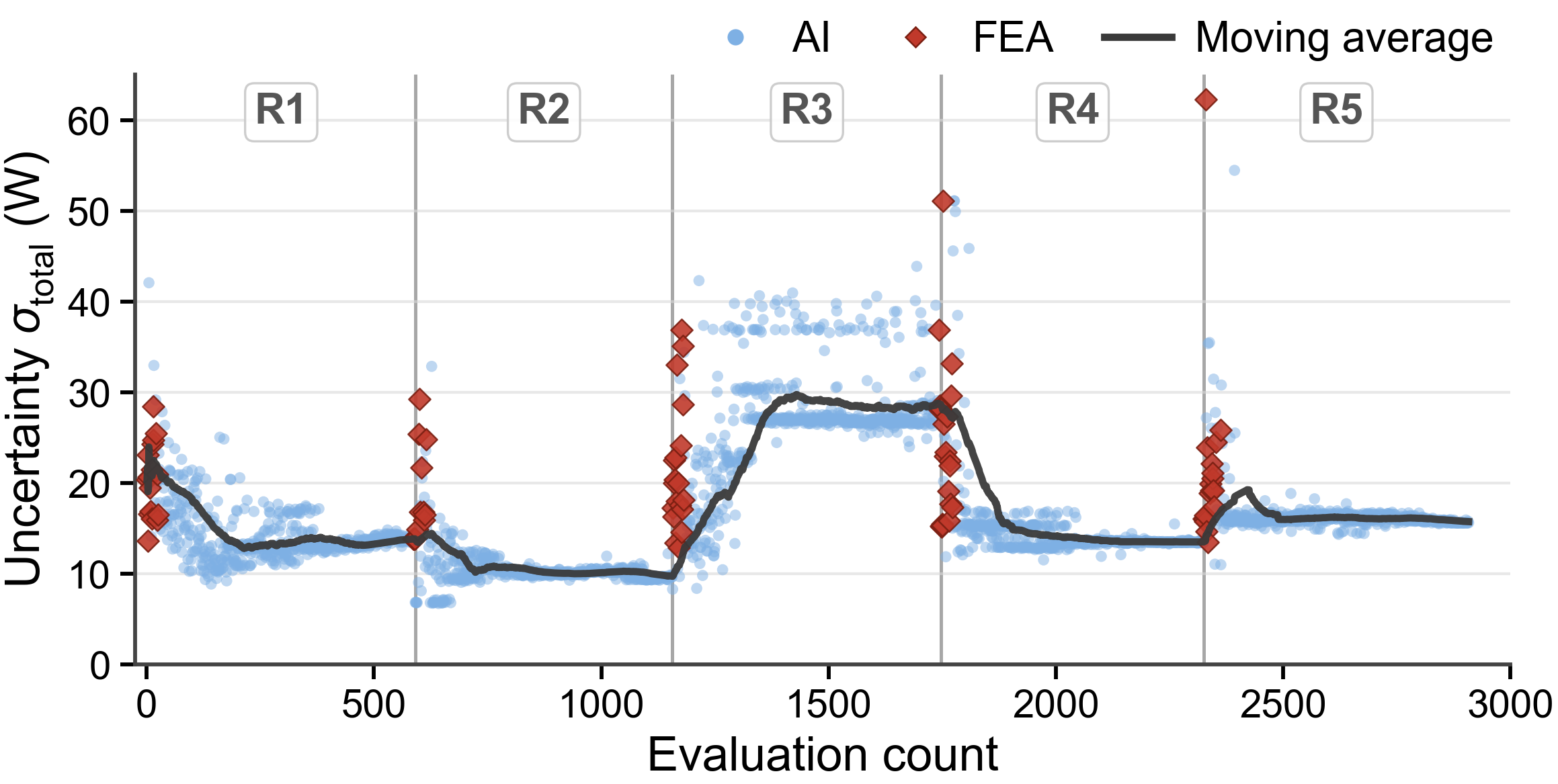}\\[-0.2em]
  {\small (a) $\sigma_{\mathrm{th,switch}}=12$\,W}
\end{minipage}\hfill
\begin{minipage}{0.49\linewidth}\centering
  \includegraphics[width=\linewidth]{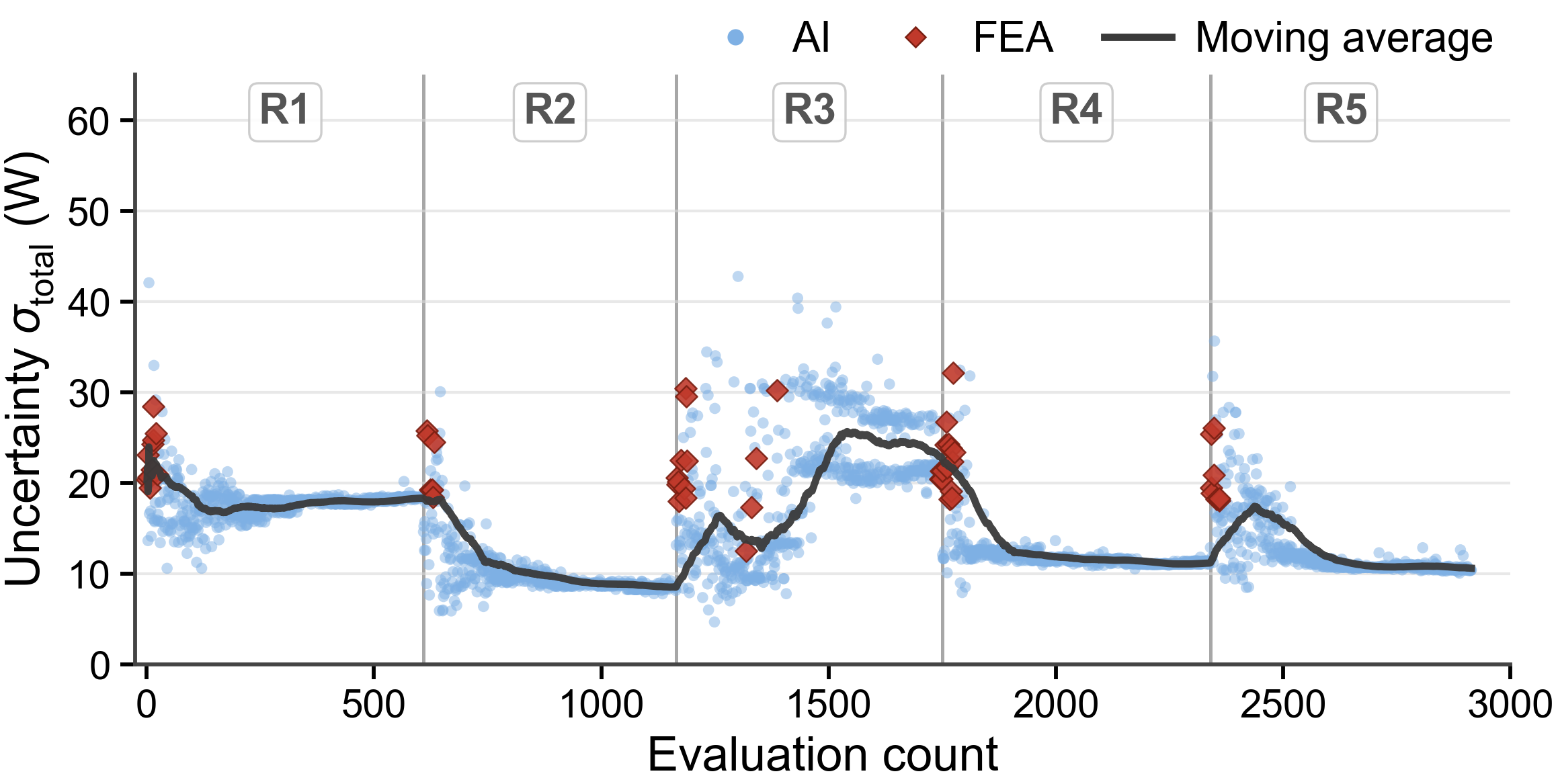}\\[-0.2em]
  {\small (b) $\sigma_{\mathrm{th,switch}}=18$\,W}
\end{minipage}

\vspace{0.6em}

\begin{minipage}{0.49\linewidth}\centering
  \includegraphics[width=\linewidth]{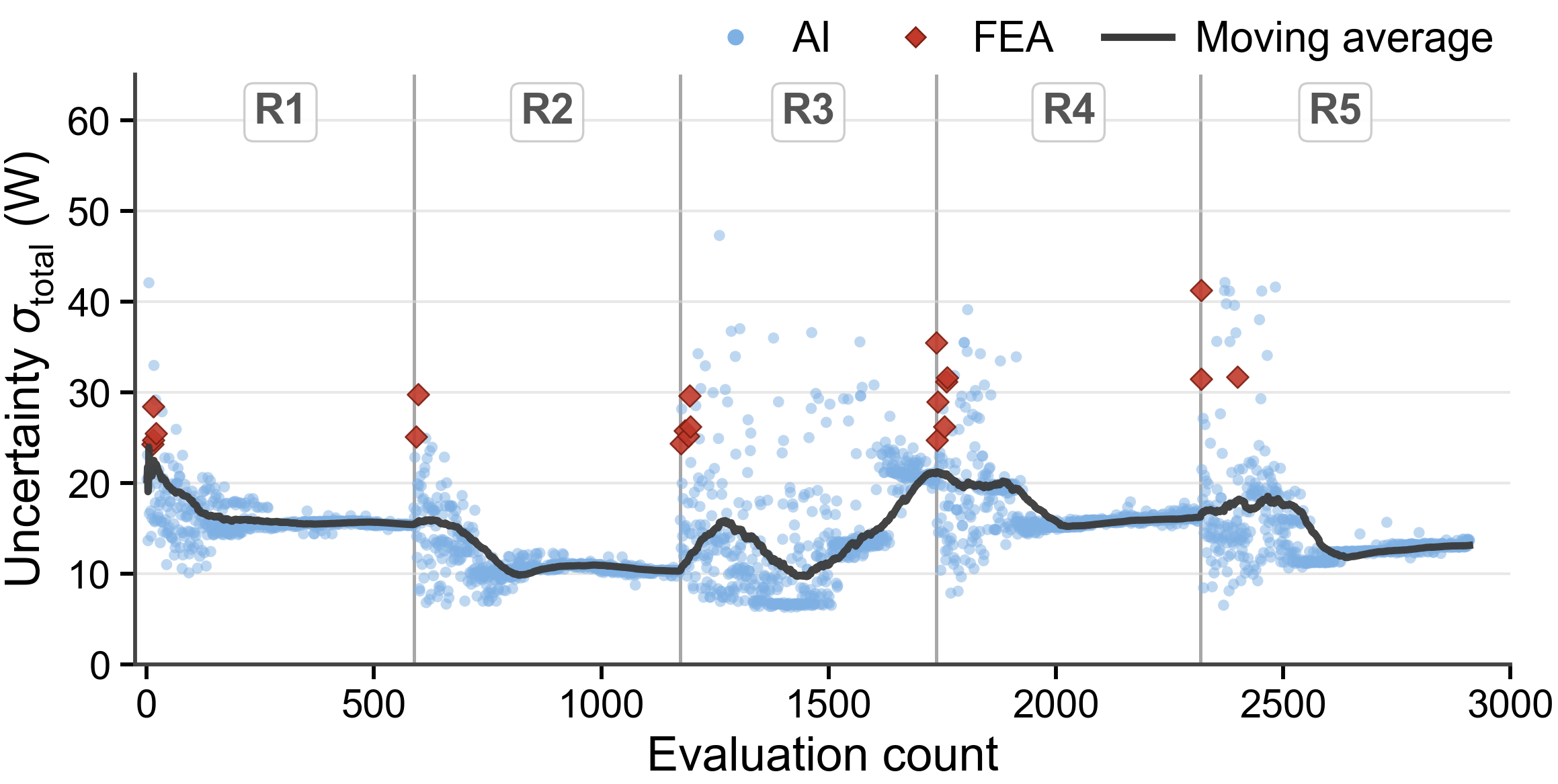}\\[-0.2em]
  {\small (c) $\sigma_{\mathrm{th,switch}}=24$\,W}
\end{minipage}\hfill
\begin{minipage}{0.49\linewidth}\centering
  \includegraphics[width=\linewidth]{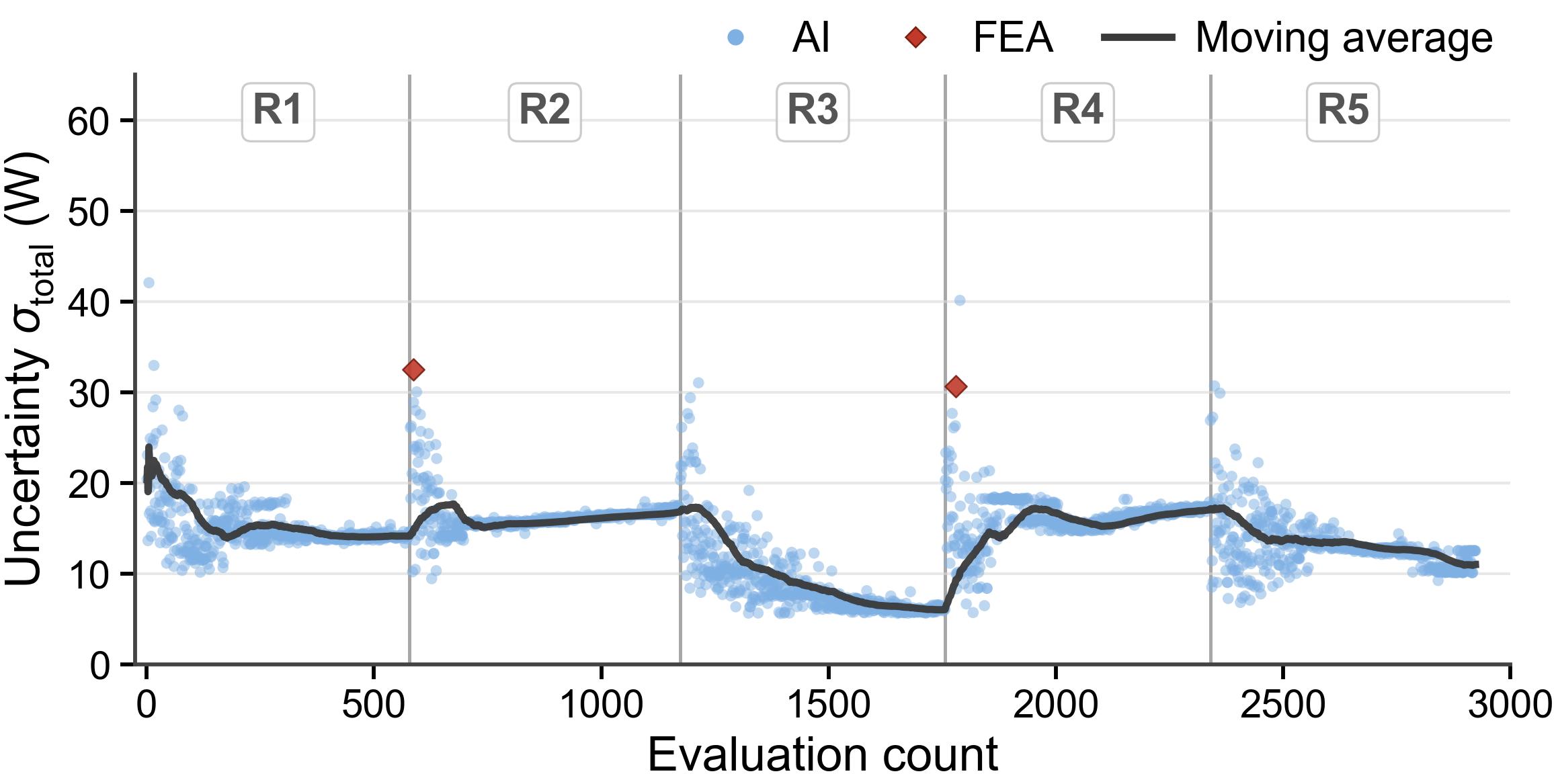}\\[-0.2em]
  {\small (d) $\sigma_{\mathrm{th,switch}}=30$\,W}
\end{minipage}
\caption{Sensitivity analysis of the switching threshold at (a)~$12$, (b)~$18$, (c)~$24$, and (d)~$30$\,W. Each evaluation is marked by the solver used (red for FEA, light blue for the AI surrogate), and the AI-predicted uncertainty at each point is compared across the thresholds.}
\label{fig:sensitivity_uncertainty}
\end{figure}

\begin{figure}[!htbp]
\centering
\begin{minipage}{0.49\linewidth}\centering
  \includegraphics[width=\linewidth]{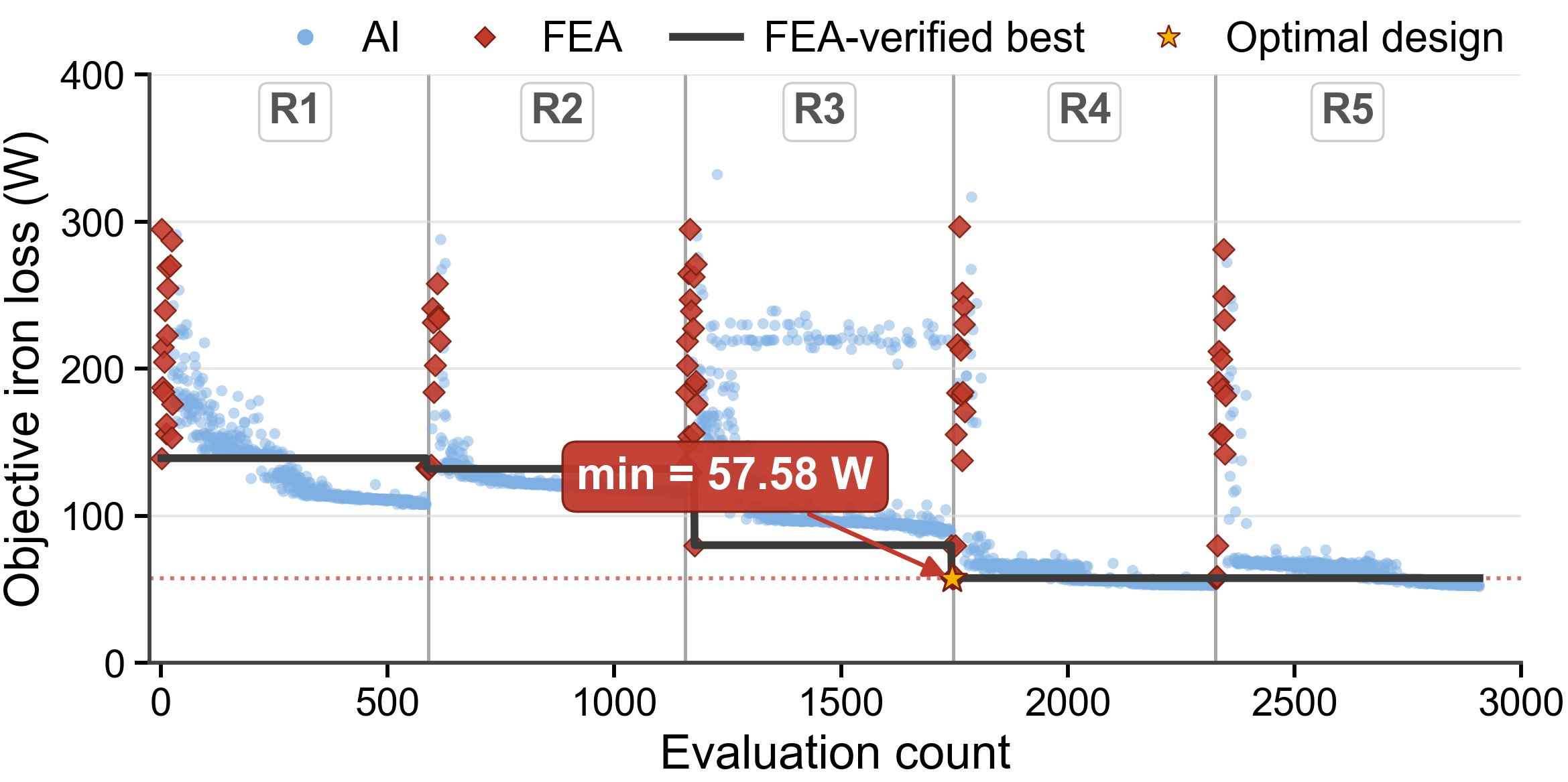}\\[-0.2em]
  {\small (a) $\sigma_{\mathrm{th,switch}}=12$\,W}
\end{minipage}\hfill
\begin{minipage}{0.49\linewidth}\centering
  \includegraphics[width=\linewidth]{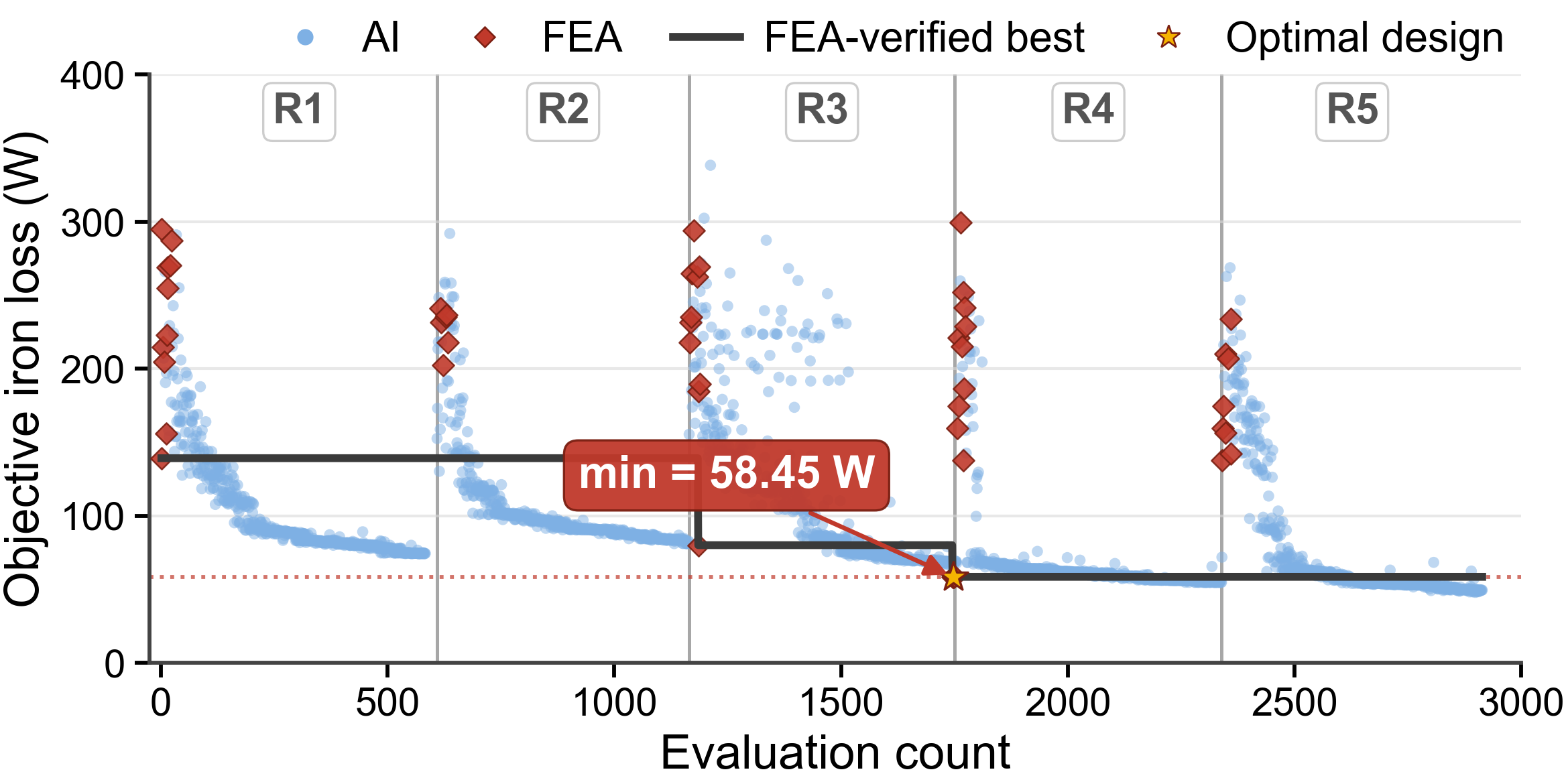}\\[-0.2em]
  {\small (b) $\sigma_{\mathrm{th,switch}}=18$\,W}
\end{minipage}

\vspace{0.6em}

\begin{minipage}{0.49\linewidth}\centering
  \includegraphics[width=\linewidth]{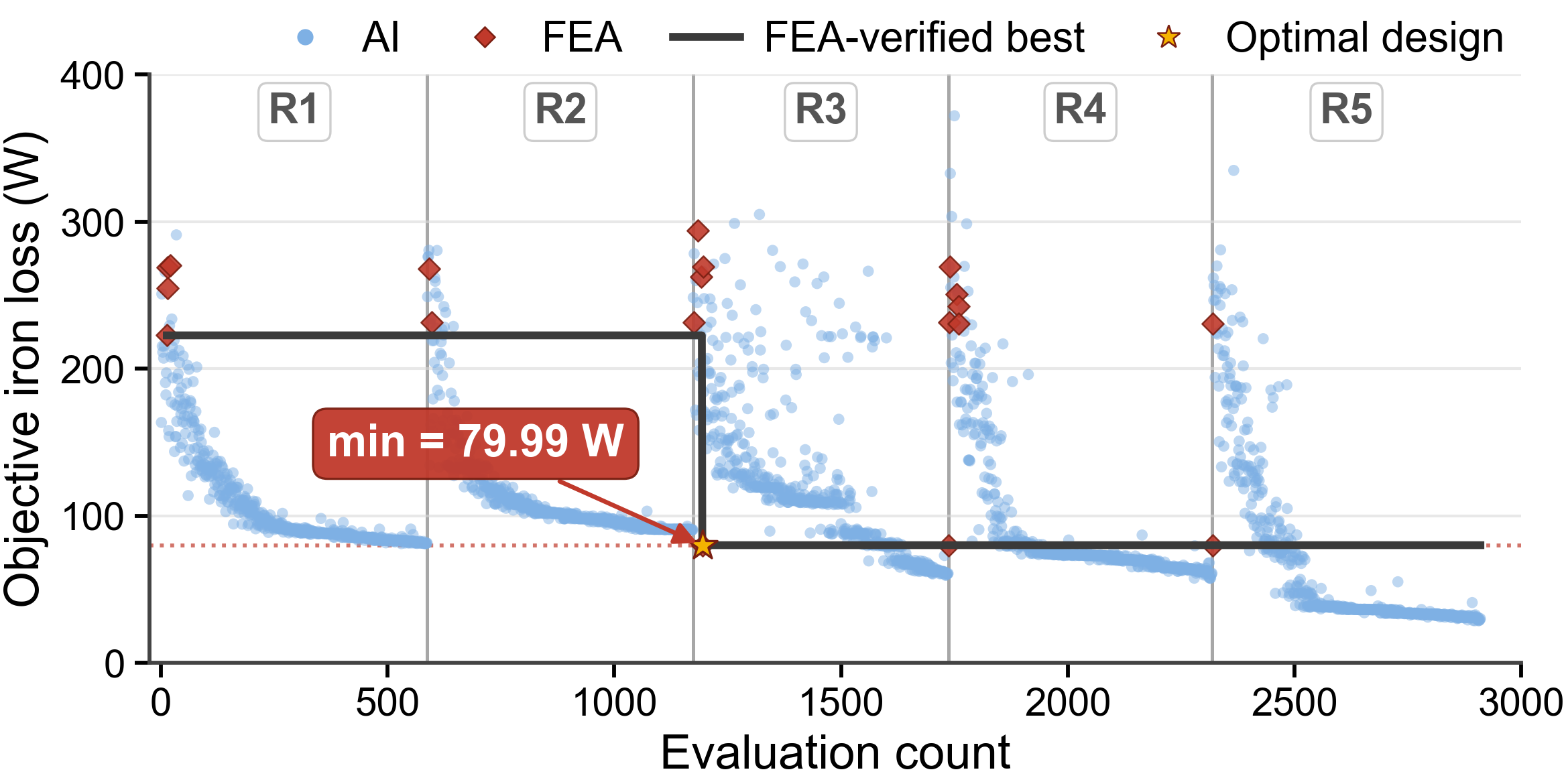}\\[-0.2em]
  {\small (c) $\sigma_{\mathrm{th,switch}}=24$\,W}
\end{minipage}\hfill
\begin{minipage}{0.49\linewidth}\centering
  \includegraphics[width=\linewidth]{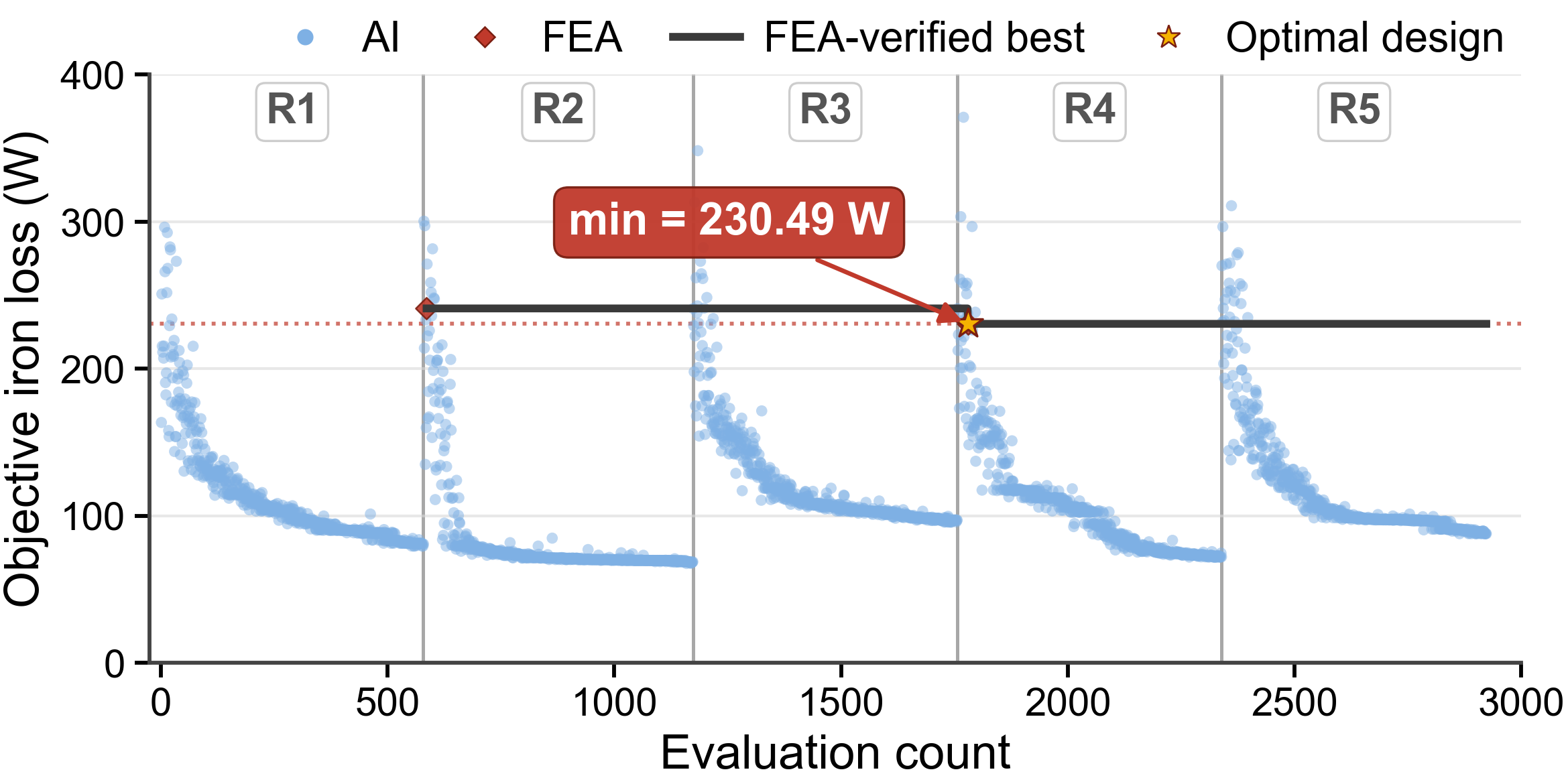}\\[-0.2em]
  {\small (d) $\sigma_{\mathrm{th,switch}}=30$\,W}
\end{minipage}
\caption{Best iron loss per evaluation under switching thresholds of (a)~$12$, (b)~$18$, (c)~$24$, and (d)~$30$\,W. The solver used at each evaluation is marked, together with the minimum iron loss among the FEA-evaluated points. As the threshold increases, fewer FEA calls are made and the minimum FEA-verified iron loss rises.}
\label{fig:sensitivity_objective}
\end{figure}

\begin{figure}[!htbp]
\centering
\includegraphics[width=0.85\linewidth]{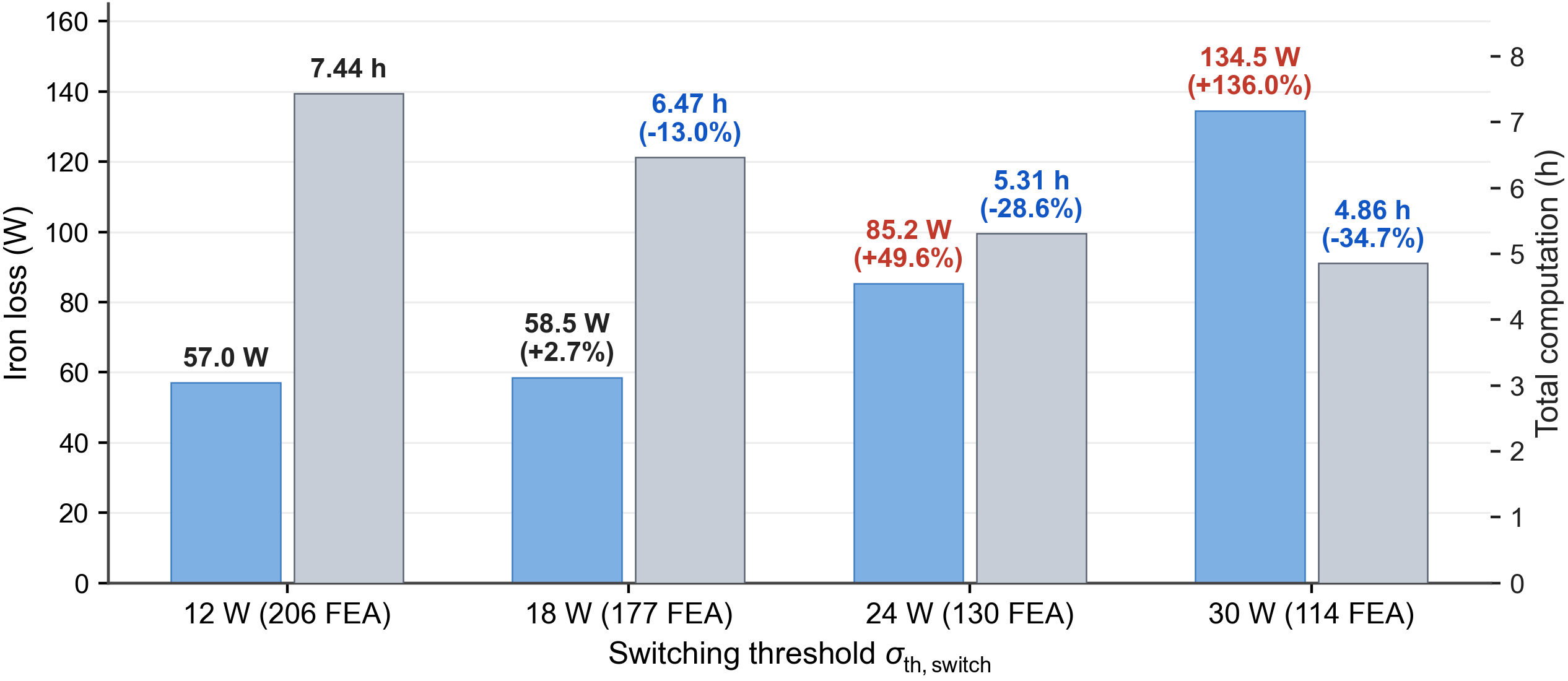}
\caption{Best iron loss (colored bars, left axis) and total computation time (gray bars, right axis) under each switching threshold, with total FEA calls in parentheses. Percentages are relative to the $12$\,W threshold. A higher threshold reduces the number of FEA calls and the computation time but degrades the optimization performance.}
\label{fig:sensitivity_summary}
\end{figure}

In \emph{rule-based switching}, the threshold is held at a single fixed value for the entire optimization run. Because the switching threshold $\sigma_{\mathrm{th,switch}}$ controls how often the optimizer falls back to high-fidelity FEA, it directly determines both the reliability of the surrogate-guided search and its computational cost. To examine its effect, we vary $\sigma_{\mathrm{th,switch}}$ over $\{12,18,24,30\}$\,W and, for each setting, run the full five-round outer-loop optimization. During each run we track the live total uncertainty $\sigma(x)$ of every evaluated candidate, the best objective value, and the total number of FEA calls consumed. Figures~\ref{fig:sensitivity_uncertainty} and~\ref{fig:sensitivity_objective} visualize the representative seed~2026, and Fig.~\ref{fig:sensitivity_summary} summarizes the four-seed averages of the final selected design.

Figure~\ref{fig:sensitivity_uncertainty} shows the live total uncertainty $\sigma(x)$ of every evaluated candidate over the entire optimization, where each panel corresponds to one threshold and the round boundaries (R1--R5) are marked. Notably, the round-wise uncertainty does \emph{not} decay monotonically: because active learning continually relocates the population, the uncertainty rises whenever the search moves into design regions outside the current training distribution and falls again once those regions are corrected by FEA and incorporated into the surrogate.

For the tight thresholds (panels (a) $12$\,W and (b) $18$\,W), the uncertainty shows a pronounced transient spike in the middle rounds (around R3). There the GA explores low-iron-loss geometries that lie outside the initially trained design space, so the surrogate is locally out-of-distribution and the per-candidate uncertainty jumps. Crucially, because the threshold is low, these high-uncertainty candidates trigger frequent FEA correction (red diamonds), and the newly labeled out-of-distribution samples are merged for active learning; after the following round the uncertainty re-stabilizes to a low level even in the newly discovered low-loss region.

For the loose thresholds (panels (c) $24$\,W and (d) $30$\,W), the uncertainty instead stays consistently low and almost flat, with very few FEA corrections. This low uncertainty is misleading: because the threshold is high, the search rarely leaves the initially trained region and candidates are almost never sent to FEA for verification, so the low values indicate that the optimizer remains confined to the in-distribution region, not that the surrogate has actually learned the high-performing region.

The best-objective (iron-loss) trajectory in Fig.~\ref{fig:sensitivity_objective} mirrors this: under the tight thresholds the objective decreases most sharply in precisely the rounds where the uncertainty spikes occur, whereas under the loose thresholds the FEA-verified best settles at a much higher loss. The uncertainty spike and the objective improvement thus reflect the same exploration phase, in which the widest exploration, the largest objective improvement, and the most frequent FEA correction coincide in the middle rounds.

Figure~\ref{fig:sensitivity_summary} aggregates these effects over the four seeds for the final selected design. The most conservative $12$\,W threshold attains the best objective of $57.0$\,W iron loss because it consumes the most FEA, about $206$ calls in total including the $100$ training-set FEA samples common to all thresholds, but it also incurs the highest total computation of $7.44$\,h. The $18$\,W threshold gives up only a small amount of objective performance, reaching $58.52$\,W, while reducing the FEA calls to $177$ and the computation to $6.47$\,h. The loose $24$\,W and $30$\,W thresholds save further computation ($5.31$\,h and $4.86$\,h) but degrade performance markedly, worsening the iron loss to $85.2$\,W and $134.5$\,W. The $18$\,W threshold therefore offers the most reasonable fixed trade-off: relative to the $12$\,W setting it sacrifices only $1.5$\,W of iron loss while saving $29$ FEA calls and nearly an hour of computation, and it avoids the large degradation of the loose settings. It is accordingly adopted as the rule-based switching setting for the remaining experiments, serving both as the baseline for the agent-based switching of Section~4.2 and as the FEA--AI hybrid model configuration in the ablation study of Section~5. At the same time, because the most informative FEA corrections concentrate in the round-varying exploration phases, no single rule-based threshold is optimal throughout the search, which motivates the agent-based switching examined in Section~4.2.

\subsection{Agent-based switching}
\label{sec:llm_control}

Because the sensitivity study's own conclusion points to a threshold that should adapt across rounds, we test an \emph{agent-based} switching mechanism in which the threshold is adaptively updated at the start of every round by an LLM controller running on the same local GPT-oss 20B backbone used by the agents. This role differs from LLM-as-optimizer approaches, in which the LLM itself proposes candidate solutions~\cite{yang2024opro}: here the controller proposes no designs and only decides, from live search signals, how much high-fidelity verification the surrogate-guided search receives. The experimental configuration is identical to the sensitivity test in every other respect, keeping the same deep-ensemble surrogate, GA operators, four random seeds, active-learning updates, and round-level termination rule; \emph{only the switching mechanism is changed}: the rule-based fixed $\sigma_{\mathrm{th,switch}}$ is replaced by a round-dependent threshold $\sigma_{\mathrm{th,switch}}^{(r)}$. The first round uses the sensitivity-test optimum $\sigma_{\mathrm{th,switch}}^{(1)}=18$\,W as its initial value, and the controller updates it from the second round onward.

The controller must determine the extent to which the surrogate can currently be trusted, which it infers primarily from a \emph{calibration ratio} computed at no additional FEA cost from the designs already evaluated by FEA in the previous round. Let $\mathcal{F}_{r-1}$ be that set of FEA-evaluated designs, with ground-truth labels $y(x)$; the calibration ratio is
\begin{equation}
\rho_r=\frac{\dfrac{1}{|\mathcal{F}_{r-1}|}\sum_{x\in\mathcal{F}_{r-1}}\big|\mu(x)-y(x)\big|}{\dfrac{1}{|\mathcal{F}_{r-1}|}\sum_{x\in\mathcal{F}_{r-1}}\sigma(x)},
\label{eq:calib_ratio}
\end{equation}
i.e., the ratio of the mean absolute surrogate error to the mean predicted standard deviation, both in watts (we write $\rho$ for $\rho_r$ when the round is clear from context), and it requires no additional FEA because the labels already exist:
\begin{itemize}
  \item $\rho_r>1$: the real error exceeds the predicted $\sigma$, so the surrogate is \emph{over-confident} $\rightarrow$ \emph{lower} $\sigma_{\mathrm{th,switch}}^{(r)}$ to acquire additional ground-truth FEA evaluations;
  \item $\rho_r\approx1$: well calibrated $\rightarrow$ hold the threshold;
  \item $\rho_r<1$: more accurate than its predicted uncertainty indicates (\emph{under-confident}) $\rightarrow$ \emph{raise} $\sigma_{\mathrm{th,switch}}^{(r)}$ to reduce FEA usage.
\end{itemize}
Three secondary signals refine the decision when calibration is available: the round-mean uncertainty trend $\Delta\bar{\sigma}$, the best-objective trend $\Delta\mathrm{obj}$, and FEA-budget pacing. A falling uncertainty trend indicates that active learning is improving the surrogate and permits a higher threshold; likewise, a low-uncertainty objective plateau implies convergence and permits a higher threshold, whereas a worsening objective implies the search is stuck on a surrogate artifact and calls for a lower one. An explicit safeguard covers the case $\mathcal{F}_{r-1}=\emptyset$ (no FEA last round, so $\rho_r$ is undefined). Because an absence of calibration evidence can never justify trusting the surrogate \emph{more}, the controller is then forbidden from raising the threshold and instead lowers it below the round-mean uncertainty to re-trigger FEA. This rule structurally prevents a runaway behavior in which the absence of FEA eliminates the calibration evidence and allows the threshold to loosen indefinitely. The backbone receives these signals as a compact JSON state under a rule-constrained, role-assigning prompt of the same template family as the other agents (Section~2), reasons over them, and returns $\sigma_{\mathrm{th,switch}}^{(r)}\in[8,35]$\,W subject to a smoothness limit of $6$\,W per round; a deterministic policy encoding the same update logic serves as a fallback when the LLM is unavailable.

\begin{table}[!htbp]
\centering
\caption{Round-by-round trace of the LLM threshold controller: calibration ratio $\rho$ (Eq.~\ref{eq:calib_ratio}), switching threshold $\tau=\sigma_{\mathrm{th,switch}}^{(r)}$, per-round FEA calls, best iron loss, and controller reasoning quoted from the decision log.}
\label{tab:llm_reasoning}
\footnotesize
\begin{tabular}{cccccp{7.4cm}}
\toprule
Round & $\rho$ & $\tau$ (W) & FEA calls & Best iron loss (W) & Controller reasoning \\
\midrule
R1 & --- & $\mathbf{18}$ & $40$ & $139.1$ & Initial value taken from the sensitivity sweep of Section~4.1; no calibration signal yet. \\
R2 & $0.46$ & $\mathbf{20}$ & $7$ & $139.1$ & ``Surrogate is under-confident ($\rho<1$), so a modest $\tau$ rise saves FEA while still catching errors.'' \\
R3 & $0.19$ & $\mathbf{24}$ & $3$ & $80.0$ & ``It is still under-confident and its uncertainty is improving, so we can safely raise $\tau$.'' \\
R4 & $4.08$ & $\mathbf{20}$ & $15$ & $60.2$ & ``$\rho>1$ means the surrogate is over-confident, so we cut $\tau$ to trigger more FEA here.'' \\
R5 & $0.69$ & $\mathbf{22}$ & $5$ & $\mathbf{49.6}$ & ``Calibration has recovered and the objective still improves, so a modest raise is safe.'' \\
\bottomrule
\end{tabular}
\end{table}

\begin{figure}[!htbp]
\centering
\includegraphics[width=0.8\linewidth]{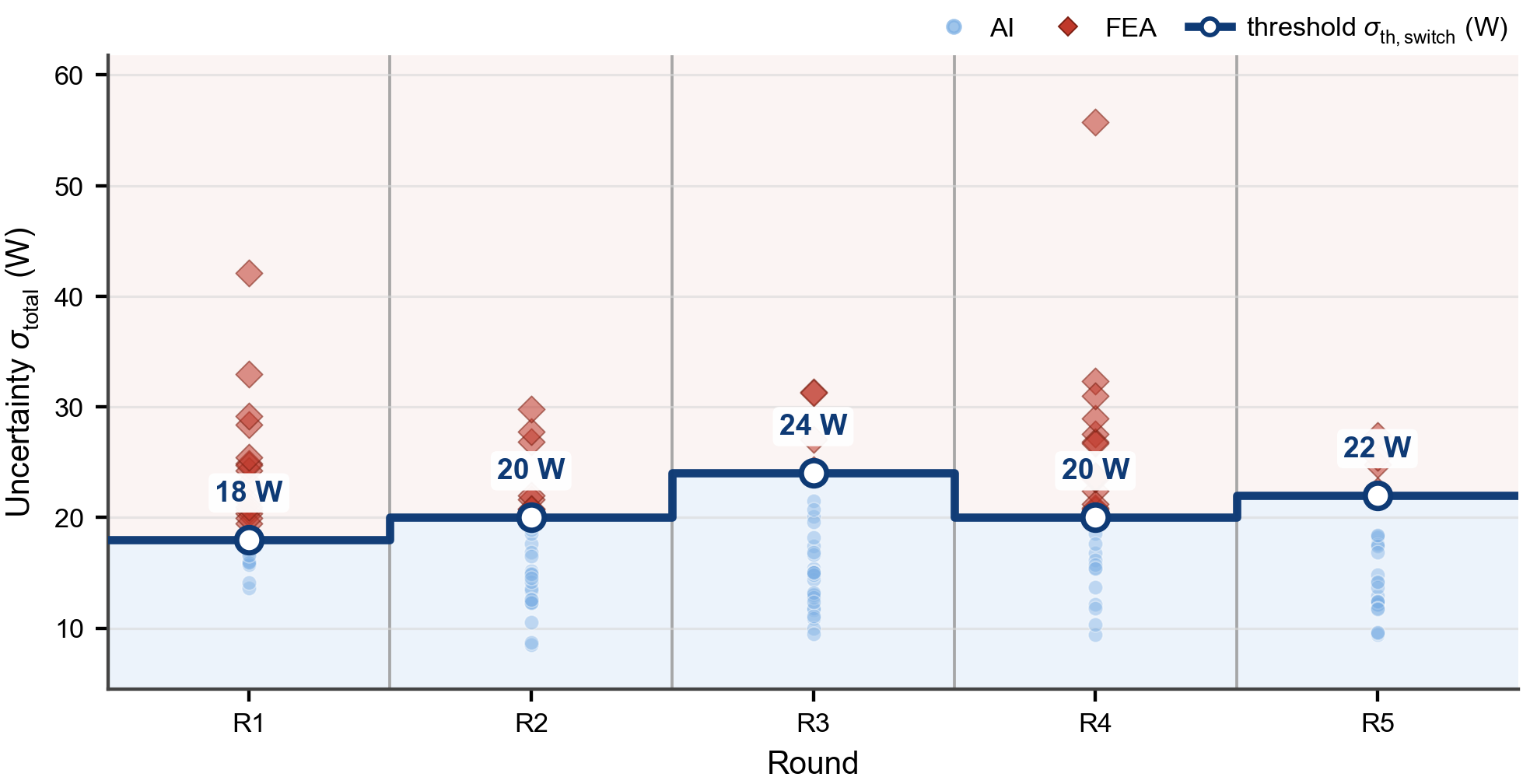}
\caption{Per-round switching threshold $\sigma_{\mathrm{th,switch}}^{(r)}$ chosen by the LLM controller through the reasoning traced in Table~\ref{tab:llm_reasoning}, and the resulting change in FEA and AI-surrogate usage across the rounds.}
\label{fig:llm_threshold_switching}
\end{figure}

Table~\ref{tab:llm_reasoning} traces, round by round, the calibration signal, the threshold it produces, the FEA calls that follow, the best objective reached, and the controller's own reasoning behind each update, while Fig.~\ref{fig:llm_threshold_switching} shows the resulting threshold together with the per-candidate switching gate for the representative seed~2026. The threshold is not monotonic: it is \emph{raised} whenever the surrogate proves trustworthy and \emph{lowered} the moment it does not. In rounds~2 and~3 the previous-round calibration shows the surrogate to be strongly under-confident ($\rho=0.46$ then $0.19$), and by round~3 the round-mean uncertainty is also falling, so the controller raises the threshold from $18$ to $20$ and then to $24$\,W to avoid spending FEA on a surrogate whose predictions are more accurate than its uncertainty estimates indicate. The critical behavior occurs in round~4: after the search discovers a much lower-loss region in round~3 (the best iron loss drops from about $139$ to $80$\,W), the surrogate becomes severely over-confident there (its predicted $\sigma$ of about $27$\,W is far smaller than the actual error of roughly $110$\,W, giving $\rho=4.08$), and the controller immediately lowers the threshold back to $20$\,W to re-acquire FEA ground truth in the new region, raising the per-round FEA count from $3$ back to $15$. By round~5 calibration has recovered ($\rho=0.69$) while the objective keeps improving, so, with the calibration signal taking precedence over the rising uncertainty trend, the threshold is raised slightly to $22$\,W. Because these re-acquired FEA labels are folded back into the surrogate through round-wise active-learning retraining, the round-5 recovery reflects a surrogate that has become genuinely better calibrated in the newly discovered near-optimal region, not merely a threshold that has moved. This ability to tighten the threshold precisely when the surrogate becomes unreliable cannot be achieved with a single rule-based threshold.

\begin{figure}[!htbp]
\centering
\includegraphics[width=0.6\linewidth]{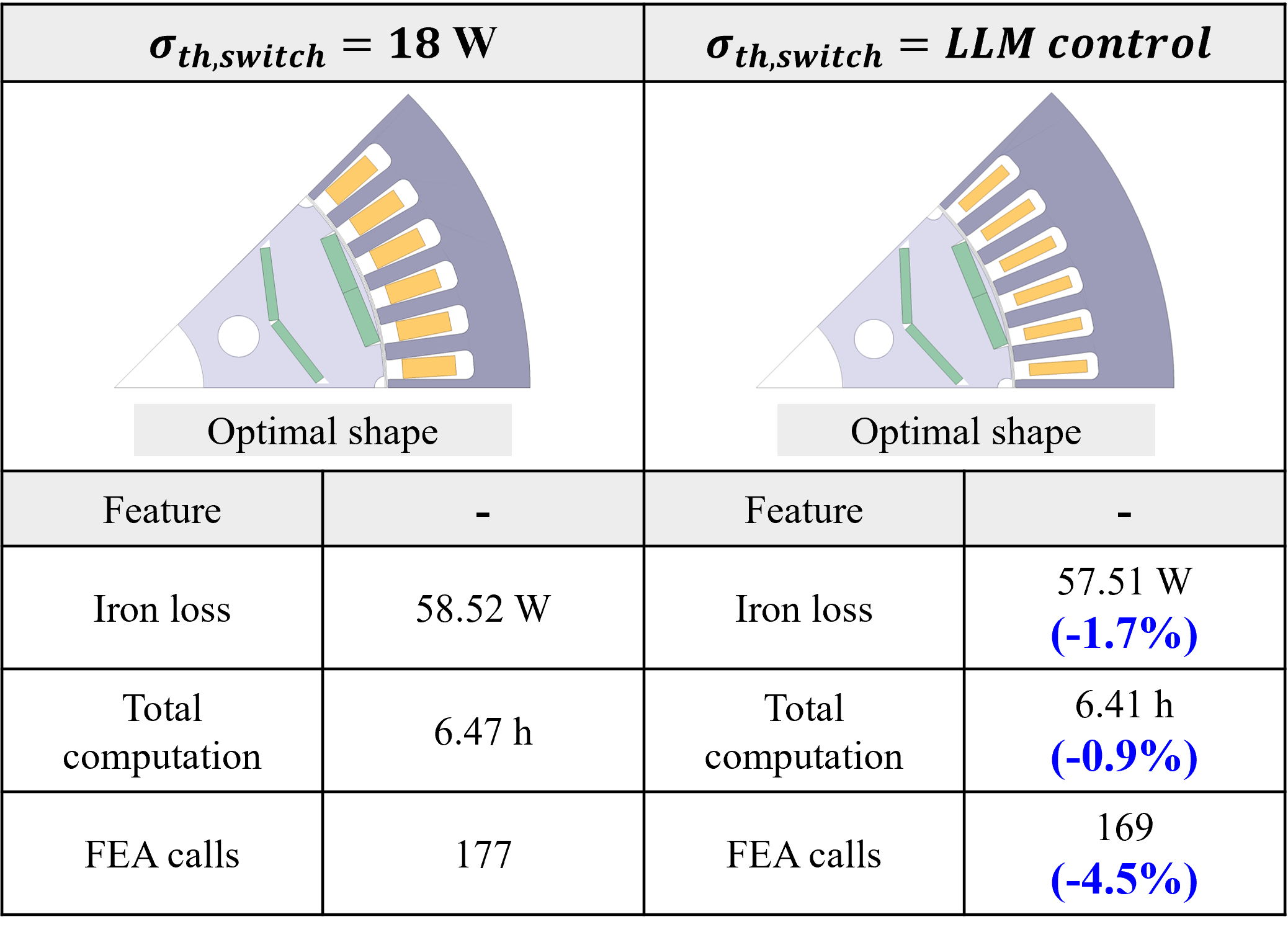}
\caption{Agent-based versus rule-based ($18$\,W) switching: optimal shape, iron loss, total computation time, and total FEA calls. Adaptively updating the threshold every round yields a slightly lower iron loss with fewer FEA calls and less computation than the rule-based setting.}
\label{fig:llm_fixed_compare}
\end{figure}

Figure~\ref{fig:llm_fixed_compare} compares the agent-based switching against the rule-based $18$\,W setting adopted in Section~4.1, averaged over the four seeds. Note that the $18$\,W baseline was selected on these same runs, so it enters the comparison favorably tuned; the comparison is therefore conservative with respect to the controller. On the seed average, the controller achieves a more favorable trade-off, reaching a slightly lower iron loss of $57.51$ versus $58.52$\,W with fewer FEA calls, $169$ versus $177$, and less computation, $6.41$ versus $6.47$\,h. These margins are small, about $1$\,W of iron loss and eight FEA calls, indicating that per-round reasoning matches, and tends to improve on, the best manually tuned rule-based threshold without requiring the sensitivity sweep that produced it. A structural explanation is that the rule-based policy commits to a single uncertainty level for the whole run, whereas the controller monitors live signals (calibration, the uncertainty trend, and objective progress) and reallocates FEA toward the rounds that need it, tightening when the surrogate is exposed as over-confident and relaxing when it is reliable. Replacing a manually set constant with lightweight per-round reasoning thus attains the best rule-based trade-off automatically and suggests richer adaptive-control policies as a direction for future work.

\section{Ablation study: same-FEA and same-evaluation budget}

The ablation study compares four optimization strategies:
\begin{itemize}
  \item \textbf{FEA-only GA:} evaluates every candidate with high-fidelity FEA; the most reliable strategy, but every evaluation incurs the full solver cost, so the computation time is substantial;
  \item \textbf{AI-only GA:} searches with the trained surrogate alone; evaluation is fast, but reliability is low because prediction errors are never corrected and grow in out-of-distribution regions;
  \item \textbf{FEA--AI hybrid GA (rule):} the FEA--AI hybrid model with rule-based switching, keeping the threshold fixed at the $18$\,W value of Section~4.1;
  \item \textbf{FEA--AI hybrid GA (agent):} the FEA--AI hybrid model with agent-based switching, adaptively updating the threshold each round via the LLM controller of Section~4.2.
\end{itemize}
Each strategy is evaluated on both a single-objective (Section~5.1) and a multi-objective (Section~5.2) problem, under two budget definitions each, a \emph{same-FEA} budget and a \emph{same-evaluation} budget, giving a $2\times2$ set of comparisons.

All four strategies share one round-based GA configuration (population $20$, $7$ generations, $12$ parents, top $4$ carried between rounds, giving $\approx100$ evaluations per round; standard GA for single-objective and NSGA-II for multi-objective), so the study isolates the effect of the evaluation strategy alone. This per-round GA is deliberately smaller than that of Section~4, so the absolute iron losses below are not directly comparable to the Section~4.1 values.

\subsection{Single-objective optimization}

The two budget definitions probe complementary questions: the same-FEA setting compares the optimization performance attainable under the same FEA budget and computation time, whereas the same-evaluation setting measures how much computational cost each strategy saves while performing the same number of GA evaluations.

\subsubsection{Same FEA budget}

\begin{figure}[!htbp]
\centering
\includegraphics[width=0.85\linewidth]{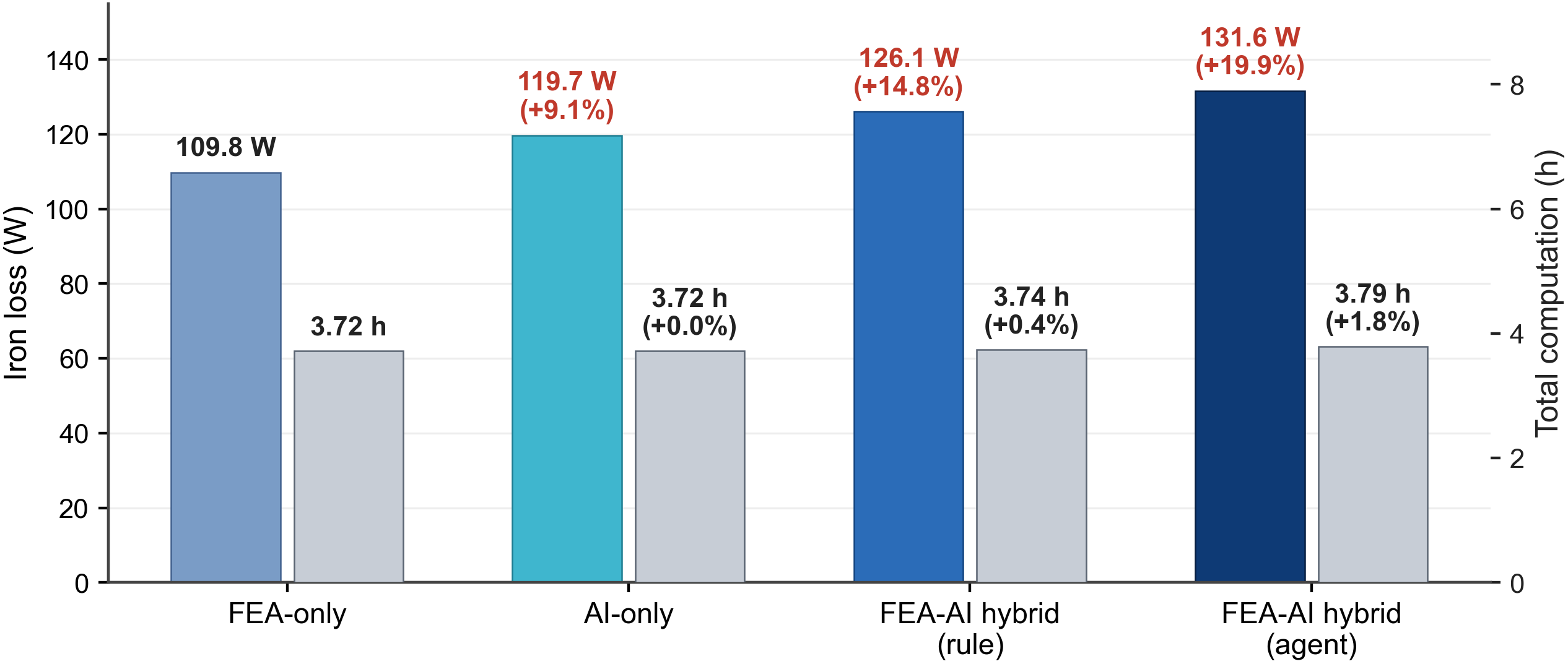}\\[1pt]
{\small (a) FEA budget of $100$ calls}\\[6pt]
\includegraphics[width=0.85\linewidth]{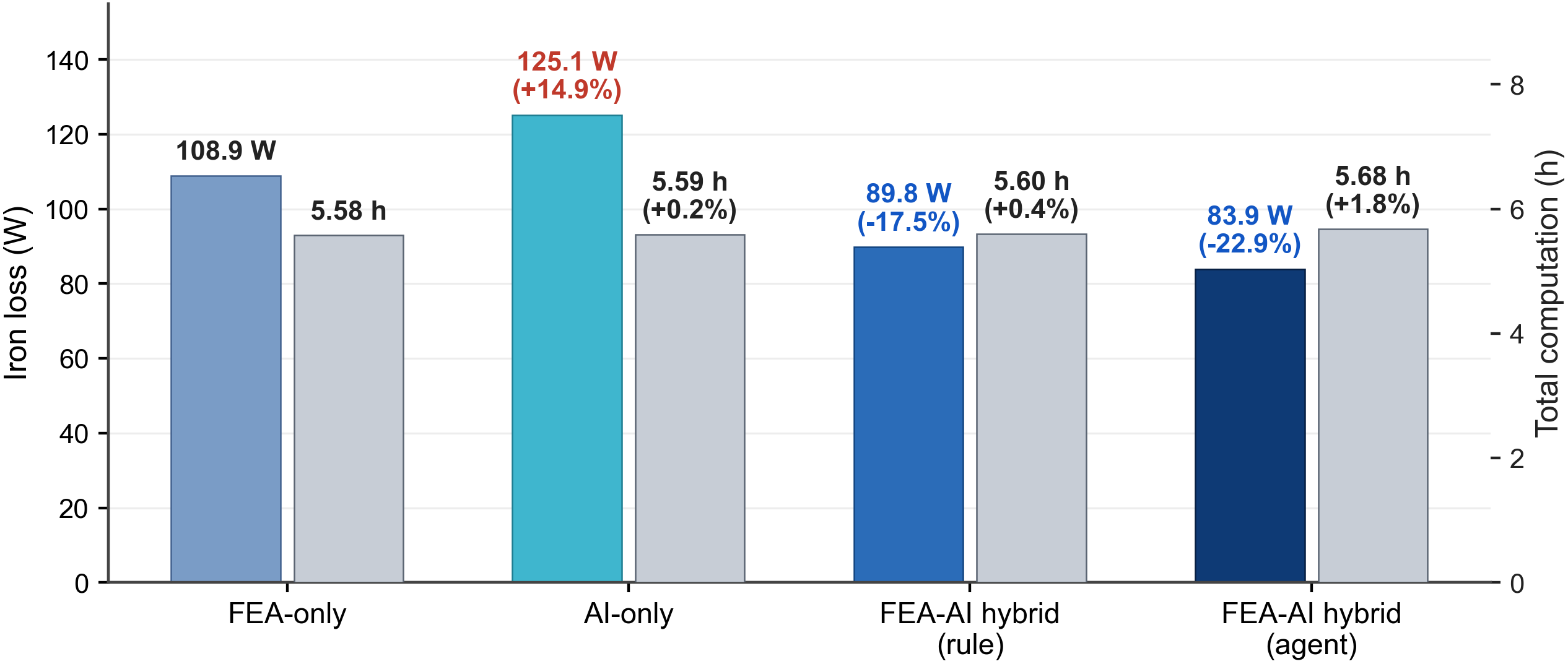}\\[1pt]
{\small (b) FEA budget of $150$ calls}\\[6pt]
\includegraphics[width=0.85\linewidth]{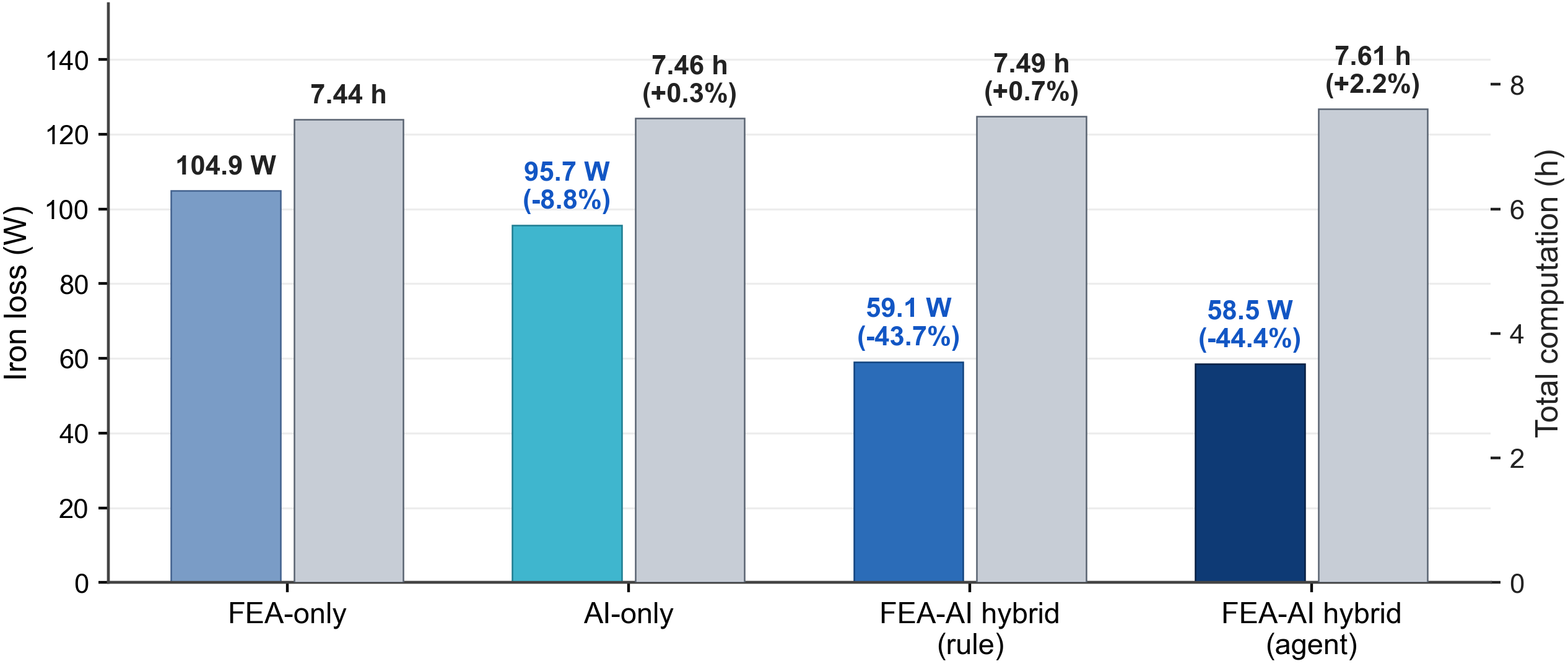}\\[1pt]
{\small (c) FEA budget of $200$ calls}
\caption{Same-FEA-budget comparison at (a)~$100$, (b)~$150$, and (c)~$200$ FEA calls. Colored bars denote the best iron loss (left axis), gray bars the total computation time (right axis), and percentages are relative to the FEA-only GA at the same budget. As the FEA budget increases, the advantage of the FEA--AI hybrid model over the FEA-only GA widens.}
\label{fig:compare_same_fea}
\end{figure}

The first study compares the four strategies under three high-fidelity budgets of $100$, $150$, and $200$ FEA calls. At each budget, the FEA-only GA spends every call on direct evaluation; the AI-only GA invests the entire budget in training samples for its surrogate and then searches by inference alone, with its evaluation count matched to that of the rule-based variant; and the two FEA--AI hybrid variants split the budget between surrogate training and uncertainty-triggered online correction, $50$ training samples plus $50$ online calls at the $100$-call budget and $100$ training samples plus $50$ or $100$ online calls at the $150$- and $200$-call budgets. The two variants differ only in the switching threshold, fixed at the $\sigma_{\mathrm{th,switch}}=18$\,W value of Section~4.1 for the rule-based variant and adaptively updated each round as in Section~4.2 for the agent-based variant. Figure~\ref{fig:compare_same_fea} reports the four-seed-average iron loss and total computation time at each budget; within each budget the four strategies take nearly identical wall-clock time, differing by at most $2.2\%$, so the comparison isolates where the high-fidelity calls are spent.

At the $100$-call budget (Fig.~\ref{fig:compare_same_fea}(a)), no strategy optimizes effectively, and the FEA--AI hybrid GA performs worst: relative to the FEA-only result of $109.79$\,W, the AI-only design ends $9.1\%$ higher, and the rule-based and agent-based variants end $14.8\%$ and $19.9\%$ higher. The reason lies in the training split: half of the budget leaves only $50$ training samples, and with such limited data not only the mean prediction but also the uncertainty estimate itself is poorly calibrated, so the switching verifies the wrong candidates. The agent-based variant performs worst because its per-round reasoning adapts the threshold to signals that are dominated by noise.

At the $150$-call budget (Fig.~\ref{fig:compare_same_fea}(b)), the ordering reverses and the strength of the FEA--AI hybrid model emerges: now trained on $100$ samples with $50$ online corrections, the rule-based and agent-based variants reach $17.5\%$ and $22.9\%$ lower iron loss than the FEA-only GA, whereas the AI-only design remains $14.9\%$ worse. Between the two variants, agent-based switching outperforms the rule-based setting, $83.88$ versus $89.76$\,W.

At the $200$-call budget (Fig.~\ref{fig:compare_same_fea}(c)), the advantage of the FEA--AI hybrid model becomes decisive: the rule-based and agent-based variants reach $43.7\%$ and $44.4\%$ lower iron loss than the FEA-only GA, nearly halving it, and the AI-only design, although improved by its larger training set, remains far above both variants. The agent-based variant again finishes ahead of the rule-based one, $58.51$ versus $59.09$\,W.

Across the three budgets the trend is clear: as the FEA budget grows, the FEA--AI hybrid model changes from the worst-performing strategy at $100$ calls to the strongest by a widening margin, $17.5$--$22.9\%$ better than FEA-only at $150$ calls and $43.7$--$44.4\%$ at $200$. The FEA-only GA benefits only marginally from the larger budget, improving by only about $5$\,W from $100$ to $200$ calls, because every additional call is consumed by whichever candidate the GA proposes next. The FEA--AI hybrid model instead allocates the same increment to three complementary uses, additional training data, additional uncertainty-targeted verification, and additional active-learning updates, so once the surrogate is trustworthy on average, its evaluated candidate pool grows from hundreds to over $1{,}700$ designs within the same budget. Because practical GA-based motor optimization involves thousands to tens of thousands of evaluations, the benefit of the FEA--AI hybrid model is expected to grow accordingly. At the two larger budgets the agent-based variant stays slightly ahead of the rule-based one because, as analyzed in Section~4.2, its per-round reasoning over the live calibration, uncertainty, and objective signals allocates the online FEA budget more effectively than the fixed $18$\,W threshold.

\begin{figure}[!htbp]
\centering
\includegraphics[width=0.8\linewidth]{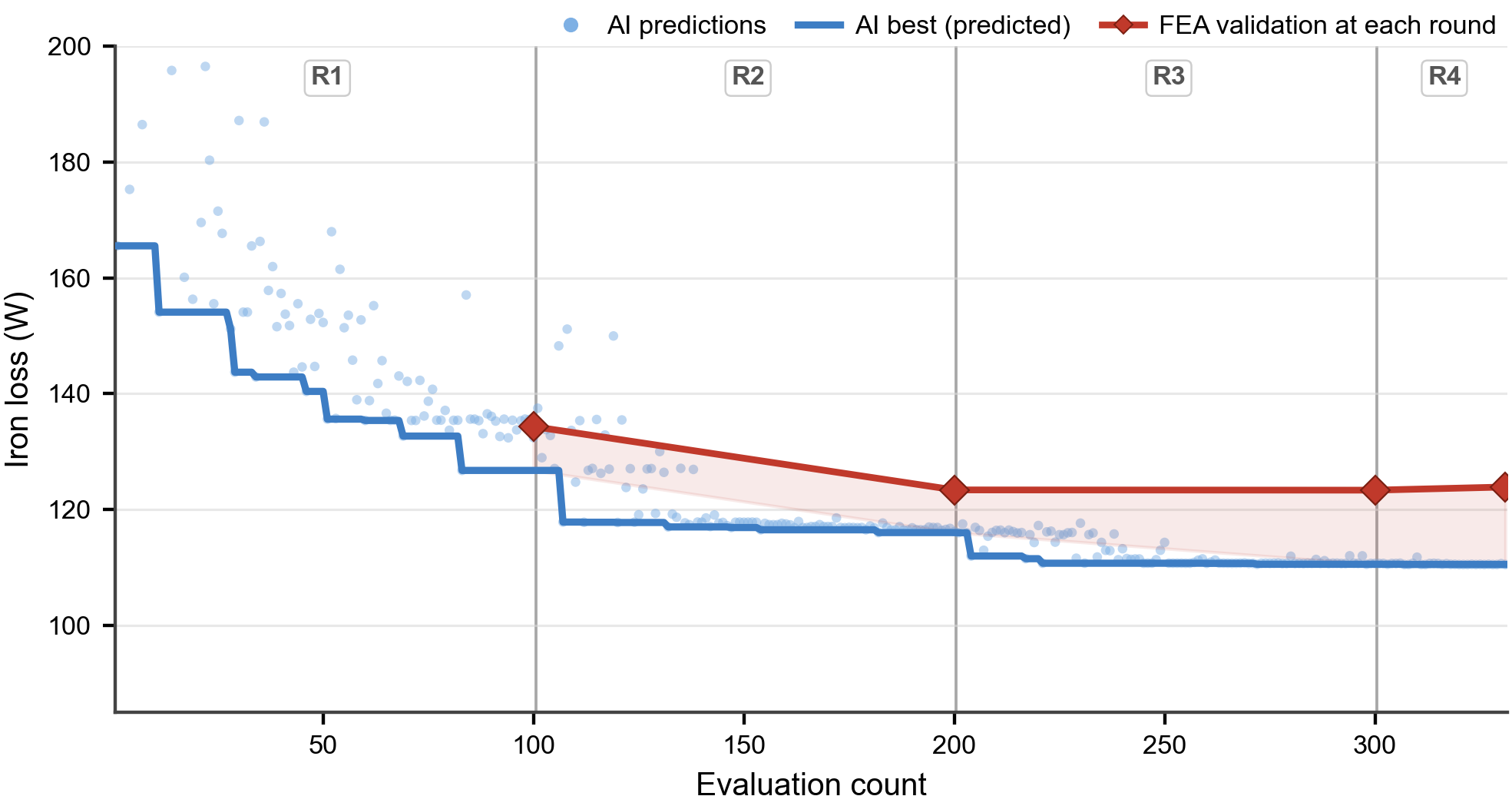}
\caption{Optimization trajectory of the AI-only strategy compared with the FEA-verified value at each round. The AI-predicted iron loss keeps decreasing, whereas the FEA verification shows the actual iron loss increasing.}
\label{fig:ai_only_rounds}
\end{figure}

The behavior of the AI-only strategy becomes clear when its search is traced round by round (Fig.~\ref{fig:ai_only_rounds}). The \emph{AI best (predicted)} curve keeps decreasing, but the \emph{FEA validation at each round}, a post-hoc diagnostic not counted toward the budgets, tells a different story: the validated loss improves only from the first to the second round, and over rounds R3--R4 the prediction continues to converge while the validated loss drifts slightly upward, so the actual performance of the selected designs degrades rather than improves. The search thus drifts toward a region that appears favorable only to the surrogate, which is precisely the failure mode that the switching threshold of the FEA--AI hybrid model is designed to catch.

\subsubsection{Same evaluation budget}

The second study instead fixes the total number of evaluations at $300$ and lets the FEA-call count vary, so that the strategies are compared at equal search effort. The FEA-only GA uses FEA for every evaluation and therefore consumes $300$ FEA calls; the AI-only GA and both FEA--AI hybrid variants use a surrogate trained on $100$ FEA samples and run until $300$ evaluations are reached, with the two variants adding uncertainty-triggered online FEA on top of that $100$-sample training set.

\begin{figure}[!htbp]
\centering
\includegraphics[width=0.85\linewidth]{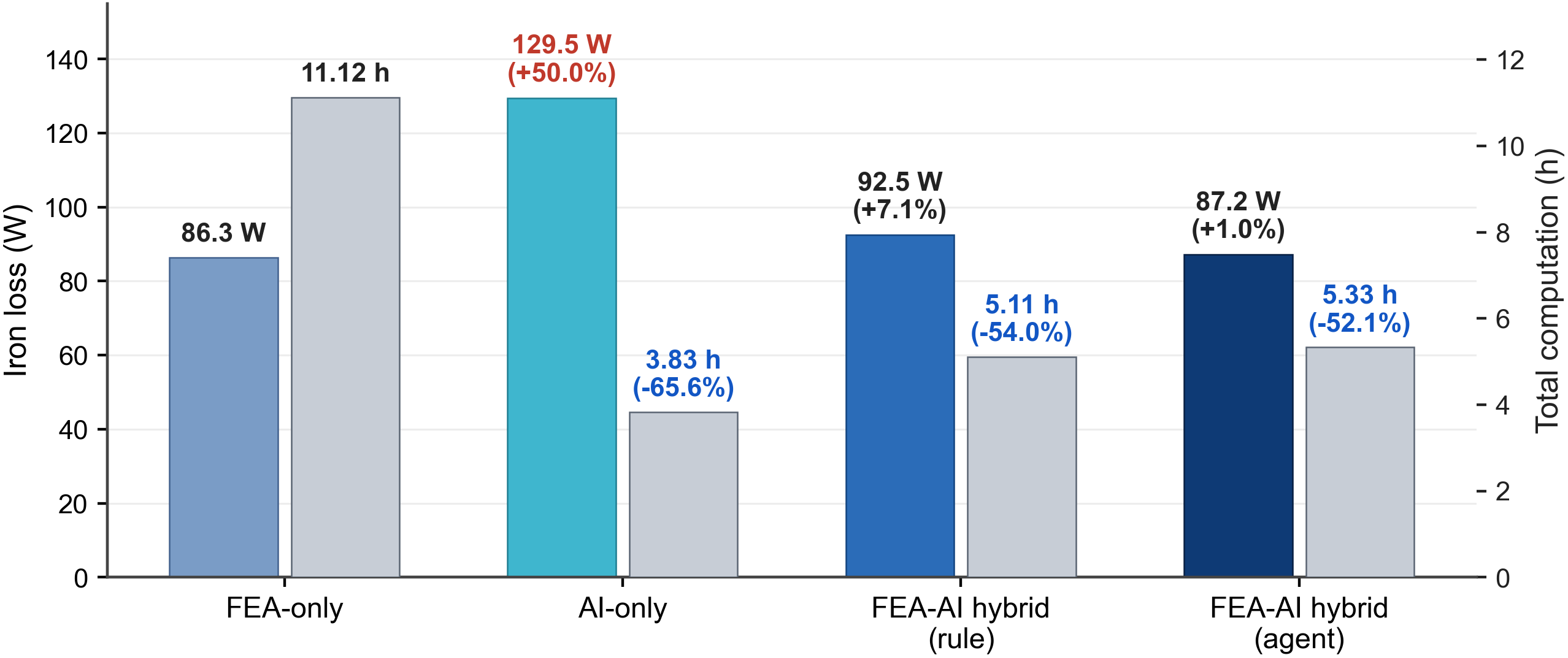}
\caption{Same-evaluation-budget comparison under $300$ evaluations per strategy. Colored bars denote the best iron loss (left axis), gray bars the total computation time (right axis), and percentages are relative to the FEA-only GA. The FEA--AI hybrid variants stay close to the FEA-only optimum at less than half its computation time, whereas the AI-only search is fastest but returns the worst validated design.}
\label{fig:compare_same_eval}
\end{figure}

Because every strategy now performs the same $300$ evaluations, the objective ranking follows the amount of high-fidelity information actually used (Fig.~\ref{fig:compare_same_eval}). The FEA-only GA, which validates all $300$ evaluations with FEA, finds the best design at $86.35$\,W, but at by far the highest computational cost of $11.12$\,h for its $300$ FEA calls. The AI-only GA, relying on its $100$-sample surrogate with no online correction, again relies excessively on its own predictions and produces the worst design at $129.51$\,W, albeit in the shortest time of $3.83$\,h. The two FEA--AI hybrid variants sit in between: the rule-based variant uses $136$ FEA calls and reaches $92.50$\,W, and the agent-based variant uses $144$ FEA calls and reaches $87.18$\,W, both in less than half the FEA-only computation time at $5.11$ and $5.33$\,h. The agent-based variant is particularly notable: with fewer than half the high-fidelity calls of the FEA-only search, it reaches within $1$\,W of the FEA-only optimum. Compared with the rule-based variant it spends slightly more FEA, $144$ versus $136$ calls, but its per-round reasoning places these calls where the surrogate is least trustworthy, converting the small extra budget into a clearly better objective of $87.18$ versus $92.50$\,W. More importantly for practice, although the FEA-only search attains the best objective when the evaluation count is small, real GA-based motor optimization routinely requires thousands to tens of thousands of evaluations, at which scale an all-FEA search is computationally infeasible. The FEA--AI hybrid variants retain most of the objective quality, and the agent-based variant nearly all of it, while roughly halving the computation. Their advantage grows as the evaluation count scales up, because the fixed training cost is spread over an increasing number of surrogate-served evaluations.

\subsection{Multi-objective optimization}

We now extend the four-way comparison to a multi-objective setting whose two competing objectives are the minimization of iron loss and the maximization of mean torque~\cite{ji2024design,vidanalage2018multimodal}, evaluated at a single operating point as in the single-objective study. Because the two objectives conflict, each run returns a Pareto front of non-dominated designs rather than a single best design, and the surrogate now predicts both objectives, each with its own predictive uncertainty (the deep ensemble of Section~2.2, one mean--variance output pair per objective).

The single-objective standard GA is replaced by NSGA-II~\cite{deb2002nsga2}, which ranks candidates by non-dominated sorting and crowding distance so that the search converges toward the Pareto front while preserving its spread. All other GA settings match the single-objective study of Section~5.1; only the multi-objective-intrinsic changes apply: survivor selection follows the NSGA-II non-dominated-sorting rule instead of elitist truncation, and the designs carried between rounds are the current Pareto-front members rather than the top-$K$ designs.

The FEA--AI hybrid model operates as in the single-objective case but now gates on both objectives, with the torque uncertainty rescaled to watts by the training-label-range ratio ($\approx4.2$\,W/Nm) so that the single threshold $\sigma_{\mathrm{th,switch}}$ applies to both. The threshold is set as before (rule-based $18$\,W from Section~4.1 versus per-round agent-based switching as in Section~4.2), and the same four strategies are compared under both budget definitions.

Because the outcome of each run is a Pareto front rather than a scalar, solution quality is assessed with two standard multi-objective performance indicators~\cite{audet2021performance}. The hypervolume (HV)~\cite{zitzler1999spea,zitzler2003performance} is the area of the objective-space region dominated by the obtained front and bounded by a fixed reference point $R=(360\,\mathrm{W},\,97\,\mathrm{Nm})$, chosen slightly worse than every compared solution so that all fronts dominate it. Both objectives are min--max normalized between an ideal point of $(60\,\mathrm{W},\,170\,\mathrm{Nm})$ and $R$ before the area is computed, so the reported HV lies in $[0,1]$, and HV is the only unary indicator known to be strictly Pareto-compliant~\cite{zitzler2003performance}. The inverted generational distance-plus (IGD+)~\cite{ishibuchi2015igdplus} is the mean dominance-compliant distance, in the same normalized space, from a reference front to the obtained front. Because the true Pareto front is unknown, the reference front is assembled separately for each budget comparison, as the non-dominated set of the FEA-verified solutions pooled across strategies and seeds (unverified AI-only predictions are excluded). Both indicators are reported together with the wall-clock time and are averaged over the random seeds. The AI-only front is the surrogate's own prediction; it is therefore reported as a predicted, FEA-unverified quantity and marked with an asterisk.

\subsubsection{Same FEA budget}

\begin{figure}[!htbp]
\centering
\includegraphics[width=\linewidth]{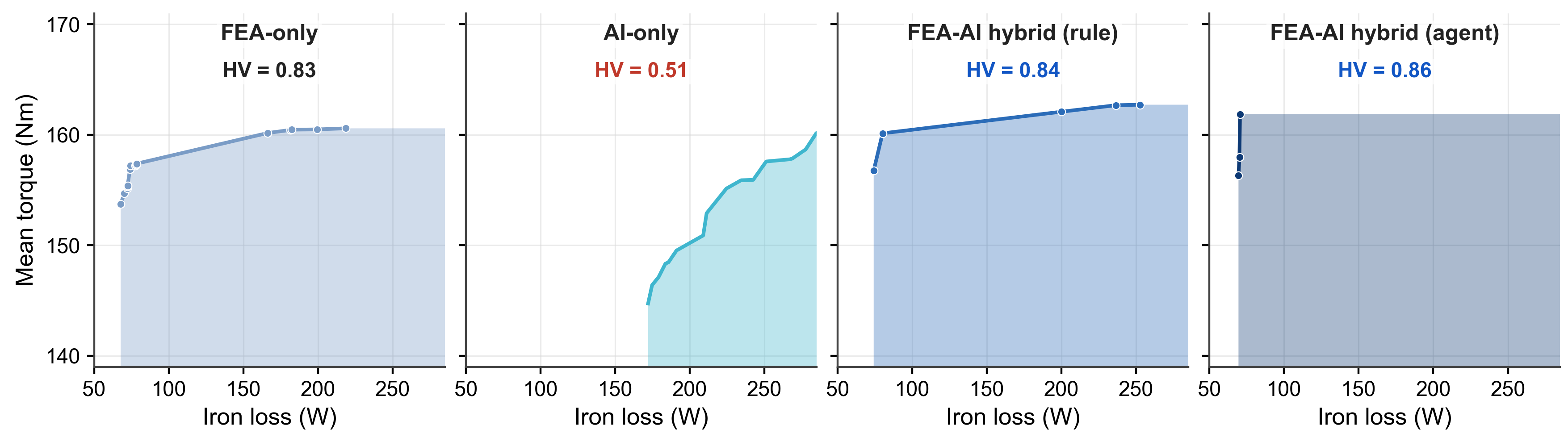}\\[1pt]
{\footnotesize (a) FEA budget of $100$ calls}\\[6pt]
\includegraphics[width=\linewidth]{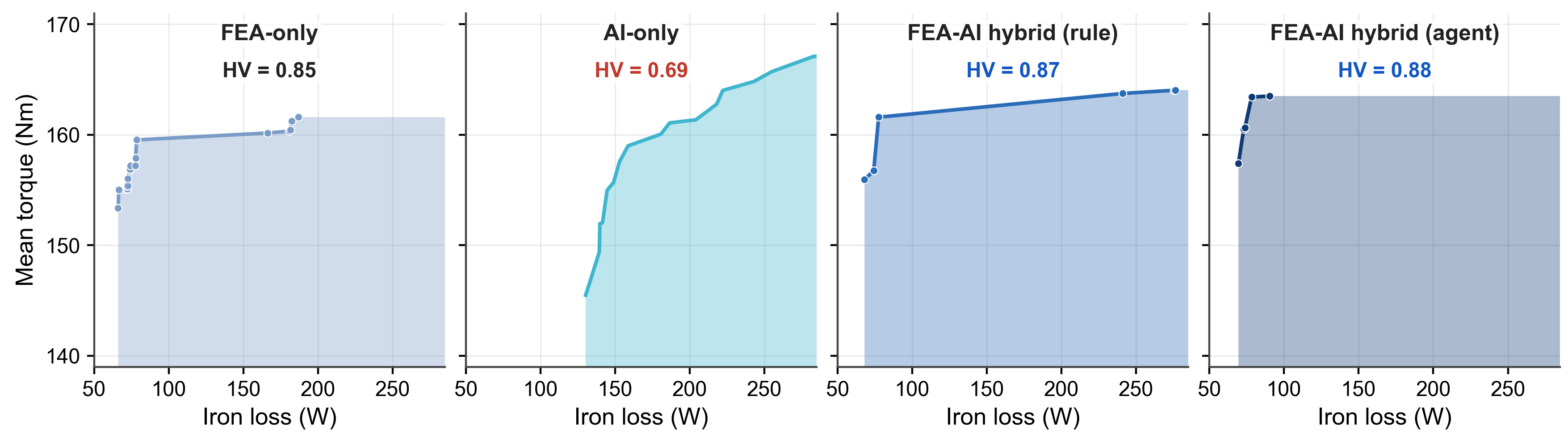}\\[1pt]
{\footnotesize (b) FEA budget of $150$ calls}\\[6pt]
\includegraphics[width=\linewidth]{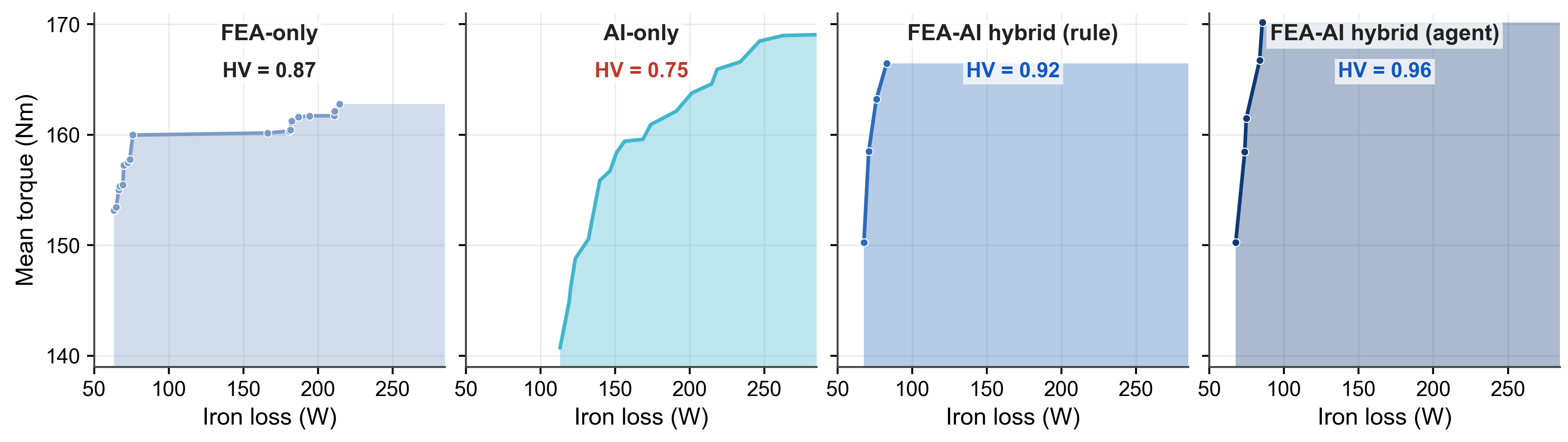}\\[1pt]
{\footnotesize (c) FEA budget of $200$ calls}
\caption{Pareto fronts of the four strategies at (a)~$100$, (b)~$150$, and (c)~$200$ FEA calls in the multi-objective same-FEA-budget study. Shaded areas give the hypervolume relative to $R=(360\,\mathrm{W},\,97\,\mathrm{Nm})$, and the AI-only front is surrogate-predicted and unverified. A wider shaded area indicates a better front, and the FEA--AI hybrid model attains the widest area, with its advantage growing as the FEA budget increases.}
\label{fig:mo_same_fea}
\end{figure}

The same-FEA design mirrors the single-objective study of Section~5.1.1: at each of the three budgets of $100$, $150$, and $200$ FEA calls, the FEA-only NSGA-II spends every call on direct evaluation, the AI-only search invests the entire budget in training samples and then searches by surrogate inference alone, and the two FEA--AI hybrid variants split the budget between surrogate training and uncertainty-triggered online correction exactly as in Section~5.1.1. Figure~\ref{fig:mo_same_fea} shows the resulting Pareto fronts for a representative seed at each budget, and Table~\ref{tab:mo_same_fea} reports the seed-average indicators.

\begin{table}[!htbp]
\centering
\caption{Multi-objective same-FEA-budget comparison: performance at budgets of $100$, $150$, and $200$ FEA calls; percentages are relative to FEA-only at the same budget.}
\label{tab:mo_same_fea}
\small
\begin{tabular}{llccc}
\toprule
Budget & Method & HV$_{\mathrm{norm}}$ $\uparrow$ & IGD+ $\downarrow$ & Time (h) $\downarrow$ \\
\midrule
\multirow{4}{*}{$100$}
 & FEA-only              & $0.673$ & $0.133$ & $3.76$ \\
 & AI-only$^{*}$          & $0.511$ ($-24.1\%$) & $0.256$ ($+92.5\%$) & $3.76$ ($+0.0\%$) \\
 & FEA--AI hybrid (rule)  & {\bfseries\boldmath $0.704$ ($+4.6\%$)}  & {\bfseries\boldmath $0.109$ ($-18.0\%$)} & $3.78$ ($+0.5\%$) \\
 & FEA--AI hybrid (agent) & {\bfseries\boldmath $0.702$ ($+4.3\%$)}  & {\bfseries\boldmath $0.122$ ($-8.3\%$)}  & $3.81$ ($+1.3\%$) \\
\midrule
\multirow{4}{*}{$150$}
 & FEA-only              & $0.704$ & $0.150$ & $5.78$ \\
 & AI-only$^{*}$          & $0.684$ ($-2.8\%$)  & $0.236$ ($+57.3\%$) & $5.79$ ($+0.2\%$) \\
 & FEA--AI hybrid (rule)  & {\bfseries\boldmath $0.714$ ($+1.4\%$)}  & $0.156$ ($+4.0\%$)  & $5.80$ ($+0.3\%$) \\
 & FEA--AI hybrid (agent) & {\bfseries\boldmath $0.708$ ($+0.6\%$)}  & $0.161$ ($+7.3\%$)  & $5.83$ ($+0.9\%$) \\
\midrule
\multirow{4}{*}{$200$}
 & FEA-only              & $0.726$ & $0.162$ & $7.63$ \\
 & AI-only$^{*}$          & $0.689$ ($-5.1\%$)  & $0.272$ ($+67.9\%$) & $7.66$ ($+0.4\%$) \\
 & FEA--AI hybrid (rule)  & {\bfseries\boldmath $0.889$ ($+22.5\%$)} & {\bfseries\boldmath $0.039$ ($-75.9\%$)} & $7.69$ ($+0.8\%$) \\
 & FEA--AI hybrid (agent) & {\bfseries\boldmath $0.822$ ($+13.2\%$)} & {\bfseries\boldmath $0.098$ ($-39.5\%$)} & $7.82$ ($+2.5\%$) \\
\bottomrule
\multicolumn{5}{@{}l}{\footnotesize $^{*}$~Surrogate-predicted values, not verified by FEA.}
\end{tabular}
\end{table}

\begin{figure}[!htbp]
\centering
\includegraphics[width=0.7\linewidth]{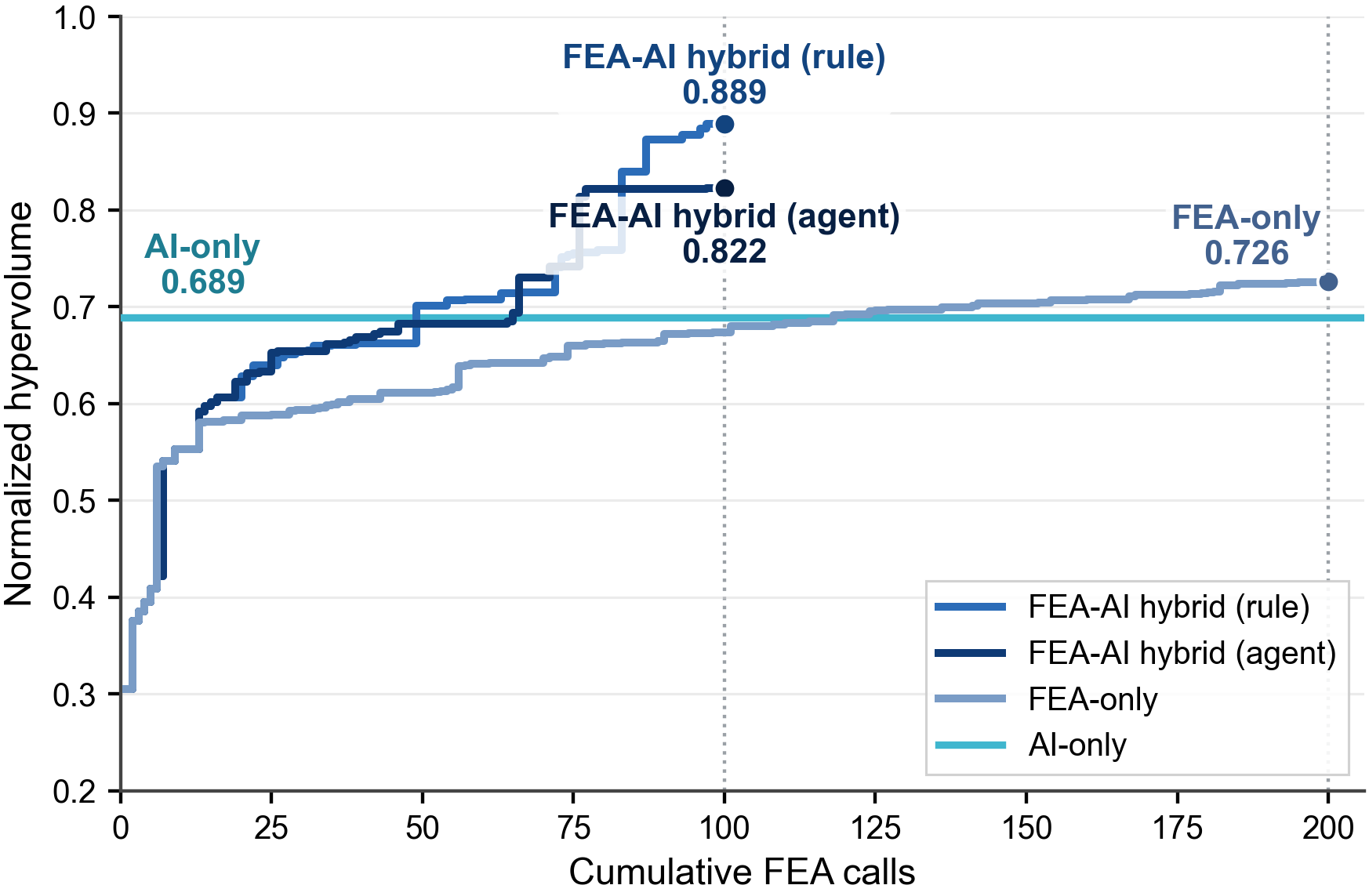}
\caption{Normalized hypervolume of the FEA-verified front versus cumulative online FEA calls at the $200$-call budget. The AI-only level is surrogate-predicted and unverified. Both FEA--AI hybrid variants exceed the final FEA-only front quality after roughly one third of its FEA calls and keep improving within the same budget.}
\label{fig:mo_hv_vs_fea}
\end{figure}

The single-objective trend carries over to the multi-objective case: the FEA--AI hybrid model improves monotonically with the FEA budget and progressively outperforms both baselines. At the $100$-call budget it already attains a slightly higher hypervolume and a better IGD+ than FEA-only; at the $150$-call budget the fronts are on par at identical cost; and at the $200$-call budget the advantage becomes decisive, with the rule-based variant reaching $22.5\%$ more hypervolume and a $75.9\%$ lower IGD+ than FEA-only. Notably, the same $50$-sample surrogate that misled the single-objective search proves beneficial at the smallest budget: a single-objective search depends on locating a single deep optimum, which a biased surrogate can misplace, whereas the hypervolume rewards the breadth of the front, and even a coarse surrogate widens the screened candidate pool while every front candidate is still FEA-verified by the outer loop. The mechanism is the same as in Section~5.1.1: the surrogate lets the FEA--AI hybrid variants screen several times more candidates and reserve FEA for the uncertainty-flagged ones. At the $200$-call budget, Fig.~\ref{fig:mo_hv_vs_fea} shows both variants passing, after roughly one third of the FEA-only call count, the front quality that FEA-only attains only at the end of its $200$ calls, and finishing far above it within their $100$-call online budget.

The AI-only front again proves unreliable: its predicted hypervolume falls below that of FEA-only at every budget and its IGD+ is the worst throughout. Re-validating all $57$ predicted Pareto designs of the $150$-call study with FEA shows that $28$ of them, about half, are infeasible geometries that cannot be simulated at all, and the surviving designs deviate from their predicted objectives by large margins. An AI-only front therefore cannot be used without complete FEA re-evaluation.

One ordering, however, reverses relative to the single-objective study: on the seed-average indicators, rule-based switching stays ahead of agent-based switching at every budget, even though the agent-based front is the best on the representative seed of Fig.~\ref{fig:mo_same_fea}. With a second objective the switching decision becomes harder: the single threshold must gate two rescaled uncertainties and the per-round signals are noisier, so the controller's reasoning is less effective than in the single-objective case. The threshold logs illustrate this behavior: on the longest runs the controller continues to raise the threshold in the late rounds just as the online budget is exhausted, so FEA verification nearly stops and the front stops improving, whereas the fixed $18$\,W rule continues to verify candidates until the end. Incorporating budget and objective awareness into the controller is a natural refinement.

\subsubsection{Same evaluation budget}

\begin{figure}[!htbp]
\centering
\includegraphics[width=\linewidth]{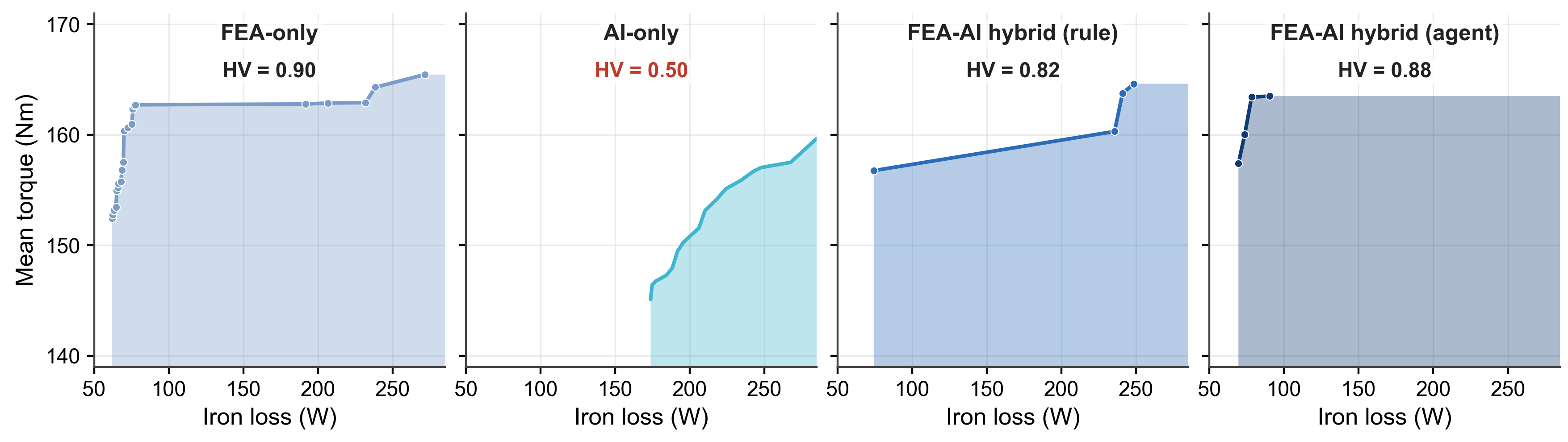}
\caption{Pareto fronts of the four strategies in the multi-objective same-evaluation study under $300$ evaluations per strategy. Shaded areas give the hypervolume relative to $R=(360\,\mathrm{W},\,97\,\mathrm{Nm})$, and the AI-only front is surrogate-predicted and unverified. The FEA--AI hybrid variants retain most of the FEA-only front quality with fewer than half the FEA calls and computation time.}
\label{fig:mo_same_eval}
\end{figure}

As in the single-objective study, every strategy is given the same budget of $300$ candidate evaluations, and the number of FEA calls is left free to vary. Figure~\ref{fig:mo_same_eval} shows the resulting Pareto fronts for a representative seed, and Table~\ref{tab:mo_same_eval} reports the four-seed-average indicators. For every surrogate-assisted strategy, the reported FEA count and time include the $100$ FEA samples ($\approx3.7$\,h) used to train the surrogate, so that each method is charged for the high-fidelity data it consumes.

\begin{table}[!htbp]
\centering
\caption{Multi-objective same-evaluation comparison: performance under $300$ candidate evaluations; FEA calls and time include the $100$ surrogate-training samples, and percentages are relative to FEA-only.}
\label{tab:mo_same_eval}
\small
\begin{tabular}{lcccc}
\toprule
Method & HV$_{\mathrm{norm}}$ $\uparrow$ & IGD+ $\downarrow$ & FEA calls $\downarrow$ & Time (h) $\downarrow$ \\
\midrule
FEA-only          & $0.752$ & $0.136$ & $300$   & $11.55$ \\
AI-only$^{*}$      & $0.486$ ($-35.4\%$) & $0.384$ ($+182.4\%$) & $100$ ($-66.7\%$)   & $3.83$ ($-66.8\%$) \\
FEA--AI hybrid (rule)  & $0.679$ ($-9.7\%$) & $0.198$ ($+45.6\%$) & $139.3$ ($-53.6\%$) & {\bfseries\boldmath $5.21$ ($-54.9\%$)} \\
FEA--AI hybrid (agent) & $0.692$ ($-8.0\%$) & $0.177$ ($+30.1\%$) & $140.0$ ($-53.3\%$) & {\bfseries\boldmath $5.24$ ($-54.6\%$)} \\
\bottomrule
\multicolumn{5}{@{}l}{\footnotesize $^{*}$~Surrogate-predicted values, not verified by FEA.}
\end{tabular}
\end{table}

Because all $300$ evaluations of the FEA-only NSGA-II are high-fidelity, it attains the best Pareto front, as expected, with a hypervolume of $0.752$ and an IGD+ of $0.136$, but at by far the highest cost of $300$ FEA calls and $11.55$\,h. The two FEA--AI hybrid variants reach a front of nearly the same quality: rule-based switching attains a hypervolume of $0.679$ and agent-based switching $0.692$, about $90$--$92\%$ of the FEA-only value, with IGD+ values of $0.198$ and $0.177$. They do so with fewer than half the FEA calls, $139$ and $140$ versus $300$, and less than half the computation time, $5.21$ and $5.24$\,h versus $11.55$\,h.

Between the two variants, agent-based switching is thus slightly better on both indicators at essentially the same FEA cost. The per-round threshold logs illustrate how the controller adapts. Reasoning over the calibration ratio $\rho$ of Eq.~\ref{eq:calib_ratio}, the controller adapts the threshold in opposite directions on different runs: on a run whose surrogate proves under-confident ($\rho<1$), it raises the threshold stepwise from $18$ to $22$\,W to stop spending FEA on predictions that are already accurate. On another run, by contrast, where calibration exposes strong over-confidence ($\rho$ up to ${\approx}2.3$), it cuts the threshold stepwise from $18$ to $12$ and then $10$\,W to force FEA verification of the newly discovered regions of the Pareto front. A fixed rule-based threshold cannot adapt in both directions across runs. On the same over-confident run, the rule-based $18$\,W policy leaves the miscalibration uncorrected ($\rho\approx1.8$--$1.9$) and its hypervolume stalls after the second round, whereas the agent-based variant's threshold cuts keep the hypervolume improving through the third and final round of the budget at a comparable FEA count.

Finally, the AI-only GA again proves unreliable: its FEA-unverified front (asterisked in Table~\ref{tab:mo_same_eval}) yields the worst indicators, a hypervolume of $0.486$ and an IGD+ of $0.384$, and none of its designs could be used without FEA re-validation.

Overall, the multi-objective results mirror the single-objective case: the FEA--AI hybrid model matches the FEA-only front quality at the smaller same-FEA budgets and clearly exceeds it at the largest, retains most of that quality at half the cost under the same evaluation budget, and its selective, uncertainty-guided use of FEA keeps it far more reliable than a surrogate-only search. Between the two variants, agent-based switching is ahead in three of the four studies (both single-objective studies and the multi-objective same-evaluation study), while rule-based switching stays ahead throughout the multi-objective same-FEA budgets, where the noisier two-objective signals weaken the controller's per-round reasoning; the switching principle itself transfers robustly from single- to multi-objective motor optimization.

\section{Conclusion}

This work presented an end-to-end multi-agent framework that specializes a local LLM into stage-dedicated agents through RAG, thereby automating the entire IPMSM design optimization workflow, from natural-language problem formulation and autonomous dataset construction to iterative optimization. Within this framework, the \textit{Optimization agent} employs an uncertainty-aware FEA--AI hybrid model, which switches each evaluation between the AI surrogate model and FEA according to predictive uncertainty, improving both efficiency and reliability of the search. The main findings are as follows:
\begin{itemize}
  \item \textbf{Effective domain-specialized RAG.} In a comparison of four LLM backbones (GPT-oss 20B, Llama-3 8B, GPT-5-mini, and Gemini-2.5-flash) on $90$ motor-domain questions, domain-specific textbook retrieval improved the performance of GPT-oss 20B over its No-RAG baseline on all three motor design question types: from near zero to $67\%$ on book-specific lookups, from $43\%$ to $77\%$ on numerical problems, and from $47\%$ to $80\%$ on conceptual questions. This quantitatively confirms the benefit of grounding the LLM in a single motor textbook through RAG for IPMSM design. Accordingly, we adopt this RAG-grounded LLM as the common backbone of all agents in the autonomous design workflow.
  \item \textbf{Autonomous design-space correction during training data resampling.} Starting from an improperly defined design space in which only $28\%$ of the geometries sampled for the AI training dataset were analysis-feasible, the agent-based resampling loop raised the geometry success ratio to $60\%$, $71\%$, and $84\%$ over three successive iterations, entirely without user intervention. At each iteration, the agent statistically attributes the infeasible geometries recorded in the validation log to the design variables responsible for them through ANOVA. It then narrows the bounds of these fail-prone variables while keeping the remaining ranges intact, ensuring that the design space is progressively constrained toward the feasible region without sacrificing sampling diversity.
  \item \textbf{Superior optimization performance under the same FEA budget.} Given the same number of FEA calls in the GA-based IPMSM design optimization process, the FEA--AI hybrid model reached up to $44\%$ lower iron loss than the FEA-only search on the single-objective problem and up to $22.5\%$ higher Pareto-front hypervolume on the multi-objective problem, surpassing the FEA-only front quality with roughly one third of its high-fidelity calls (Fig.~\ref{fig:mo_hv_vs_fea}). This gain in the FEA--AI hybrid model arises because the fast evaluations of the AI surrogate model enable a much broader exploration of the design space within the same high-fidelity budget.
  \item \textbf{Reduced FEA calls under the same evaluation budget.} Under the same number of GA evaluations, the FEA--AI hybrid model reduced the number of FEA calls by more than half and thereby shortened the total computation time by $52$--$55\%$. Nevertheless, the solution quality remained close to that of the FEA-only search, staying within about $1\%$ of its optimum on the single-objective problem and retaining $90$--$92\%$ of its hypervolume on the multi-objective problem. This saving efficiency increases even more when the active learning progresses since it lowers the predictive uncertainty of the AI surrogate model, so that fewer evaluations trigger FEA calls.
  \item \textbf{Reliable optimization framework.} The comprehensive ablation results reveal a clear contrast among the three strategies. 1) The FEA-only search finds accurate optimal solutions but at a prohibitive computational cost. 2) The AI-only search, in contrast, is fast but unreliable: it converges to false optima (Fig.~\ref{fig:ai_only_rounds}), produces the worst FEA-validated designs under the same evaluation budget, and about half ($49\%$) of its predicted Pareto designs prove infeasible. 3) The FEA--AI hybrid model achieves the best overall trade-off among solution quality, reliability, and computation cost. This advantage is attributed to its uncertainty-gated FEA verification: fast surrogate evaluations drive the exploration, while FEA is selectively applied where the predictions are least reliable.
\end{itemize}

A primary limitation of this study is that the framework operates on a parameterized IPMSM geometry; thus, the design space is spanned by twelve template variables, leaving free-form or topology-level design changes outside its current scope. Future work will therefore extend the framework toward more general geometry representations. More broadly, none of its components is specific to motors: with the domain reference, the parametric template, and the solver exchanged, the same combination of a domain-specialized local LLM, RAG-grounded multi-agent collaboration, and the uncertainty-aware FEA--AI hybrid model can be extended to other simulation-driven design domains while keeping proprietary data in-house. An interactive, browser-based demonstration of the full pipeline is described in Appendix~\ref{app:gui}.

\section*{Acknowledgments}
This work was supported by grants from the Ministry of Science and ICT (GTL24033-000, N10250154, and No. 2022-0-00986), the Ministry of Trade, Industry and Energy (RS-2025-02317327 and RS-2025-25444634), the Ministry of Oceans and Fisheries (PET0050), and Korea Hydro \& Nuclear Power Co., Ltd. (No. 8-Tech-07).

\section*{CRediT authorship contribution statement}
\textbf{Jinseong Han:} Conceptualization, Methodology, Software, Validation, Formal analysis, Investigation, Data curation, Writing -- original draft, Writing -- review \& editing, Visualization.
\textbf{Sunwoong Yang:} Writing -- review \& editing, Supervision, Methodology, Conceptualization.
\textbf{Namwoo Kang:} Writing -- review \& editing, Supervision, Project administration, Funding acquisition.

\section*{Declaration of competing interest}
The authors declare that they have no known competing financial interests or personal relationships that could have appeared to influence the work reported in this paper.

\section*{Data availability}
Data will be made available on request.

\appendix
\section{Problem-definition agent interactions}
\label{app:problem_def}

\begin{figure}[!htbp]
\centering
\includegraphics[width=0.98\linewidth]{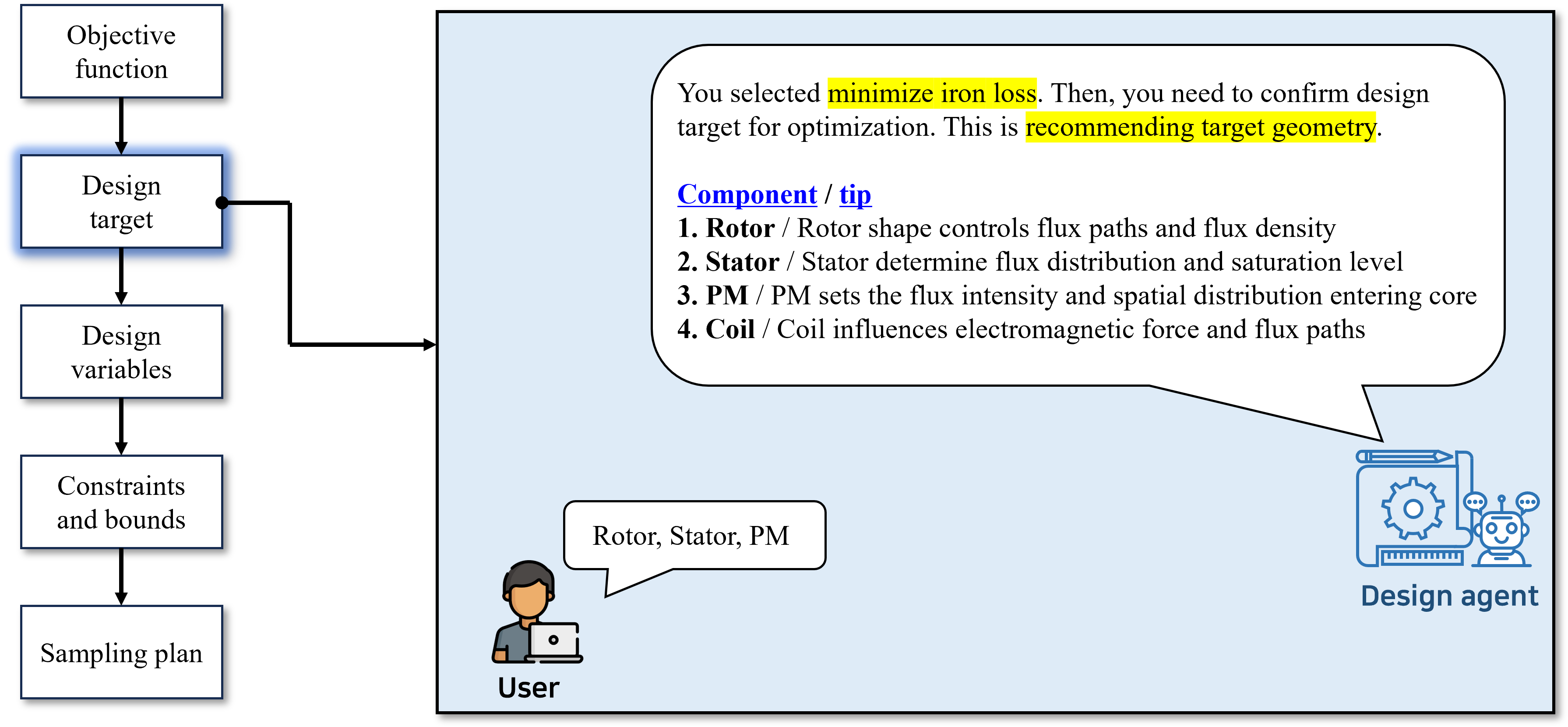}
\caption{\textit{Problem-definition agent} interaction for the \emph{design-target selection} stage.}
\label{fig:pd_design_target}
\end{figure}

\begin{figure}[!htbp]
\centering
\includegraphics[width=0.98\linewidth]{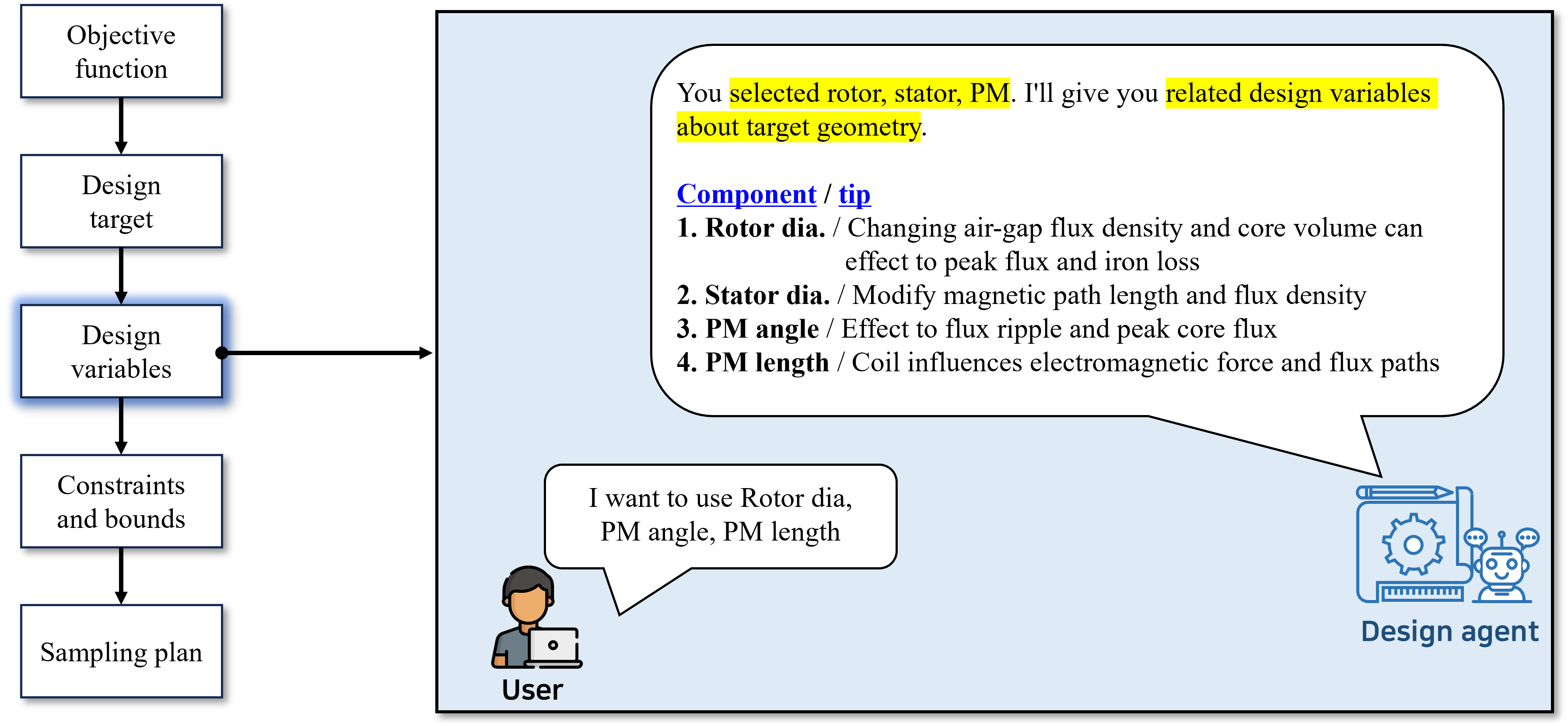}
\caption{\textit{Problem-definition agent} interaction for the \emph{design-variable selection} stage.}
\label{fig:pd_design_variables}
\end{figure}

\begin{figure}[!htbp]
\centering
\includegraphics[width=0.98\linewidth]{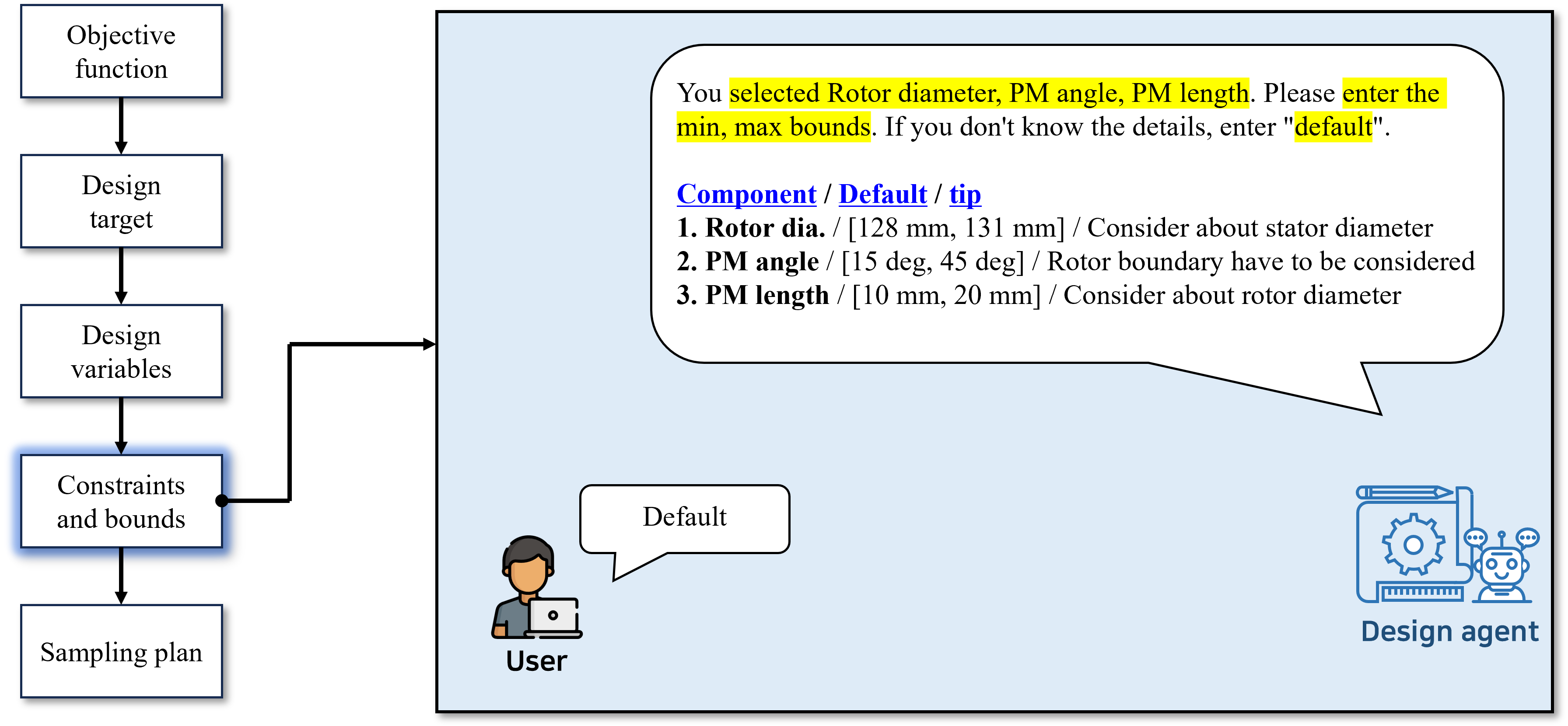}
\caption{\textit{Problem-definition agent} interaction for the \emph{constraints-and-bounds} stage.}
\label{fig:pd_constraints}
\end{figure}

\begin{figure}[!htbp]
\centering
\includegraphics[width=0.98\linewidth]{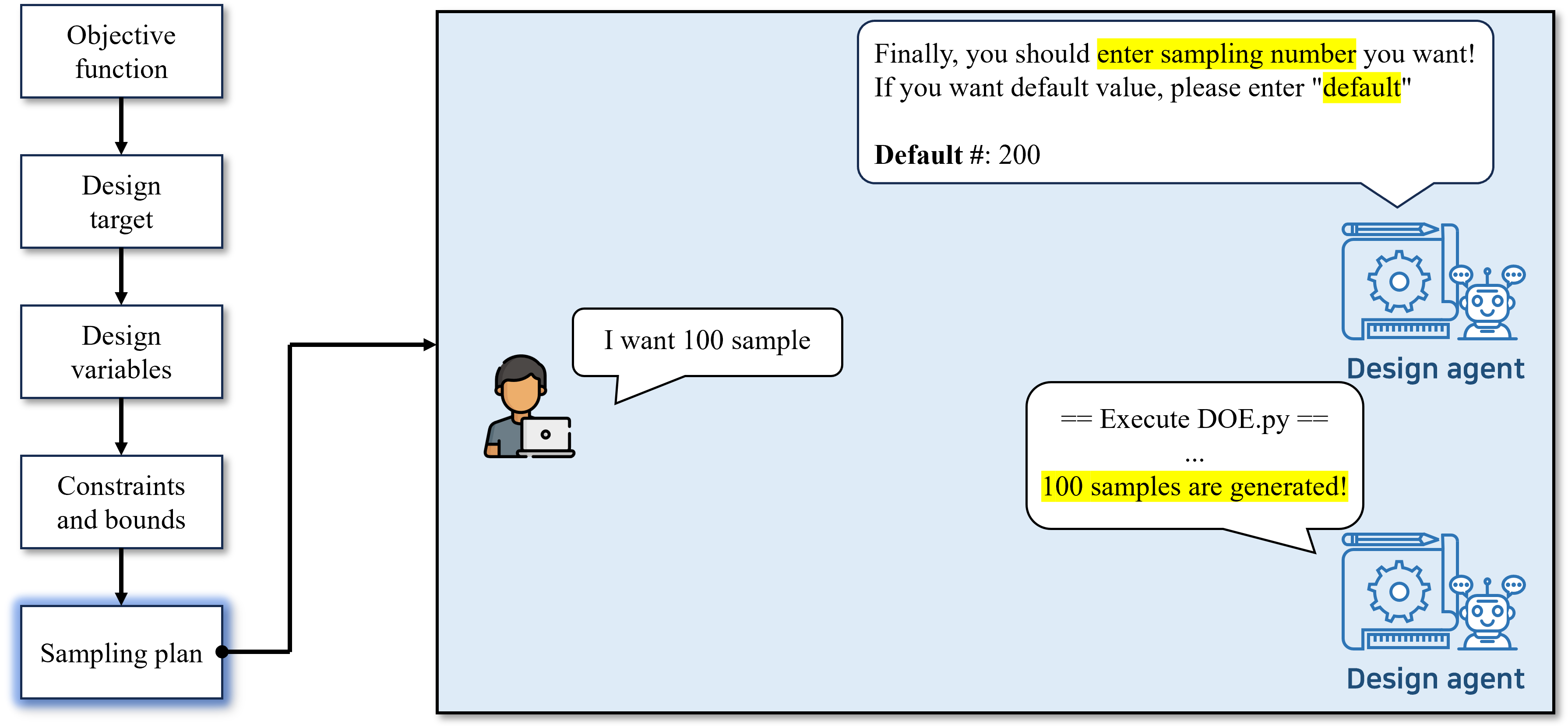}
\caption{\textit{Problem-definition agent} interaction for the \emph{sampling-plan} stage.}
\label{fig:pd_sampling_plan}
\end{figure}

The \textit{problem definition agent} of Section~2.1 guides the user through a five-stage decision flow (objective function $\rightarrow$ design target $\rightarrow$ design variables $\rightarrow$ constraints and bounds $\rightarrow$ sampling plan). The objective-selection stage is shown in the main text (Fig.~\ref{fig:design_agent_chat}); the remaining four stages are collected here in Figs.~\ref{fig:pd_design_target}--\ref{fig:pd_sampling_plan}. In each figure the flow diagram on the left marks the active stage, and the panel on the right shows the corresponding retrieval-grounded interaction, in which the agent returns component or variable candidates with engineering tips and the user confirms or overrides the recommendation before the flow advances to the next stage.

\section{Full response comparison for the RAG saliency question}
\label{app:rag_response}

\begin{figure}[!htbp]
\centering
\includegraphics[width=0.98\linewidth]{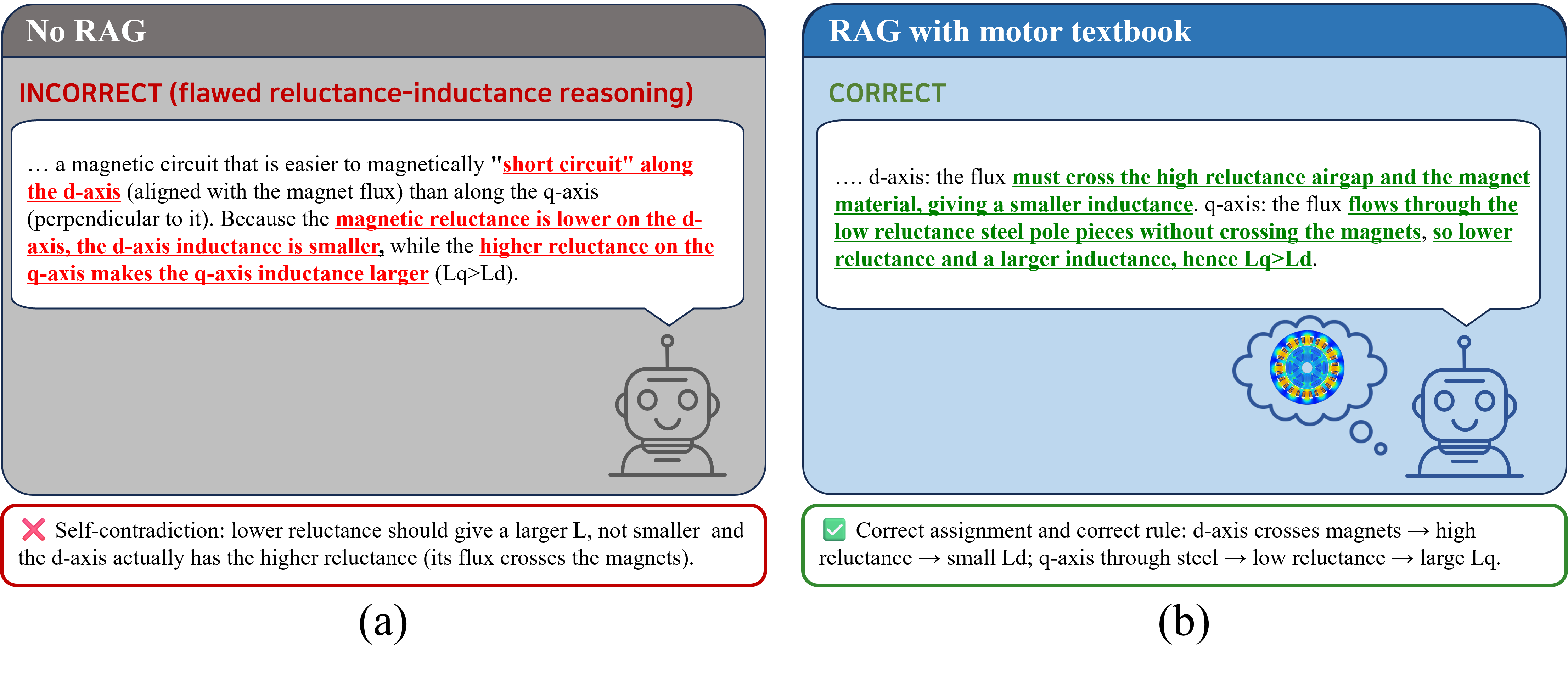}
\caption{Response comparison for an IPMSM saliency question ($L_q>L_d$ and reluctance torque) in the RAG study: (a)~GPT-oss 20B without RAG (mechanistically incorrect); (b)~with motor-textbook RAG (correct mechanism)~\cite{gieras2009permanent}.}
\label{fig:rag_saliency_comparison}
\end{figure}

Complementing the per-question examples of Table~\ref{tab:rag_examples}, Fig.~\ref{fig:rag_saliency_comparison} reproduces the full generated responses for the saliency prompt of Section~3.1 (\textit{``In an IPMSM, why is the $q$-axis inductance larger than the $d$-axis inductance ($L_q>L_d$), and what is reluctance torque?''}) under the same local GPT-oss 20B backbone, (a)~without RAG and (b)~with motor-textbook RAG. The ungrounded answer is fluent but reaches the correct conclusion through a physically inverted argument, whereas the retrieval-grounded answer explains the low-reluctance $q$-axis versus high-reluctance $d$-axis mechanism correctly and defines reluctance torque accordingly.

\section{Interactive multi-agent demonstration interface}
\label{app:gui}

To improve the usability of the proposed framework for non-expert users and to make its behavior easy to inspect and reproduce, we additionally developed a lightweight, browser-based demonstration interface that animates the full multi-agent pipeline in operation. The interface runs entirely in a web browser without any installation and is publicly available at \url{https://hanjinseong.github.io/ipmsm-design/}.

\begin{figure}[!htbp]
\centering
\includegraphics[width=0.98\linewidth]{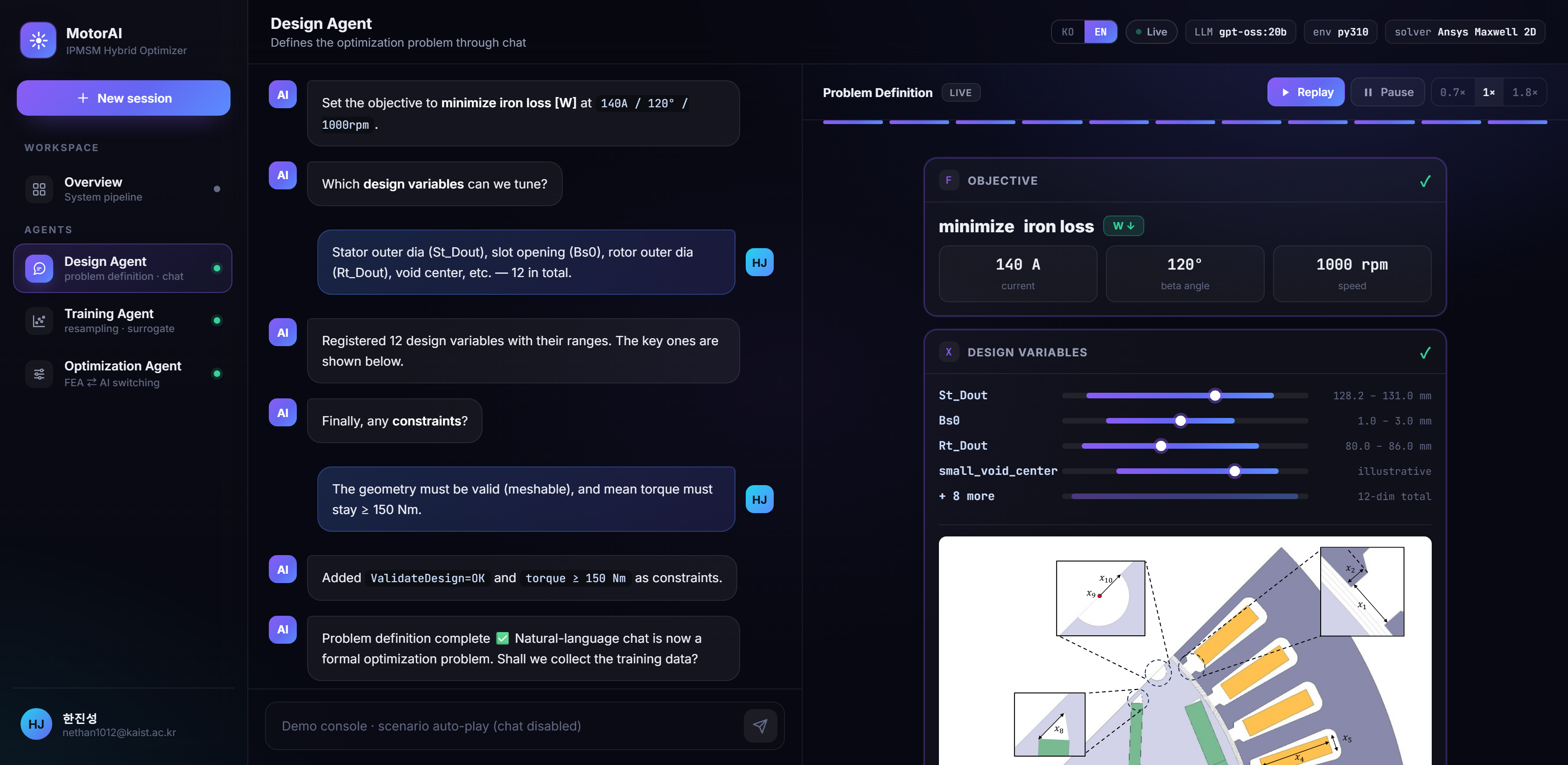}
\caption{\textit{Design-agent} view of the demonstration interface: natural-language problem formulation and the live optimization card.}
\label{fig:gui_design}
\end{figure}

\begin{figure}[!htbp]
\centering
\includegraphics[width=0.98\linewidth]{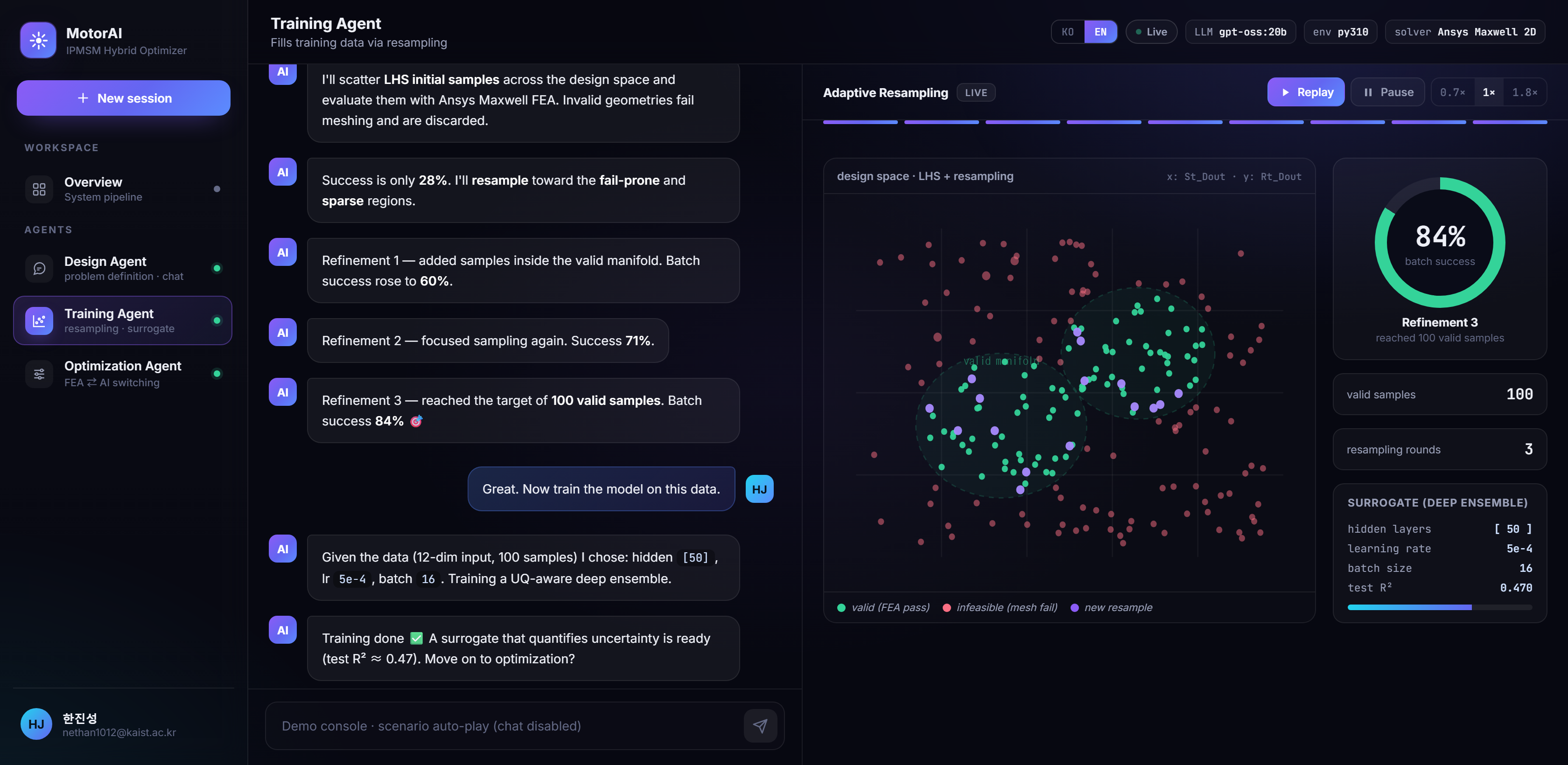}
\caption{\textit{Training-agent} view of the demonstration interface: log-informed resampling and surrogate training.}
\label{fig:gui_training}
\end{figure}

\begin{figure}[!htbp]
\centering
\includegraphics[width=0.98\linewidth]{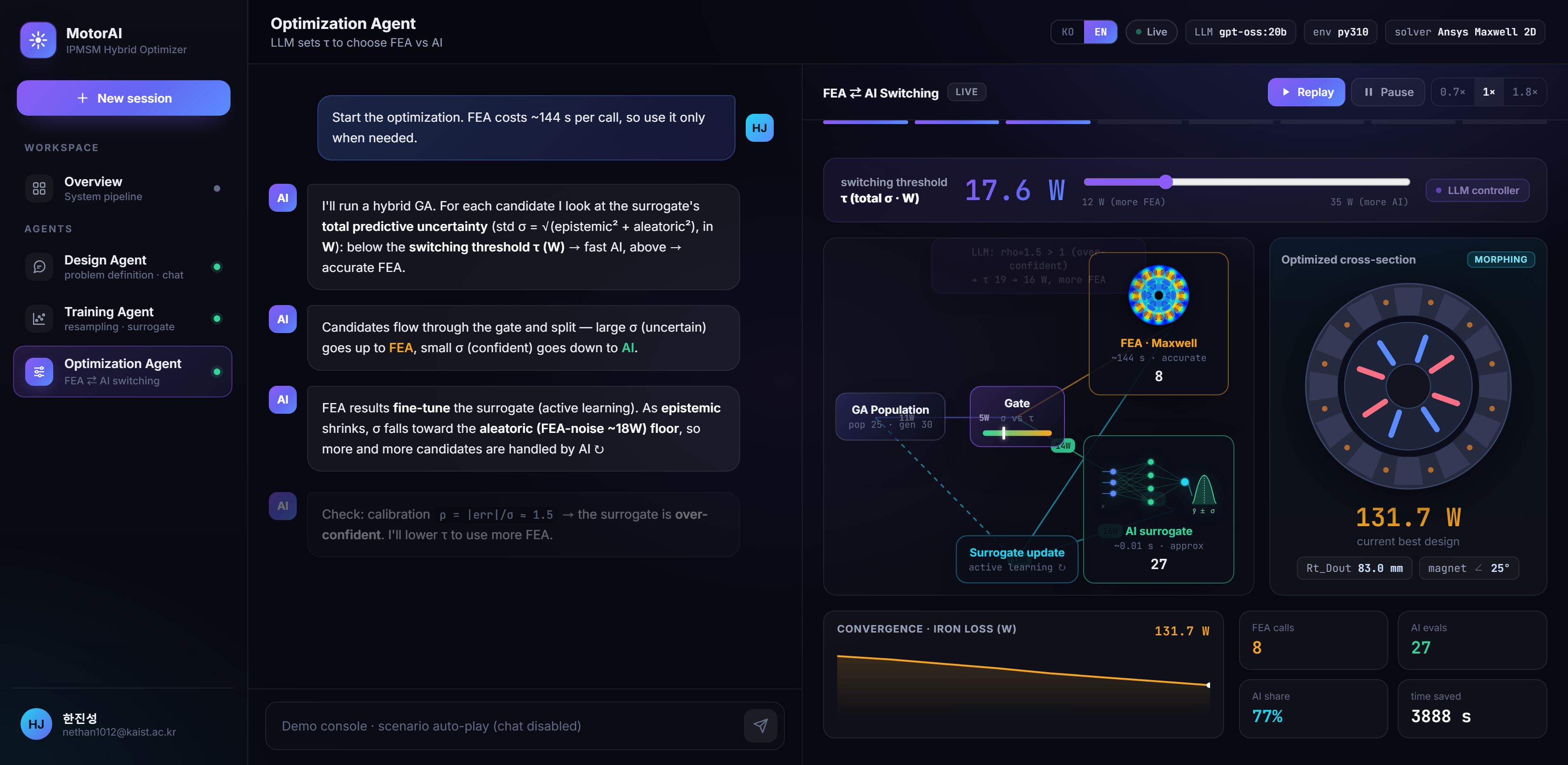}
\caption{\textit{Optimization-agent} view of the demonstration interface, publicly available at \url{https://hanjinseong.github.io/ipmsm-design/}: LLM-narrated walkthrough of the uncertainty-aware FEA--AI switching.}
\label{fig:gui_homepage}
\end{figure}

A left-hand panel exposes the three agents of the framework, and selecting an agent opens a step-by-step, LLM-narrated walkthrough of that stage with a live visualization on the right. Figures~\ref{fig:gui_design}--\ref{fig:gui_homepage} show the three agent views in turn: the \textit{Design agent} (problem definition), the \textit{Training agent} (resampling and surrogate construction), and the \textit{Optimization agent} (uncertainty-aware FEA--AI switching).

The \textit{Design-agent} view (Fig.~\ref{fig:gui_design}) mirrors the problem-definition flow of Section~2.1: through natural-language chat the user fixes the objective (minimize iron loss at a stated operating point), the design variables and their ranges, and the constraints, which the right-hand panel consolidates into a live optimization card together with the parameterized IPMSM geometry. The \textit{Training-agent} view (Fig.~\ref{fig:gui_training}) animates the log-informed resampling loop of Sections~2.2 and~3.2: LHS samples are scattered over the design space, infeasible geometries are discarded, and successive resampling refinements raise the geometry success ratio from $28\%$ to $84\%$ until the target of $100$ valid samples is reached, after which the UQ-aware deep-ensemble surrogate is trained.

The \textit{Optimization-agent} view (Fig.~\ref{fig:gui_homepage}) gives an LLM-narrated walkthrough of the FEA--AI hybrid search: the agent explains how each candidate's total predictive standard deviation $\sigma$ (the square root of the aleatoric-plus-epistemic variance of Eq.~\ref{eq:ensemble_uncertainty_decomposition}) is compared against the switching threshold $\sigma_{\mathrm{th,switch}}$ (denoted $\tau$ in the interface), how low-uncertainty candidates are served by the surrogate while high-uncertainty ones are sent to high-fidelity FEA, how the FEA labels fine-tune the surrogate through active learning, and how the calibration ratio is monitored to retighten the threshold when the surrogate becomes over-confident. The right-hand panel provides a live visualization of this process: an adjustable switching-threshold slider, an animated FEA\,$\rightleftarrows$\,AI gate that routes the GA population either to the Ansys Maxwell solver or to the deep-ensemble surrogate, a convergence trace of the iron-loss objective, the continuously morphing optimized cross-section, and running counters for FEA calls, AI evaluations, the AI-handled fraction, and the wall-clock time saved relative to an all-FEA search. This walkthrough mirrors the switching and active-learning mechanism of Section~2.3 and the calibration-based retightening of Section~4.2, letting a user watch, in real time, how uncertainty-driven switching allocates a limited FEA budget and how the search converges toward a reliable optimum.

\clearpage
\bibliographystyle{unsrt}
\bibliography{references}

\end{document}